%% file: main.tex
\documentclass{article} 
\usepackage{iclr2025_conference,times}

\input{macros}

\usepackage{natbib}
\usepackage{titletoc}
\usepackage{colortbl}
\usepackage[capitalize]{cleveref}
\usepackage{lscape}

\usepackage{mathtools}
\usepackage{manfnt}
\newcommand{\cube}[1]{\text{\mancube}^{#1}}

\theoremstyle{plain}
\newtheorem{theorem}{Theorem}[section]
\newtheorem{proposition}[theorem]{Proposition}

\theoremstyle{definition}
\newtheorem{definition}[theorem]{Definition}

\theoremstyle{remark}

\makeatletter
\def\@fnsymbol#1{\ensuremath{\ifcase#1\or \dagger\or \ddagger\or
   \mathsection\or \mathparagraph\or \|\or **\or \dagger\dagger
   \or \ddagger\ddagger \else\@ctrerr\fi}}
    \makeatother

\NewDocumentCommand{\colorlinkmailto}{m m m}{%
    \href{mailto:#1}{\textcolor{#2}{#3}}%
}

\definecolor{mediumblue}{RGB}{100, 149, 237}

\title{Transformers Learn Low Sensitivity \\Functions: Investigations and Implications }
\author{Bhavya Vasudeva$^\dagger$,  
Deqing Fu$^\dagger$, 
Tianyi Zhou,  Elliott Kau$^*$, 
Youqi Huang$^*$, 
Vatsal Sharan\\
University of Southern California\\
{\fontsize{9}{11}\colorlinkmailto{bvasudev@usc.edu}{black}{\texttt{bvasudev@usc.edu}}, \colorlinkmailto{deqingfu@usc.edu}{black}{\texttt{deqingfu@usc.edu}}}
}

\begin{document}

\maketitle
\footnotetext{$^\dagger$Co-first authors. $^*$Co-third authors. $^{1}$The code is available at \url{https://github.com/estija/sensitivity}.}
\input{sections/abstract}

\input{sections/introduction}

\section{Related Work}
We discuss related work on understanding transformers, sensitivity and spectral bias, and simplicity bias in deep learning in this section, and related work on implicit biases of gradient methods, the robustness of transformers, spurious correlations, and data augmentation in \cref{app:rel-work}.

\input{sections/related-work}

\input{sections/spectral-bias}

\input{sections/beyond-boolean}

\input{sections/toy-expts}

\input{sections/vision-expts}

\input{sections/nlp-expts}
\input{sections/implications}

\input{sections/conclusion}
\section*{Acknowledgements}
The authors thank the anonymous reviewers for helpful suggestions and feedback. This work was supported by an NSF CAREER Award CCF-2239265, an Amazon Research Award, \dfedit{an Open Philanthropy research grant, and a Google Research Scholar Award}. YH was supported by a USC CURVE fellowship. The authors acknowledge the use of USC CARC's Discovery cluster and the USC NLP cluster. This work was done in part while BV, DF, TZ, and VS were visiting the Simons Institute for the Theory of Computing. 

\bibliography{citations}
\bibliographystyle{iclr2025_conference}
\newpage
\appendix
\onecolumn

\section*{Appendix}
\startcontents[appendix]
\addcontentsline{toc}{chapter}{Appendix}
\renewcommand{\thesection}{\Alph{section}} 

\printcontents[appendix]{}{1}{\setcounter{tocdepth}{3}}

\setcounter{section}{0}
\input{sections/extra-results}
\input{sections/proof}
\input{sections/rel-work-app}
\input{sections/exp-settings}
\input{sections/limitations}


\end{document}

%% file: macros.tex
\usepackage{latexsym}
\usepackage{graphicx,amsmath,
amsfonts,amssymb,amsthm,bm,
url,breakurl,epsf,color,
caption,multirow,comment,boldline}
\usepackage[backref=page]{hyperref}
\usepackage{newtxtt}
\usepackage{xspace}

\hypersetup{
    colorlinks=true,
    linkcolor=[rgb]{0.87,0.0,0.45},
    filecolor=magenta,      
    urlcolor=[rgb]{0.3,0.75,0.7},
    citecolor=[rgb]{0.1,0.3,0.85}
    }

\usepackage{epsfig}
\usepackage[shortlabels]{enumitem}
\everymath=\expandafter{\the\everymath\displaystyle}
\DeclareMathSizes{24}{24}{24}{24}

\usepackage[suppress]{color-edits}
\addauthor[Bhavya]{bv}{magenta}
\addauthor[Deqing]{df}{orange}
\addauthor[Tianyi]{tz}{red}
\addauthor[Youqi]{yh}{green}
\addauthor[Elliott]{ek}{pink}
\addauthor[Vatsal]{vs}{blue}

\usepackage{bbm,dsfont}
\usepackage{booktabs,siunitx}
\usepackage{isomath}
\usepackage{upgreek}
\usepackage{wrapfig}
\usepackage{cite}
\usepackage{nicefrac}       
\usepackage{microtype}      
\usepackage{subfigure,subcaption}
\usepackage[bottom,hang,flushmargin,multiple,symbol]{footmisc} 
\usepackage{fnpct}

\usepackage[utf8]{inputenc} 
\usepackage[T1]{fontenc}    
\usepackage{amsopn,amstext}       
\usepackage{graphics}

\newcommand{\vct}[1]{\boldsymbol{#1}} 
\newcommand{\mat}[1]{\boldsymbol{#1}} 

\newcommand{\field}[1]{\mathbb{#1}}
\newcommand{\R}{\field{R}} 
\newcommand{\E}{\field{E}} 

\newcommand{\norm}[1]{\left\|#1\right\|}

\DeclareMathOperator{\prob}{Pr}

\newtheorem*{remark*}{Remark}

\newcommand{\x}{{\vct{x}}}

\newcommand{\y}{{\vct{y}}}

\newcommand{\calP}{{\mathcal{P}}}

\newcommand{\ind}{\mathds{1}}

\newcommand{\sign}{\textsc{sign}}

\newcommand{\mtx}[1]{\bm{#1}}
\newcommand{\attn}{\mathrm{ATTN}\xspace}
\newcommand{\WV}{\mtx{W}_{V}}
\newcommand{\WK}{\mtx{W}_{K}}
\newcommand{\WQ}{\mtx{W}_{Q}}
\newcommand{\Tau}{T}
\newcommand{\U}{\mtx{U}}

\newcommand{\sft}[1]{\bm{\varphi}(#1)}

\newcommand{\X}{{\mtx{X}}}
\newcommand{\Tb}{{\mtx{\theta}}}
\newcommand{\f}{\Phi}
\newcommand{\inpb}[2]{\left\langle #1, #2 \right\rangle}
\newcommand{\concat}{\operatorname{concat}\xspace}
\newcommand{\xb}{\bm{x}}

\newcommand{\e}{\bm{e}}
\newcommand{\V}{\mathcal{V}}
\newcommand{\Vsp}{\V_{\text{sparse}}}
\newcommand{\Vfr}{\V_{\text{frequent}}}
\newcommand{\Vir}{\V_{\text{irrelevant}}}
\newcommand{\I}{\mathcal{I}}
\newcommand{\Isp}{\I_{\text{sparse}}}
\newcommand{\Ifr}{\I_{\text{frequent}}}
\newcommand{\Iirr}{\I_{\text{irrelevant}}}

\newcommand{\vb}{\bm{v}}
\newcommand{\vs}{\vb_{\text{sp}}}
\newcommand{\vf}{\vb_{\text{freq}}}
\newcommand{\vi}{\vb_{\text{irrel}}}

\newcommand{\D}{\mathcal{D}}
\newcommand{\Pc}{\mathcal{P}}
\newcommand{\Nc}{\mathcal{N}}
\newcommand{\db}{\tilde{d}}

\newcommand{\calN}{{\mathcal{N}}}

\usepackage{duckuments}
\usepackage{dblfloatfix} 

\renewcommand{\paragraph}[1]{\textbf{#1}}

\newcounter{this-list}
\newenvironment{tightenumerate}{
\vspace{2pt}
\begin{list}{\arabic{this-list}.}{\usecounter{this-list}
                                 \setcounter{this-list}{0}
  \setlength{\itemsep}{0pt}%
  \setlength{\parsep}{0pt}%
  \setlength{\topsep}{0pt}%
    \setlength{\partopsep}{0pt}%
  \setlength{\leftmargin}{3.5ex}%
  \setlength{\labelwidth}{4.5ex}%
  \setlength{\labelsep}{1ex}%
}} {\end{list}\vspace{2pt}}

%% file: sections/abstract.tex
\begin{abstract}
Transformers achieve state-of-the-art accuracy and robustness across many tasks, but an understanding of their inductive biases and how those biases differ from other neural network architectures remains elusive. In this work, we identify the sensitivity of the model to token-wise random perturbations in the input as a unified metric which explains the inductive bias of transformers across different data modalities and distinguishes them from other architectures. We show that transformers have lower sensitivity$^1$ than MLPs,  CNNs, ConvMixers and LSTMs, across both vision and language tasks. We also show that this low-sensitivity bias has important implications: i) lower sensitivity correlates with improved robustness; it can also be used as an efficient intervention to further improve the robustness of transformers; ii) it corresponds to flatter minima in the loss landscape; and iii) it can serve as a progress measure for grokking. We support these findings with theoretical results showing (weak) spectral bias of transformers in the NTK regime, and improved robustness due to the lower sensitivity. 
\end{abstract}

%% file: sections/introduction.tex
\section{Introduction} \label{sec:intro}

Transformers, originally introduced for language problems \citep{Vaswani2017AttentionIA}, have become a universal backbone across machine learning --- including applications such as vision \citep{dosovitskiy2021an} and protein structure prediction \citep{Jumper2021HighlyAP}.  Several recent works have also found that not only do transformers achieve better accuracy, but they are also more robust to various corruptions and changes in the data distribution \citep{shao2021adversarial,mahmood2021robustness,bhojanapalli2021understanding,paul2022vision}.  
Despite their practical success, relatively little is understood about what distinguishes transformers from other neural network architectures.  \emph{ If a transformer and an alternative neural network architecture (such as a CNN or an LSTM) are trained to obtain similar training accuracy on a dataset, then how do the models differ in terms of the functions they learn? Equivalently, what inductive biases do transformers have which distinguish them from other architectures?} 

Recently, for the setting of Boolean inputs, \citet{bhattamishra2023simplicity} and \citet{hahn-rofin-2024-sensitive} suggest using the notion of \emph{sensitivity} to distinguish transformers from other candidate architectures. The sensitivity of a function measures how likely the output is to change for random changes to the input. \citet{bhattamishra2023simplicity} and \citet{hahn-rofin-2024-sensitive}  show that transformers are biased to learn functions with low sensitivity on Boolean inputs.   
Sensitivity has several desirable properties as a notion of inductive  bias. It is closely related to the Fourier representation of the function and various other notions of Boolean function complexity such as the degree of the function and the size of the smallest decision tree which represents the function \citep{o'donnell_2014}. Low sensitivity functions correspond to low complexity functions based on all these notions of Boolean function complexity, and hence an inductive bias towards low-sensitivity functions is regarded as an instance of `simplicity bias' of the model \citep{valle2018deep,bhattamishra2023simplicity}. Sensitivity also has deep connections to well-studied notions of inductive biases such as spectral bias \citep{rahaman2019spectral}, which is a bias towards `simple' functions such as low frequency functions in the Fourier space.
Sensitivity has also been found to correlate with better generalization for fully-connected networks \citep{novak2018sensitivity}. \bvedit{(See \cref{app:rel-work} for a detailed discussion of the related work.)} 

\paragraph{Our results.} Sensitivity is a promising notion of inductive bias, but has mainly  been investigated for Boolean functions so far. Given the numerous modalities of data across which transformers are successful in practice, the goal of our work is to examine if appropriate extensions of the notion of sensitivity for Boolean functions help understand the inductive  bias of transformers across varied data modalities --- and if these notions help explain properties of transformers such as their improved robustness. We now provide an overview of the main claims and results of the paper. We begin our investigation with the following question:

    \begin{center}
        \emph{What are appropriate notions of sensitivity beyond Boolean data?}
    \end{center}
    
To provide a concrete starting point where we can understand the properties of sensitivity with theoretical analysis, we first consider the Boolean setup and place the low-sensitivity bias of transformers on a firmer theoretical foundation in that setting (Section \ref{sec:th-spec}). Using prior work on spectral bias in neural networks \citep{yang2020finegrained} we prove that transformers show a low-sensitivity bias on Boolean functions, and also prove that low sensitivity leads to better robustness. We then consider the above question, and propose a suitable notion of sensitivity which takes into account the underlying metric space (Section \ref{sec:sensitivity_def}). To investigate if sensitivity is a suitable notion beyond the Boolean case, we first consider a synthetic dataset where we can tease apart sensitivity from related notions which can coincide for Boolean functions --- such as a preference towards functions that depend on a sparse set of tokens. We show that transformers prefer to learn low-sensitivity functions (even if they are not sparse). Subsequently, we examine if this low-sensitivity bias is widely present across different tasks:

\begin{center}
    \emph{Does low-sensitivity serve as a unified notion of simplicity across vision and language tasks, and does it distinguish between transformers and other architectures?}
\end{center}

Here, we first conduct experiments on vision datasets. We empirically compare \dfedit{(Vision-)}Transformers with MLPs, CNNs, and ConvMixers, and observe that transformers have lower sensitivity compared to other candidate architectures (see \Cref{sec:expts-vision}). 
Similarly, we conduct experiments on language tasks and observe that transformers learn predictors with lower sensitivity than LSTM models. Furthermore, transformers tend to have uniform sensitivity to all tokens while LSTMs are more sensitive to more recent tokens (see \Cref{sec:expts-language}). Given this \bvedit{low-}sensitivity bias, we next examine its implications:

\begin{center}
    \emph{What are the implications of a bias towards low-sensitivity functions; is it helpful 
    in certain settings?}
\end{center}

We study this in three contexts: robustness, properties of the loss landscape, and training dynamics.

\begin{tightenumerate}

\item Lower Sensitivity Correlates with Better Robustness: We show that transformers have lower sensitivity and are more {robust} to corruptions when tested on the  CIFAR-10-C dataset, compared to CNNs 
(\cref{sec:implications}). We also demonstrate that sensitivity is not only predictive of robustness but also has prescriptive power: We add a regularization term at training time to encourage the model to have \bvreplace{less}{lower} sensitivity. Since sensitivity is efficient to measure empirically, this is easy to accomplish via data augmentation. We find that models explicitly trained to have lower sensitivity yield even better robustness on CIFAR-10-C. 
These results show that \emph{low sensitivity 
correlates with the improved robustness of transformers}. 

\item Lower Sensitivity Correlates with Flatter Minima: We explore the connection between sensitivity and a property of the loss landscape that has been found to correlate to good generalization --- the sharpness of the minima. We compare the sharpness of the minima with and without sensitivity regularization, and our results show that \emph{lower sensitivity correlates with flatter minima}. This indicates that sensitivity could serve as a unified notion for both robustness and generalization.

\item Sensitivity Serves as a Progress Measure for Grokking: \bvedit{We examine if sensitivity can be used to understand the training dynamics of transformers, specifically from the perspective of the phenomenon of \emph{grokking} where test accuracy abruptly improves long after the training loss or accuracy saturates. We consider modular addition, a task where transformers exhibit grokking. We show that \emph{sensitivity provides a progress measure that decreases even when the training loss does not reduce and is indicative of stages of grokking}. 
}

\end{tightenumerate}

%% file: sections/related-work.tex
\paragraph{Understanding Transformers.} The emergence of transformers as the go-to architecture for many tasks has inspired extensive work on understanding the internal mechanisms of transformers, including reverse-engineering language models \citep{wang2022interpretability}, the grokking phenomenon  \citep{power2022grokking,nanda2023progress}, manipulating attention maps \citep{hassid2022does,kobayashi2024analyzing}, automated circuit finding \citep{Conmy2023TowardsAC}, arithmetic computations \citep{hanna2023how,quirke2024understanding}, optimal token selection \citep{AtaeeTarzanagh2023TransformersAS,tarzanagh2023maxmargin,vasudeva2024implicit}, and in-context learning \citep{brown2020lm,Garg2022WhatCT,Akyrek2022WhatLA,Oswald2022TransformersLI,fu2023transformers,bhattamishra2023understanding,guo2024how}. Several works investigate why vision transformers (ViTs) outperform CNNs \citep{Trockman2022PatchesAA,raghu2021do,MelasKyriazi2021DoYE}, as well as other properties of ViTs, such as robustness to (adversarial) perturbations and distribution shifts \citep{bai2023transformers,shao2021adversarial,mahmood2021robustness,bhojanapalli2021understanding,Naseer2021IntriguingPO,paul2022vision,ghosal2022vision}. Further, several works on mechanistic interpretability of transformers share a similar recipe of measuring sensitivity --- corruption with Gaussian noise \citep{meng2022locating,Conmy2023TowardsAC} but on hidden states rather than the input space.

\paragraph{Sensitivity and Spectral Bias.} 
Sensitivity is closely related to spectral bias \citep{yang2020finegrained}, which is a bias towards `simple' functions in the Fourier space. 
Simple functions in the Fourier space generally correspond to low-frequency terms when the input space is continuous, and low-degree polynomials when the input space is discrete. Recent work has shown that deep networks prefer to use low-frequency Fourier functions on images \citep{xu2019frequency}, and low-degree Fourier terms on Boolean functions \citep{yang2020finegrained}. We note that in contrast to some other notions of spectral bias, sensitivity also has the advantage that it can be efficiently estimated on data through sampling --- in contrast, estimating all the Fourier coefficients  requires time exponential in the dimensionality of the data and hence can be computationally prohibitive \citep{xu2019frequency}.

\paragraph{Simplicity Bias in DL.} Several works \citep{Neyshabur2014InSO, valle2018deep,arpit2017closer,geirhos2020shortcut} show that NNs prefer learning `simple' functions over the data. \citet{inc-comp} show that during the early stages of SGD training, the predictions of NNs can be approximated well by linear models. \citet{morwani2023simplicity} show that 1-hidden-layer NNs exhibit simplicity bias to rely on low-dimensional projections of the data, while \citep{Huh2021TheLS} empirically show that deep NNs find solutions with lower rank embeddings. \citep{shah2020pitfalls} create synthetic datasets where features that can be separated by predictors with fewer piece-wise linear components are considered simpler, and show that in the presence of simple and complex features with equal predictive power, NNs rely heavily on simple features. \citet{geirhos2018imagenettrained} show that trained CNNs rely more on image textures rather than image shapes to make predictions. \citet{rahaman19a} use Fourier analysis tools and show that deep networks are biased towards learning low-frequency functions, and \citep{xu2019frequency,cao2021towards,bietti-ntk-2019,basri2019convergence} provide further theoretical and empirical evidence for this.

%% file: sections/spectral-bias.tex
\section{Sensitivity and Weak Spectral Bias}

\label{sec:th-spec}
In this section, we theoretically show that transformers with linear attention exhibit (weak) spectral bias to learn lower-order Fourier coefficients, which in turn implies a bias to learn low-sensitivity functions. We start with an overview of  Fourier analysis on the Boolean cube and sensitivity.

\paragraph{Fourier analysis on the Boolean cube \citep{o'donnell_2014}.} The space of real-valued functions on the Boolean cube $\cube{d}$ forms a $2^d$-dimensional space. Any such function can be written as a \emph{unique multilinear} polynomial. 
Specifically, the multilinear monomial functions, $\chi_U(\x)\!:=\!x^U\!:=\!\prod_{i\in U}x_i$, for each $U\subseteq [d]$, form a Fourier basis of the function space $\{f:\cube{d}\rightarrow \R\}$, \textit{i.e.}, their inner products satisfy $\E_{\x\sim\cube{d}}\left[\chi_U(\x)\chi_V(\x)\right]\!=\!\ind[U\!=\!V]$. Consequently, any function $f : \cube{d} \rightarrow \R$ can be written as $f(\x)\!=\!\textstyle\sum\limits_{U\subseteq [d]}\hat{f}(U)\chi_U(\x)$, for a unique set of coefficients $\hat{f}(U),U\subseteq[d]$, where $[d]\!=\!\{1,\dots,d\}$. 

\paragraph{Sensitivity in Boolean function analysis.} Sensitivity is a common complexity measure for Boolean functions. Intuitively, it captures the changes in the output of the function, averaged over the neighbours of a particular input. Formally, let $\,\cube{d}:= \{\pm1\}^d$ denote the Boolean cube in dimension $d$. The sensitivity of a Boolean function $f:\cube{d}\rightarrow\{\pm1\}$ at input $\x\in\cube{d}$ is given by
$S(f,\x) =\textstyle\sum_{i=1}^d\ind[f(\x)\neq f(\x^{\oplus i})]$,
where $\ind[\cdot]$ denotes the indicator function and $\x^{\oplus i}= (x_1,\dots , x_{i-1},-x_i, x_{i+1}, \dots, x_d)$ denotes the sequence obtained after flipping the $i^{\text{th}}$ co-ordinate of $\x$. Note that in the Boolean case, the neighbor of an input can be obtained by flipping a bit, we will define a more general notion later which holds for more complex data. The average sensitivity is measured by averaging $S(f,\x)$ across all inputs, 
\begin{align}\label{eq:sens-bool}
S(f)=\mathop{\E}_{\x\sim\cube{d}}[S(f,\x)]= \tfrac{1}{2^d}\sum\limits_{\x\in\cube{d}}S(f,\x).
\end{align}
Following \citet{bhattamishra2023simplicity}, when comparing inputs of different lengths, we consider the average sensitivity normalized by the input length, $\overline{S}(f)=\tfrac{1}{d}S(f)$. The sensitivity of a function $f$ is known to be related to the degree $D(f)$ of the multilinear polynomial which represents $f$ \citep{huang2019induced,hatami2011variations}, and low-degree functions have lower sensitivity. Specifically, a  breakthrough result \citep{huang2019induced} showed that $D(f)\leq S_{\max}^2(f)$ , where $S_{\max}(f):=\max_{\x\in\cube{d}}S(f,\x)$.

\paragraph{Attention Layer.} 
The output of a single-head self-attention layer, parameterized by key, query, value matrices $\WQ , \WK \!\in\! \R^{\db\times d_h},\WV \!\in\! \R^{\db\times d_v}$ for input $\X\!\in\!\R^{\Tau\times \db}$ with $\Tau$ tokens of $\db$ dimension, is 
$\attn(\X;\WQ,\!\WK,\!\WV):=\sft{\X\WQ\WK^\top\X^\top}\X\WV$,
where $\sft{\X\WQ\WK^\top\X^\top}\!\in\! \R^{\Tau\times \Tau}$ is the attention map with the softmax map $\sft{\cdot}: \R^\Tau\!\rightarrow\!\R^\Tau$ applied row-wise.

\paragraph{Main Results.} Consider any model with at least one self-attention layer, 
where $\X$ is obtained by reshaping $\x\in\cube{d}$, $d=\Tau \db$. Instead of applying the softmax activation, we consider linear attention and apply an identity activation element-wise with a scaling factor of $d^{-1/2}$. The following result shows that the conjugate kernel (CK) or neural tangent kernel (NTK) (see \cref{app:proof} for an overview) induced by transformers with linear attention exhibit a weak form of spectral bias, where the eigenvalues do not decrease as 
the degree of the multi-linear monomials increases, {separately} for even and odd degrees; see \cref{app:proof} for the proof. 
\begin{proposition}
\label{corr:wsb} Let $K$ be the CK or NTK of a transformer with linear attention on a Boolean cube $\cube{d}$. For any $\x,\y\in\cube{d}$, we can write $K(\x,\y)=\Psi(\inpb{\x}{\y})$ for some univariate function $\Psi:\R\rightarrow \R$. Further, for every $U\subseteq [d]$, $\chi_U$ is an eigenfunction of $K$ with eigenvalue
    \begin{align*}
        \mu_{|U|}:= \mathop{\E}_{\x\sim\cube{d}}\left[x^UK(\x,\mathbf{1})\right]= \mathop{\E}_{\x\sim\cube{d}}\left[x^U \Psi\left(d^{-1}\textstyle\sum\limits_ix_i\right)\right],
    \end{align*}
    where $\mathbf{1} := (1,\dots,1) \in \cube{d}$, and
    the eigenvalues $\mu_k$, $k \in [d]$, satisfy
    \begin{align*}
        \mu_0\geq  \mu_2\geq \dots\geq \mu_{2k} \geq \dots ,\quad
\mu_1\geq  \mu_3\geq  \dots\geq\mu_{2k+1} \geq \dots .
    \end{align*}
\end{proposition}
Note that for a given $U$, the eigenvalue $\mu_{|U|}$ only depends on $x^U$ and $\textstyle\sum_ix_i$ by definition, and hence, it is invariant under any permutation of $[d]$. 
Larger eigenvalues for lower-order monomials indicate that simpler features are learned faster. Since low sensitivity implies learning low-degree polynomials, \cref{corr:wsb} also implies a weak form of low sensitivity bias. 

We now show a connection between lower sensitivity and better robustness. Given a sample $\xb\in\cube{d}$ and some $\rho\in(0,1)$, consider a noisy sample $\xb'$, where $x'_i=x_i$ with probability $\rho$ and uniformly random, otherwise. It can be verified that the pair $(\xb,\xb')$ has correlation $\rho$. 
The following result shows that if $f$ has a lower  sensitivity $S_{\max}(f)$, then there is a lower probability $\prob [f(\xb)\!\neq\! f(\xb')]$ of inconsistent predictions on the pair $(\xb,\xb')_\rho$; see \cref{app:proof} for the proof.


\begin{proposition} \label{prop:ns}Given $\rho$-correlated pair $(\xb,\xb')$, where $\rho\!\in\!(0,1)$, and function $f\!:\!\cube{d}\!\rightarrow\!\{\pm 1\}$ with maximum sensitivity $S_{\max}(f)$, $0\leq \!\underset{(\xb,\xb')_\rho}{\prob}\![f(\xb)\!\neq\! f(\xb')] \leq 0.5(1\!-\!\rho^{(S_{\max}(f))^2})$. 
\end{proposition}

Together, \cref{corr:wsb,prop:ns} imply that transformers have low sensitivity and hence, better robustness. In \cref{sec:implications}, we present experimental evidence showing that the low sensitivity of transformers correlates with their improved robustness.

%% file: sections/beyond-boolean.tex
\section{Measuring Sensitivity beyond Boolean Data}\label{sec:sensitivity_def}

While sensitivity appears to be a promising metric to understand the inductive biases of transformers, it is only defined for Boolean data. In order to investigate the inductive biases in real-world image and language tasks, we need an equivalent metric for high-dimensional, real-valued data. 

We define the sensitivity metric for high-dimensional data, which is an analog of \cref{eq:sens-bool} as follows.
\begin{definition}\label{def:sensitivity} Given a model $\f$, dataset $\D$ and distribution $\Pc$, sensitivity is computed as:
\begin{equation}
\overline{S}(\f)=\frac{1}{\Tau}~\underset{\substack{\X\sim\D\\\xb\sim\Pc}}{\E} \left[\sum\limits_{\tau=1}^\Tau\ind[\sign(\f(\Tb;\X))\neq \sign(\f(\Tb;\X^{\oplus \tau}))]\right],
\end{equation}
where $\X^{\oplus \tau}$ is obtained by replacing the $\tau^{\text{th}}$ token in $\X$ with $\xb$.
\end{definition}

\begin{figure*}
    \centering
    \includegraphics[width=0.6\linewidth]{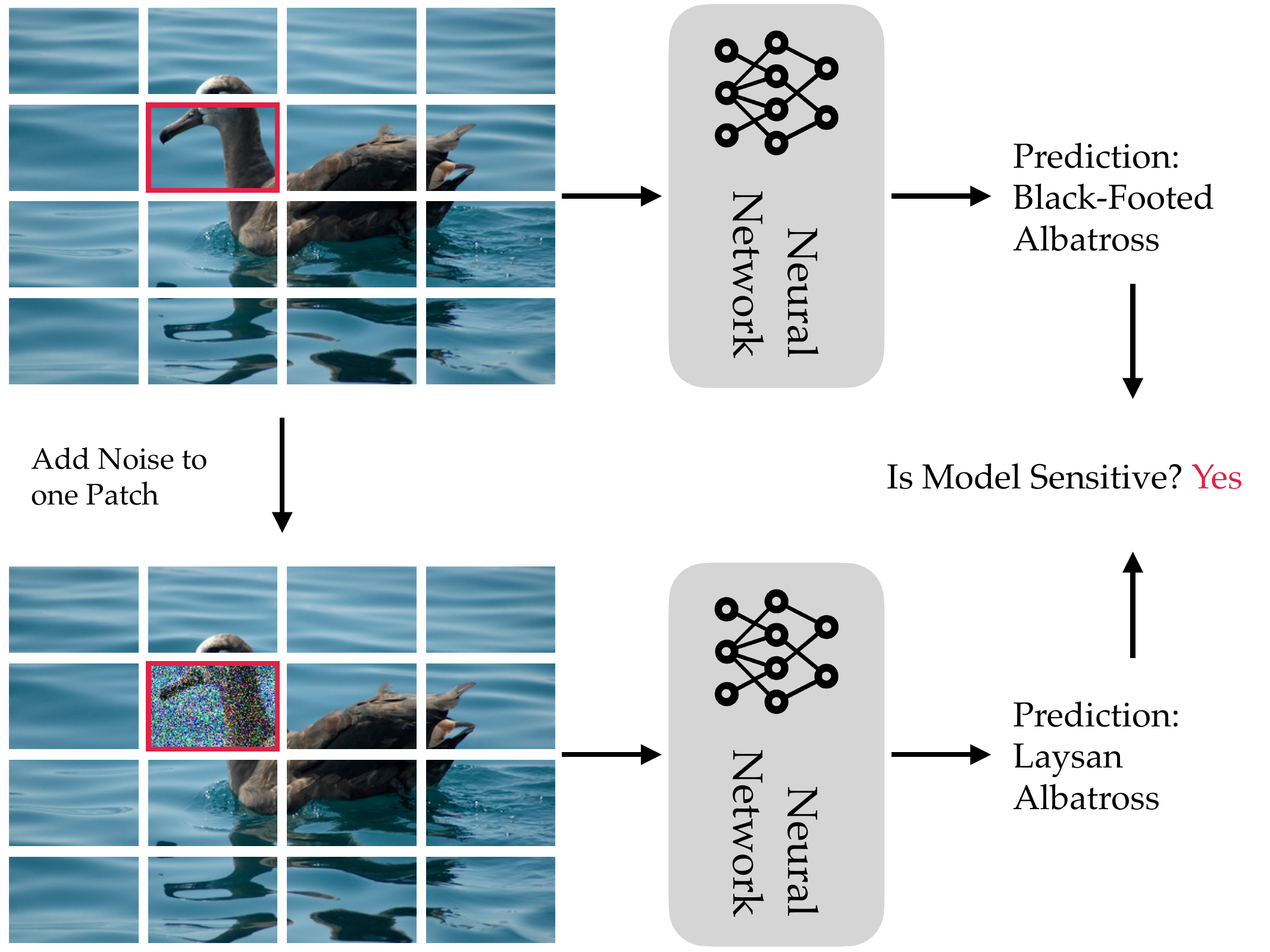}
    \captionsetup{margin=1mm}
    \caption{\textbf{Measuring Sensitivity in Vision Tasks}. A {\color{red} patch} is first selected to add Gaussian noise corruptions. Then the original image and the corrupted image are fed into the \textit{same} {\color{gray} neural network} to make predictions. If the predictions are inconsistent, then the {\color{gray} neural network} is sensitive to this {\color{red} patch}. The process is repeated for every patch to measure the overall sensitivity.}
    \label{fig:illustration-sensitivity}
\end{figure*}

An important consideration here is to define $\Pc$.
While one can replace a token with a randomly selected token to measure sensitivity, this may not ensure that the new token lies in a neighbourhood of the original token. Capturing how the output changes with local perturbations according to the metric of the underlying space is an important aspect of the sensitivity definition for Boolean functions (as discussed in \citet{gopalan2016smooth}, for e.g.), and appropriate extensions of sensitivity beyond Boolean functions should capture this property. For input spaces such as natural images or text embeddings, there is more structure in the tokens, and a randomly selected 
token can lie far from the original token's neighborhood. Therefore, instead of replacing a token with a random token, we inject small perturbations into the token to evaluate sensitivity. This allows us to control the size of the neighbourhood by selecting the strength of the noise perturbation.  
 
Formally, for each token $\vct{e}_\tau$ of an input  $\X$, let $\vct{x} \!:=\! \vct{e}_\tau\! +\! \vct{\xi}$ be a perturbed token, where $\vct{\xi} \!\sim\! \mathcal N(\vct{0}, \sigma^2 \mat{I})$ is an isotropic Gaussian with variance $\sigma^2$. We measure sensitivity by replacing $\vct{e}_\tau$ with $\vct{x}$ as per \cref{def:sensitivity}, with $\Pc$ as $\Nc(\vct{0}, \sigma^2 \mat{I})$. For image data, each token $\vct{e}_\tau$ corresponds to different patches
(see \cref{sec:expts-vision} for further details), while for language data, the tokens correspond to embeddings of sub-words (see \cref{sec:expts-language} for more details). \Cref{fig:illustration-sensitivity} illustrates the measurement for images.

 The important characteristics of this metric are that it is a unified notion of complexity across vision and language tasks, and as we will see later, it distinguishes transformers from various other architectures. For instance, we compare the sensitivity measured with token-wise perturbations as mentioned above, with random perturbations across the input in \cref{app:add-expts} and find that the gap in the latter metric is not as large as the proposed metric. 

 We also note that while for Boolean data, sensitivity aligns with other notions of complexity, such as sparsity, this may or may not be the case in settings with high-dimensional or real-valued data. In the following section, we present experimental results for a self-attention model, in a synthetic data setting to demonstrate this. Specifically, we show that in the synthetic dataset, the related notion of using a sparse set of input tokens may or may not align with low-sensitivity, but the model learns the low-sensitivity function in both cases.

%% file: sections/toy-expts.tex
\subsection{Experiments on Synthetic Data}

\label{sec:synth}
We construct a synthetic dataset to examine the inductive bias of a single-layer self-attention model. 
We show that in the presence of two solutions with the same predictive power but different sensitivity values, this model learns the low-sensitivity function. We begin by describing the experimental setup and then discuss our results. 

\paragraph{Setup.}  We compose a single-head self-attention layer 
with a linear head $\U\in\R^{\db\times \db}$ to obtain the final prediction, and write the full model as
\begin{align}\label{eq:attn-model}
\f(\Tb;\X):=\inpb{\U}{\sft{\X\WQ\WK^\top\X^\top}\X\WV}, 
\end{align}
where $\Tb:=\concat(\WQ,\WK,\WV,\U)$. We consider this model for the experiments in this section, with all the parameters initialized randomly at a small scale. Next, we describe the process to generate the dataset. We first define the vocabulary as follows:
 
\input{sections/figure_synth2}

\begin{definition}[Synthetic Vocabulary]\label{def:syn_voc}
    Consider a vocabulary of $M$ distinct tokens $\mathcal{V}\!:=\!\{\e_1,\dots\e_M\}$, where $\e_i\!\in\!\{0,1\}^d$ denotes the $i^{\text{th}}$ basis vector. We define smaller subsets of \textit{sparse} tokens and larger subsets of \textit{frequent} tokens for each label $y=\pm 1$, as well as a subset of \textit{irrelevant} tokens:  
\begin{align*}
    &\Vsp^+:=\{\e_1\}, \Vsp^-:=\{\e_2\}, \Vir:=\{\e_{2m+3},\dots,\e_M\}\\
    &\Vfr^+:=\{\e_3,\e_5,\dots,\e_{2m+1}\}, \Vfr^-:=\{\e_4,\e_6,\dots,\e_{2m+2}\}.
\end{align*}
\end{definition}

%

Let $\Tau$ denote the sequence length of each data point,  $n_f$ and $n_s$ denote the number of frequent and sparse tokens, respectively, such that $n_s\!<\!n_f\!<\!\min(m,\Tau\!-\!n_s)$, and $n_d$ be a parameter satisfying $n_d\!\leq\! n_f$. Next, we describe the process of generating the dataset $\mathcal{D}$; see \cref{fig:vocab} for an example.
\begin{definition}[Dataset Generation]\label{def:gen_data}
Consider the vocabulary in \cref{def:syn_voc}. To generate a data point $(\X,y)$, we first sample the label $y\in\{\pm 1\}$ uniformly at random. We divide the indices $[\Tau]$ into three sets $\Ifr, \Isp $ and $\Iirr$, and sample each set as follows:
\begin{itemize}[leftmargin=*,itemsep=0em, topsep=0em]
    \item  $\Ifr$ is composed of  $\left\lfloor (n_f+n_d)/2 \right\rfloor$ tokens uniformly sampled from $\Vfr^y$ and $n_f-\left\lfloor (n_f+n_d)/2\right\rfloor$ tokens uniformly sampled from $ \Vfr^{-y}$.
    \item $\Isp$ contains $n_s$ tokens uniformly sampled from $\Vsp^y$.
    \item The remaining $\Tau-n_f-n_s$ tokens in $\Iirr$ are uniformly sampled from $\Vir$. 
\end{itemize}
\end{definition}

To determine if the tokens in $\Vsp$ or those in $\Vfr$ have a more significant impact on the model's predictions, we adapt the test set generation process by altering the second step in \cref{def:gen_data}: we sample the sparse tokens from $\Vsp^{-y}$ instead of $\Vsp^{y}$. If this modification leads to a noticeable drop in the test accuracy, it suggests that the model relies on the sparse feature(s) for its predictions.

We consider two other metrics to examine the role of the attention head and the linear predictor. Define three vectors: $\vs\!:=\!\e_1\!-\!\e_2,\,\vf\!:=\!\!\!\!\!\!\!\textstyle\sum\limits_{i\in\Vfr^+}\!\!\!\!\!\e_i\!-\!\!\!\!\!\!\!\textstyle\sum\limits_{i\in\Vfr^-}\!\!\!\!\!\e_i,\,   \vi\!:=\!\!\!\!\!\!\!\!\textstyle\sum\limits_{i\in\Vir}\!\!\!\!\!\e_i$. 
We plot the average alignment (cosine similarity) between the rows of $\U\WV^\top$ and these vectors to see what tokens the prediction head relies on. Similarly, we plot the sum of the softmax scores for the three types of tokens to see which tokens are selected by the attention mechanism. 

\noindent We set $\Pc$ in \cref{def:sensitivity} as the uniform distribution over $\V$ for computing sensitivity in our experiments.

\paragraph{Results.} \Cref{fig:synth-joint} shows the train and test dynamics of the model in \cref{eq:attn-model} using synthetic datasets generated by following the process in \cref{def:gen_data} (details in \cref{tab:sens-synth}). We consider two cases: in the first case (left column), using the sparse token leads to a function with lower sensitivity, whereas in the second case (right column), using the frequent tokens leads to lower sensitivity (see \cref{tab:sens-synth} for a comparison of the sensitivity values). We observe that in the first case, the OOD test accuracy drops to $0$, the alignment with $\vs$ is close to $1$ and the attention weights on the sparse tokens are the highest. These results show that the model relies on the sparse token in this case. On the other hand, in the second case, the test accuracy remains high, the alignment with $\vf$ is close to $1$ and the attention weights on the frequent tokens are the highest, which shows that the model relies on the frequent tokens. These results show that the model exhibits a low-sensitivity bias. Note that in both cases, the model can learn a function that relies on a sparse set of inputs (using the sparse tokens), however, it uses these tokens only when doing so leads to lower sensitivity.

%% file: sections/figure_synth2.tex
\begin{figure*}[t]
    \centering
\begin{minipage}[b]{0.53\linewidth}
\begin{minipage}[b]{\linewidth}
            \centering
            \includegraphics[width=\linewidth]{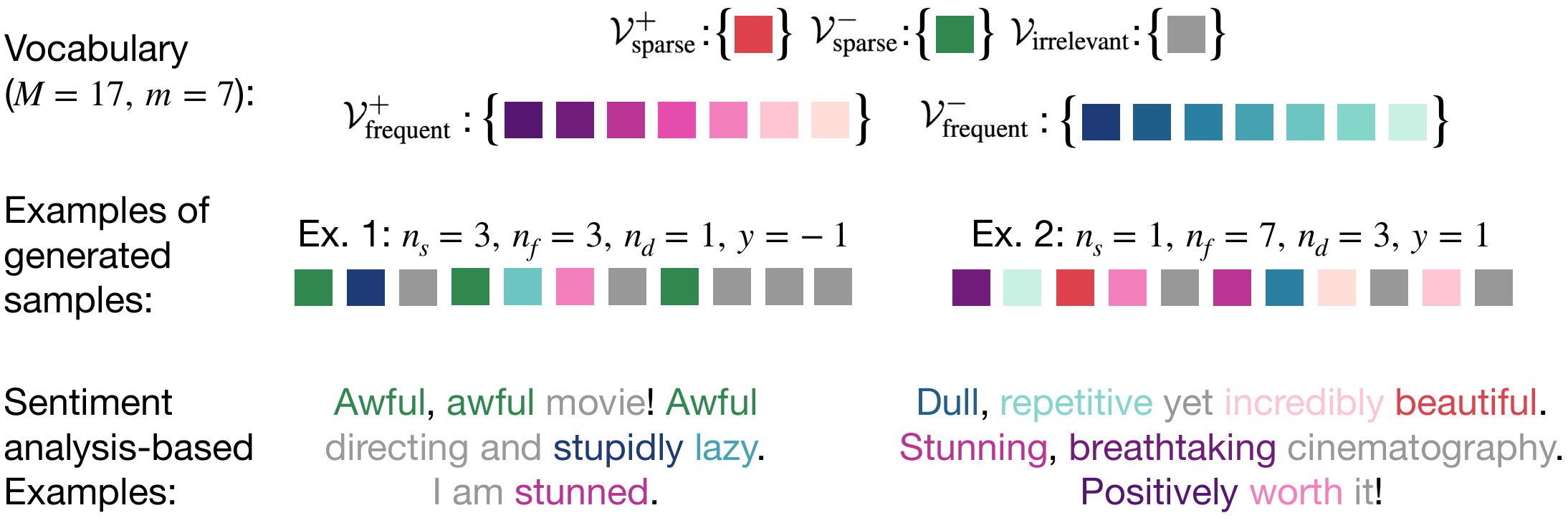}
    \caption{Visualization of the synthetic data generation process (see \cref{sec:synth} for details). For simplicity, we represent each $d$-dimensional token with a square. Middle row: In each case, given a label $y$, we randomly sample $\Tau=11$ tokens, with $n_s$ tokens from $\Vsp^y$, $\left\lfloor (n_f+n_d)/2 \right\rfloor$ tokens from $\Vfr^y$, $n_f\!-\!\left\lfloor (n_f+n_d)/2\right\rfloor$ tokens from $ \Vfr^{-y}$ and the remaining tokens from $\Vir$. Note that in the first example, since $n_s\!=\!3$ and $n_d\!=\!1$, a predictor that relies (only) on the sparse tokens is less sensitive compared to the one that relies on the frequent tokens. On the other hand, in the second example, since $n_s\!=\!1$ and $n_d\!=\!3$, the predictor that relies on the frequent tokens is less sensitive. Bottom row: We include two sentiment analysis-based examples to illustrate the synthetic data \bvedit{samples in the second row, using the same colors as the first two rows.}}
    \label{fig:vocab}
    
    \end{minipage}
\vfill
\begin{minipage}[b]{\linewidth}
\centering
\vspace{3mm}
\captionof{table}{Comparison of sensitivity values for models that use only sparse or frequent tokens for the settings considered in \Cref{fig:synth-joint}.}
\resizebox{\linewidth}{!}{%
\begin{tabular}{lccc}    
\toprule
$(n_s,n_f,n_d,m)$ & Using sparse tokens & Using frequent tokens \\
\midrule 
$(3, 5, 1, 16)$; \cref{fig:synth-joint} left col.    & $0$ & $0.2878$\\
$(1, 17, 7, 20)$; \cref{fig:synth-joint} right col.   & $0.0339$ & $0$\\
\bottomrule
\end{tabular}}

    \label{tab:sens-synth}
    
\end{minipage}   
\end{minipage}
\hfill    
\begin{minipage}[b]{0.44\linewidth}
\centering
\includegraphics[width=\linewidth]{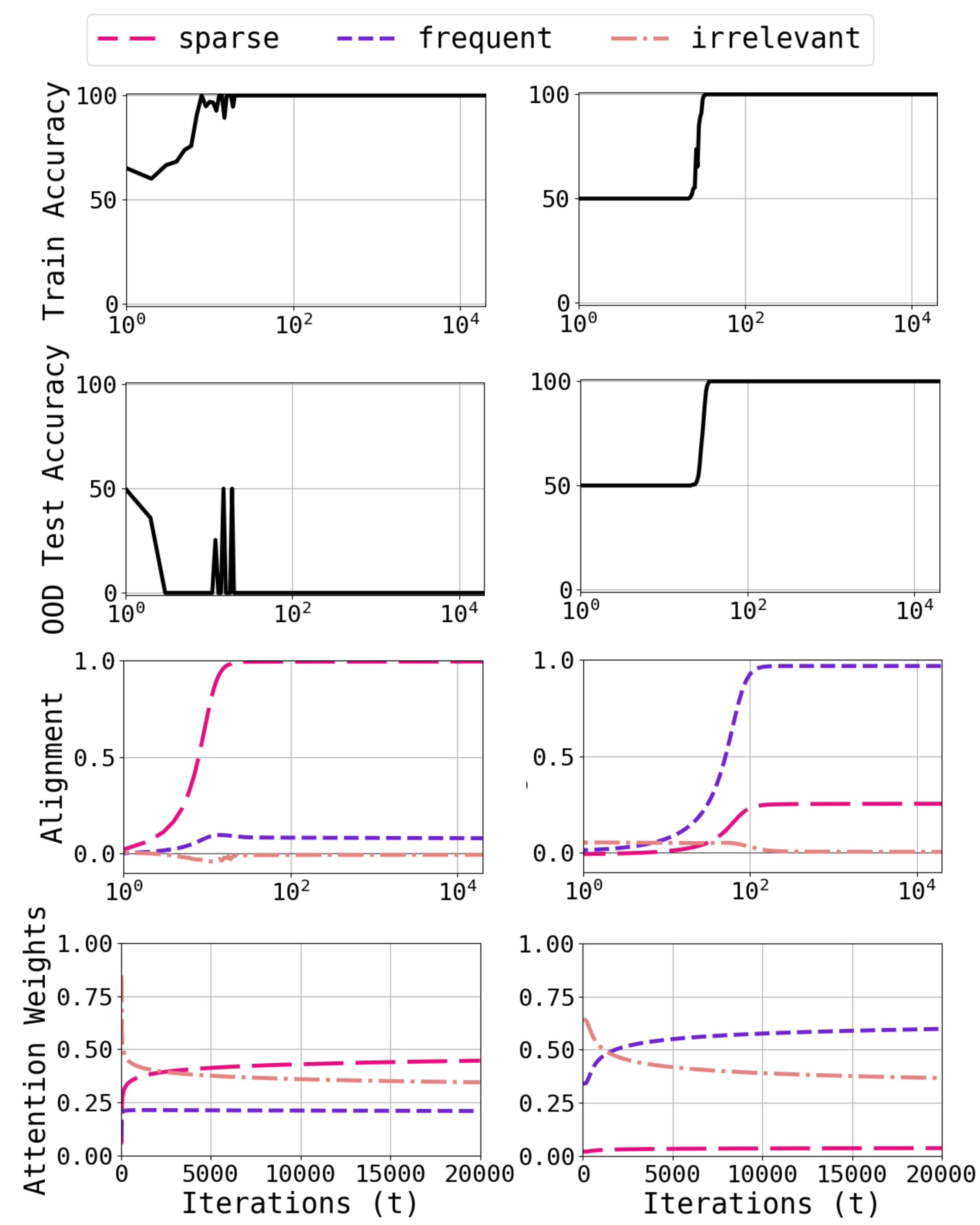}
    \caption{Train and test dynamics for a single-layer self-attention model (\cref{eq:attn-model}) using the synthetic data visualized in \cref{fig:vocab}; see \cref{sec:synth} for details. \textbf{Left column}: the predictor that uses \emph{sparse} tokens has lower sensitivity (Ex. 1 in \Cref{fig:vocab}), \textbf{Right column}: the predictor that uses \emph{frequent} tokens has lower sensitivity (Ex. 2 in \Cref{fig:vocab}); see \cref{app:add-expts-synth} for more examples.
    }
    \label{fig:synth-joint}
\end{minipage}
\end{figure*}

%% file: sections/vision-expts.tex
\section{Investigations on Vision Tasks} \label{sec:expts-vision}

In this section, we test whether our notion of sensitivity captures the inductive bias of transformers on vision tasks. We consider Vision Transformers (ViT, \citealp{dosovitskiy2021an}) which 
regard images as a sequence of patches instead of a tensor of pixels.  
\begin{definition}[Tokenization for Vision Transformers] \label{def:vit_token}
   Let $\mat{X} \in \mathbb R^{n_h \times n_w\times n_c}$ be the image with height $n_h$, width $n_w$, and number of channels $n_c$. A tokenization of $\mat{X}$ is a sequence of $T$ image patches $\{\vct{e}_1, \cdots, \vct{e}_T\}$ where each token $\vct{e}_i$ represents an image patch of dimension $d=n_w n_h n_c / T$. 
\end{definition}

Sensitivity is measured on the \textit{training} set because our goal is to understand the simplicity bias of the model at training time, to see if it prefers to learn certain simple classes of functions on the training data. Since different models could have different generalization capabilities, the sensitivity on test data might not reflect the model's preference for low-sensitivity functions at training time. Further, since the choice of optimization algorithm could in principle introduce its own bias and our goal is to understand the bias of the architecture, we train both the models with the same optimization algorithm, namely SGD; see \cref{fig:sens-opt} in the App. for a comparison with Adam.

\begin{figure}[t]
    \centering
    \includegraphics[width=0.8\linewidth]{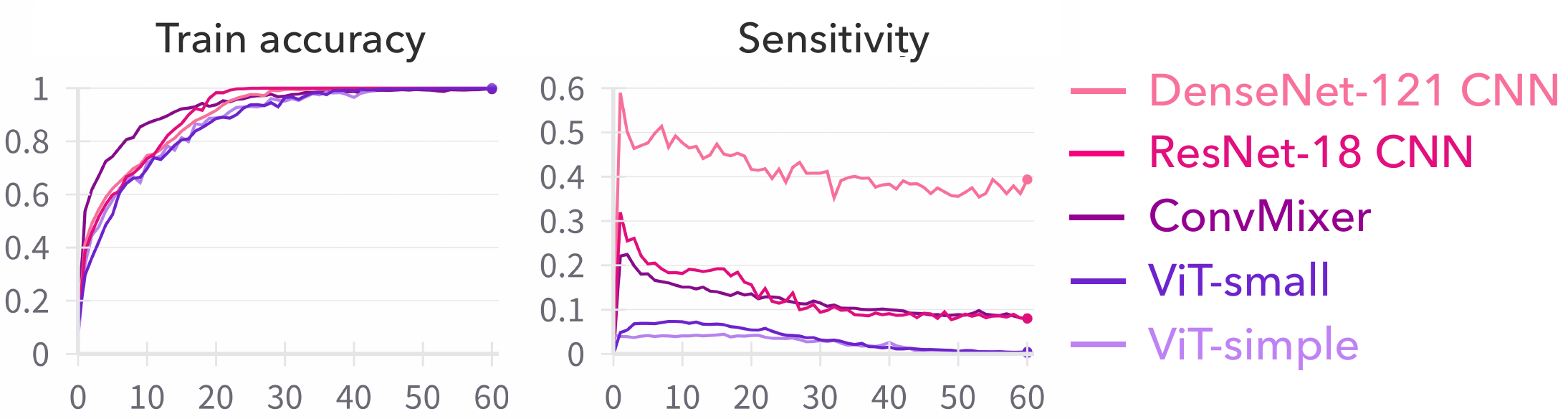}
    \caption{\textbf{Sensitivity on CIFAR-10.} Comparison of the sensitivity of two CNNs, two ViTs, and ConvMixer trained on the CIFAR-10 dataset, as a function of training epochs. For a fair comparison, the figure also shows the train accuracies (see App. \cref{fig:cifar-full} for full train dynamics). All models have similar accuracies but the ViTs have significantly lower sensitivity. 
    }
    \label{fig:cifar}
\end{figure}

We consider three datasets in this section (see \cref{app:expt-details} for details), namely CIFAR-10 \citep{cifar}, ImageNet-1k \citep{imagenet15russakovsky} and Fashion-MNIST \citep{fm}. We use $\sigma^2\!=\!1$, $15$ and $5$, respectively. We use a variant of the ViT architecture for small-scale datasets proposed in \citet{lee2021vision}, referred to as ViT-small here onwards; see \cref{app:expt-details} for more details and \cref{app:add-expts} for additional results where we show that varying model depths and number of heads does not affect the sensitivity of ViT models. We also compare the sensitivity of the ViT-small model and a ResNet-18 \citep{he2016residual} CNN on the SVHN dataset with $\sigma^2\!=\!1$ in \cref{app:add-expts}, which leads to the same conclusion. 

\paragraph{Transformers learn lower sensitivity functions than CNNs.} \Cref{fig:cifar} shows the train accuracies as well as the sensitivity comparison between two ViTs: the ViT-small model and a ViT-simple model \citep{beyer2022better}, two CNNs: a ResNet-18 and a DenseNet-121 \citep{Huang2016DenselyCC}, and a ConvMixer model \citep{Trockman2022PatchesAA}. Note that the train accuracies are comparable for all architectures, which allows for a fair comparison of sensitivity. We observe that the ViTs have significantly lower sensitivity compared to the CNNs and the ConvMixer model. At the end of the training, the sensitivity values are $0.3673$ for DenseNet-121, $0.0827$ for ResNet-18, $0.0829$ for ConvMixer, $0.0050$ for ViT-small and $0.0014$ for ViT-simple. 

\textbf{Do Transformers have lower sensitivity than CNNs because these models process inputs differently?} ViTs process inputs as a sequence of patches whereas CNNs do not, and hence a natural question to ask is if the difference in sensitivity between the two architectures is due to this difference in processing the inputs as opposed to differences in the architecture.  To investigate this, we compare ViTs with ConvMixer \citep{trockman2022patches}. 
Similar to ViTs, ConvMixer processes the input data in a patch-wise manner, but has two key differences: it does not use the self-attention mechanism, which is the core component of transformers, and it relies on convolutions for the feedforward part as well. The higher sensitivity of the ConvMixer model indicates that the low sensitivity simplicity bias of the transformers is not because they process inputs patch-wise, but rather a result of other components of the architecture. 

\textbf{Do these observations generalize to pre-trained models?} To study this, we consider the ImageNet-1k dataset \citep{imagenet15russakovsky}. We compare the sensitivity values of pre-trained ConvNext \citep{9879745} and ViT/L-16 \citep{dosovitskiy2021an} models. For comparable accuracies, ViT/L-16 has a sensitivity of $0.0191$, which is lower than that of ConvNext at $0.0342$. \dfedit{This shows that the observations on small-scale models studied in this section transfer to large-scale pretrained models.}


\paragraph{Transformers learn lower sensitivity functions than MLPs.} 
Next, we consider the Fashion-MNIST dataset and compare the sensitivity of ViT-small, a 3-hidden-layer CNN, an MLP with LeakyReLU activation and an MLP with sigmoid activation (see \cref{fig:fm} in the Appendix for the training curves). 
At the end of training, the sensitivity values are $0.0559$ for the MLP with LeakyReLU, $0.0505$ for the MLP with sigmoid, $0.0453$ for the CNN and $0.0098$ for the ViT. 

Thus, transformers learn lower sensitivity functions compared to MLPs{, ConvMixers,} and CNNs.

%% file: sections/nlp-expts.tex
\section{Investigations on Language Tasks}\label{sec:expts-language}

In this section, we 
investigate the sensitivity of transformers on natural language tasks, where each datapoint is a sequence of tokens. Similar to the comparison of ViTs with MLPs and CNNs in \Cref{sec:expts-vision}, we compare a RoBERTa \citep{liu2019roberta} transformer model with LSTMs \citep{hochreiter1997lstm}, an alternative auto-regressive model, in this section. Recall that we consider a transformer with linear attention for the results in \Cref{sec:th-spec}. Aligning with this setup, we also consider a RoBERTa model with ReLU activation in the attention layer (\textit{i.e.}, replacing $\vct{\varphi}(\cdot)$ in \cref{eq:attn-model} with $\mathrm{ReLU}(\cdot)$) for our experiments. 

We use the \bvreplace{same}{usual} RoBERTa-like tokenization procedure to process inputs for all the models so that they are represented as \texttt{<s>} $e_1, \cdots, e_T$ \texttt{</s>} where each $e_j$ represents tokens that are usually subwords and \texttt{<s>} represents the classification (CLS) token, $T$ the sequence length, and  \texttt{</s>} the separator token. We denote $e_0 \!= $ \texttt{<s>} and $e_{T+1} \!= $ \texttt{</s>}. For each token $e_j$, a token embedding $h_E(\cdot)\!:\! [M]\!\rightarrow\! \mathbb R^d$ is trained during the process, where $M$ denotes the vocabulary size. For transformers, we also train a separate positional encoder $h_P(\cdot)\!:\! [N]\!\rightarrow\! \mathbb R^d$, where $N$ denotes the maximum sequence length. We denote $\vct{e}_j^\text{\tiny{LSTM}} \!=\! h_E^\text{\tiny{LSTM}}(e_j)$ and $\vct{e}_j^\text{\tiny{RoBERTa}} \!=\! h_E^\text{\tiny{RoBERTa}}(e_j) + h_P^\text{\tiny{RoBERTa}}(j)$ as the embedding tokens of LSTM and RoBERTa, respectively. We omit the superscript for convenience. 

To control the relative magnitude of noise, the embeddings $\vct{e}_\tau\! \leftarrow\! \texttt{LayerNorm}(\vct{e}_\tau)$ are first layer-normalized \citep{ba2016layer} before the additive Gaussian corruption. To better control possible confounders, we limit both LSTM and RoBERTa to having the same number of layers. Both models are trained from scratch, \textit{without} any pretraining on larger corpora, to ensure fair comparisons. 

We consider two binary classification datasets, MRPC \citep{dolan-brockett-2005-automatically} and QQP \citep{qqp-data} (see \cref{app:expt-details} for details), 
which are relatively easy to learn without pretraining \citep{kovaleva-etal-2019-revealing}. Empirically, we set $\sigma^2 \!=\! 15$ 
(results with $\sigma^2\! = \!4$ in App. \ref{app:add-nlp-expts} yield similar observations as the results in this section). Similar to \cref{sec:expts-vision}, we measure sensitivity on the train set (results on the validation set in App. \ref{app:add-nlp-expts} yield similar observations). We include results with different depth values for RoBERTa as well as using GPT-2 in App. \ref{app:add-nlp-expts} and they lead to similar conclusions.

\input{sections/figure_nlp}

\paragraph{Transformers learn lower sensitivity functions than LSTMs.} As shown in \Cref{fig:nlp-sensitivity}, both RoBERTa models have lower sensitivity than LSTMs on both datasets, regardless of the number of datapoints trained. Even at initialization with random weights, LSTMs are more sensitive. At the end of training, the sensitivity values on the MRPC dataset are $0.15$, $0.002$ and $0.001$ for the LSTM, the RoBERTa model with softmax activation and the RoBERTa with ReLU activation, respectively. On the QQP dataset, LSTM, RoBERTa-softmax and RoBERTa-ReLU have sensitivity values of $0.09$, $0.03$ and $0.02$, respectively. Interestingly, RoBERTa with ReLU activation also has lower sensitivity than its softmax counterpart. This may be because softmax attention encourages sparsity because of which the model can be more sensitive to a particular token; see 
Ex. 2 in \cref{fig:vocab} and bottom row of \cref{fig:synth-joint} for an example where sparsity can lead to higher sensitivity.

\paragraph{LSTMs are more sensitive to later tokens.} In \cref{fig:nlp-sensitivity-position}, we plot sensitivity over the token positions. We observe that LSTMs exhibit larger sensitivity towards the end of the sequence, i.e. at later token positions. In contrast, transformers are relatively uniform. Similar observations were made by \citep{fu2023transformers} for a linear regression setting: LSTMs do more local updates and only remember the most recent observations, whereas transformers preserve global information and have longer memory.

\paragraph{Transformers are sensitive to the CLS token.} In \cref{fig:nlp-sensitivity-position}, we also observe that the RoBERTa model with softmax activation has frequent bumps in the sensitivity values at early token positions. This is because different sequences have different lengths and while computing sensitivity versus token positions, we align all the sequences to the right. These bumps at early token positions indeed correspond to the starting token after the tokenization procedure, the CLS token \texttt{<s>}. This aligns with the observation of \citet{jawahar-etal-2019-bert} that the CLS token gathers all global information. Perturbing the CLS token corrupts the aggregation and results in high sensitivity. We also observe that RoBERTa with ReLU activation seems less sensitive to the CLS token compared to its softmax counterpart. 

%% file: sections/figure_nlp.tex
\begin{figure*}[t]
\centering
    \begin{minipage}[b]{0.49\linewidth}
            \centering
            \subfigure[MRPC]{\includegraphics[width=0.49\linewidth]{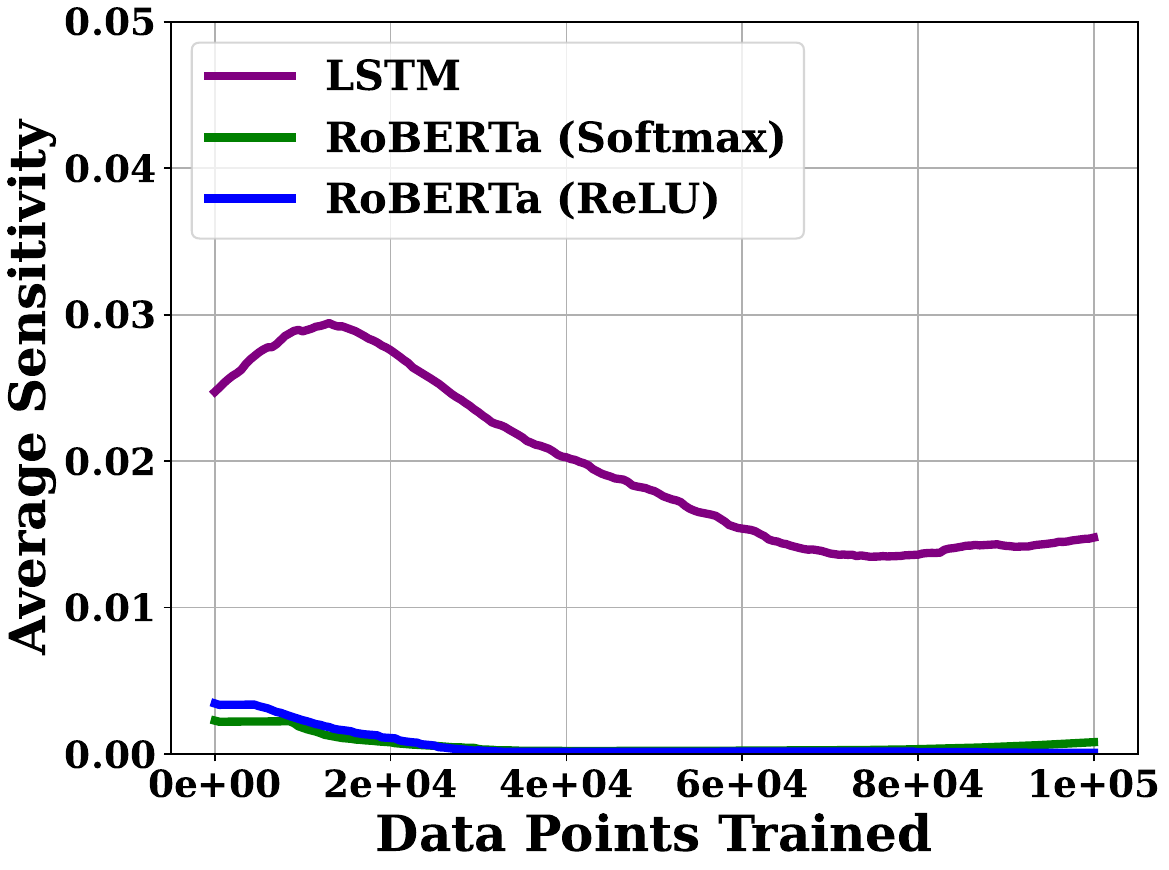}}\hspace{0.1cm}
            \subfigure[QQP]{\includegraphics[width=0.48\linewidth]{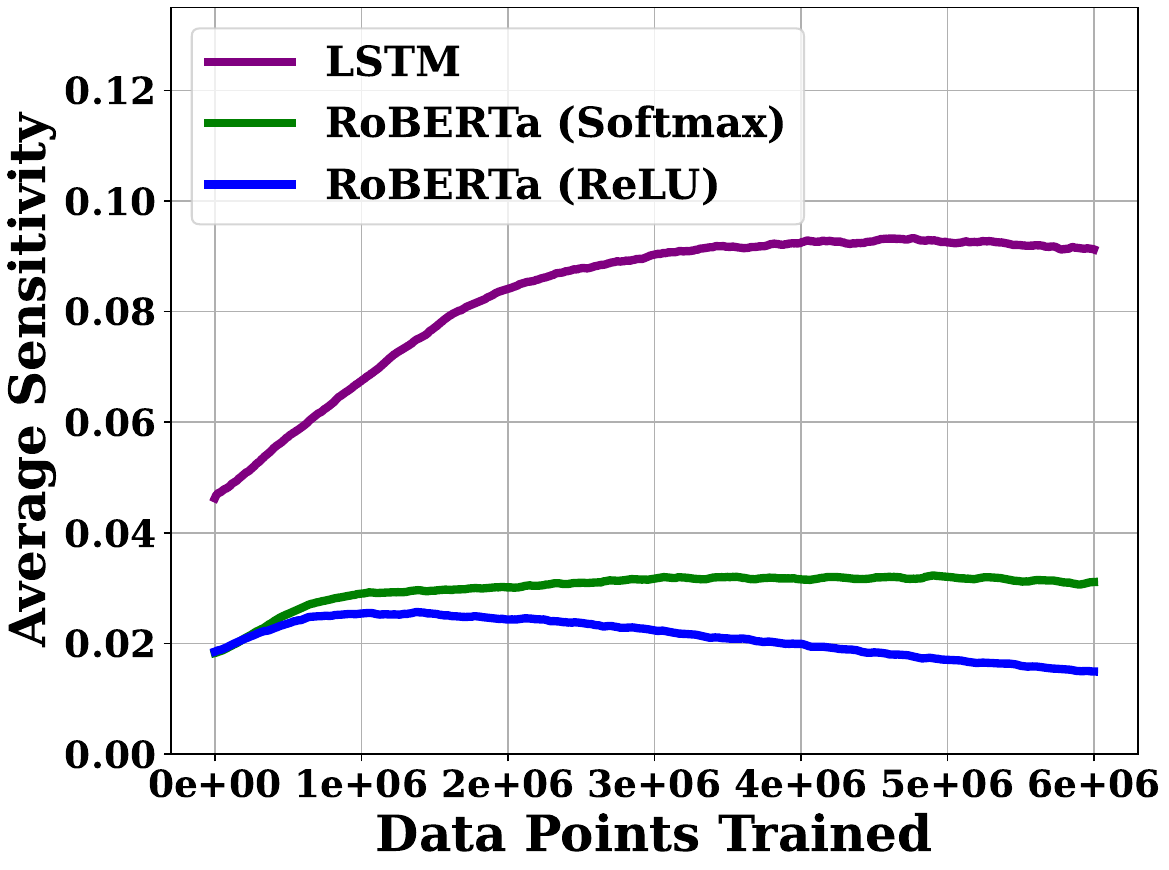}}
        
            \caption{\textbf{Sensitivity over Datapoints Trained}. On both datasets, the Transformer-based model RoBERTa displays much lower sensitivity compared to LSTMs during the entire training process. RoBERTa with ReLU activation has lower sensitivity compared to its Softmax counterpart at later stages of training.}
            \label{fig:nlp-sensitivity}
    \end{minipage}
    \hfill
    \begin{minipage}[b]{0.49\linewidth}
    \centering
    \subfigure[MRPC]{\includegraphics[width=0.48\linewidth]{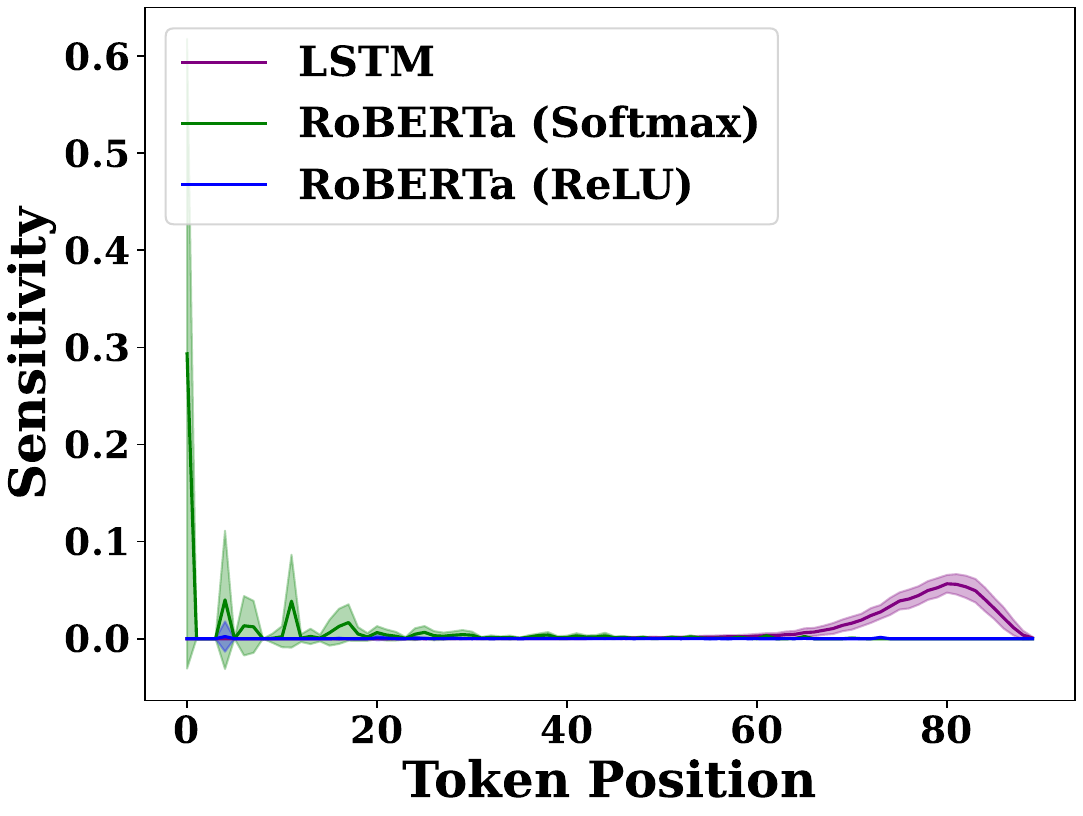}}\hspace{0.1cm}
    \subfigure[QQP]{\includegraphics[width=0.49\linewidth]{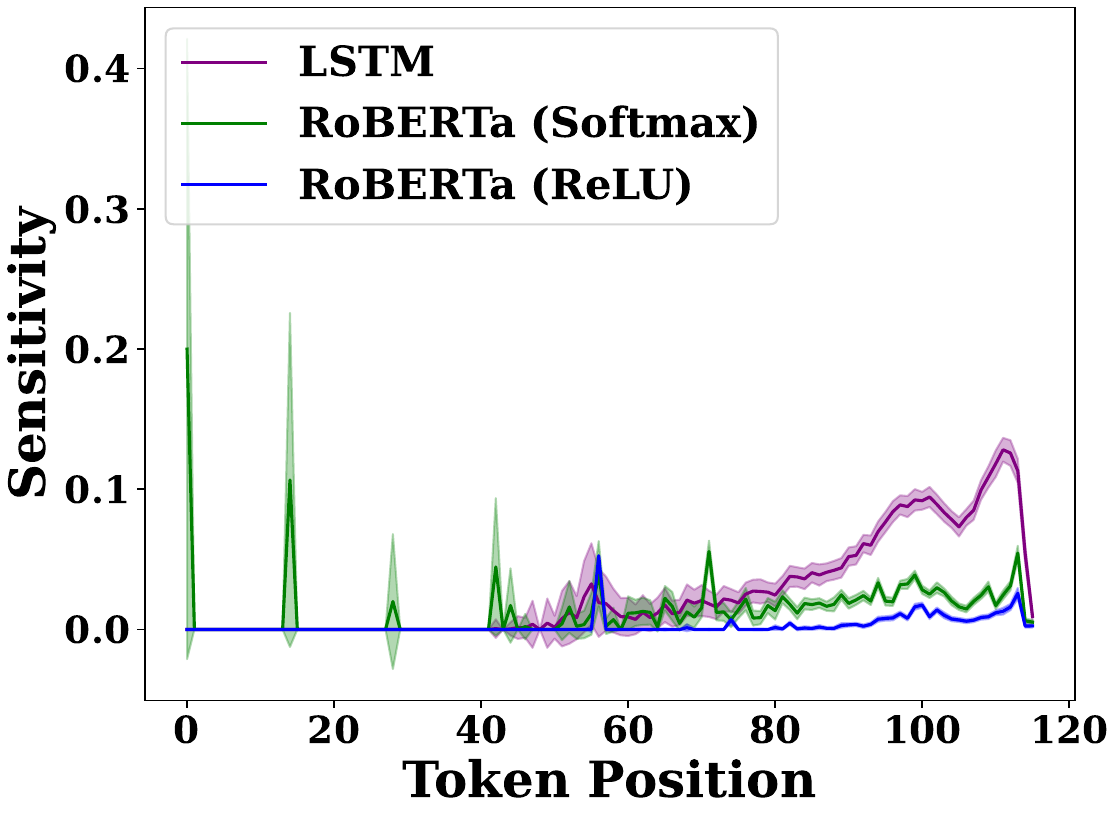}}
    \caption{\textbf{Sensitivity over Token Position}. On both datasets, LSTMs are more sensitive to later tokens than early ones, while RoBERTa's sensitivity, regardless of the activation function, is more uniform across token positions, except for a few early bumps in early tokens which come from the CLS token \texttt{<s>}. }
    \label{fig:nlp-sensitivity-position}
    \end{minipage}
    \
    \
\end{figure*}

%% file: sections/implications.tex
\section{Implications of Low Sensitivity Bias}

\label{sec:implications}
We saw in \cref{sec:expts-vision} that transformers learn lower sensitivity functions than CNNs. In this section, we first compare the test performance of these models on the CIFAR-10-C dataset and show that transformers are more robust than CNNs. Next, we add a regularization term while training the transformer, to encourage lower sensitivity. The results demonstrate that lower sensitivity leads to improved robustness. We then explore the connection between sensitivity and the flatness of the minima. Our results show that lower sensitivity leads to flatter minima.  Finally, we examine if sensitivity can be used to understand the training dynamics of transformers, where we find sensitivity to be a suitable progress measure for certain grokking instances.

%
\subsection{Lower Sensitivity leads to Improved Robustness} \label{ssec:robustness}

The CIFAR-10-C dataset \citep{hendrycks2018benchmarking} was developed to benchmark the performance of various NNs on object recognition tasks under common corruptions that are not confusing to humans. Images from the test set of CIFAR-10 are corrupted with $14$ types of algorithmically generated corruptions from blur, noise, weather, and digital
categories (see Fig. 1 in \citet{hendrycks2018benchmarking} for examples). 

\begin{figure}[t]
\centering
\hspace{5mm}
\includegraphics[width=\linewidth]{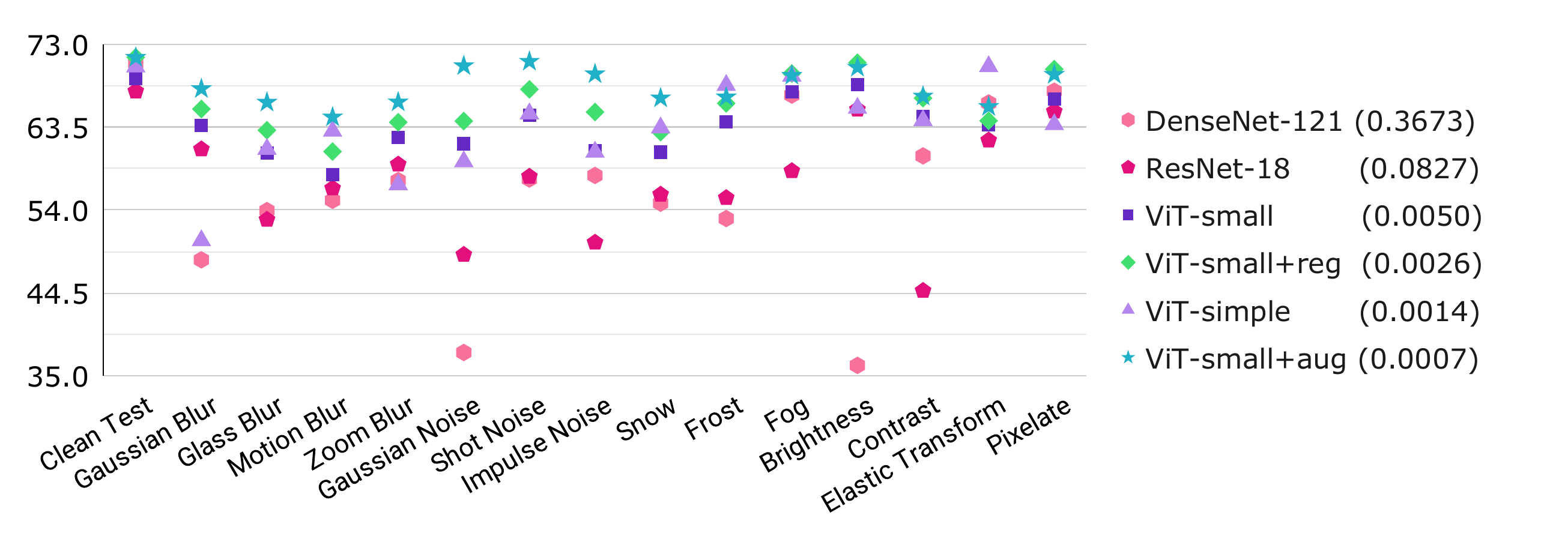}
    \caption{Comparison of the test accuracies on CIFAR-10 and on various corruptions from the CIFAR-10-C dataset (see \cref{sec:implications} for details) of various models trained on the CIFAR-10 dataset, at the last training epoch \bvedit{(see App. \cref{fig:cif10c} for a comparison of the accuracies as a function of training epochs.). We observe that the Vision Transformer models \texttt{ViT-small} and \texttt{ViT-simple} exhibit lower sensitivity and higher robustness to corruptions compared to the CNN models \texttt{DenseNet-121} and \texttt{ResNet-18}. Additionally, encouraging lower sensitivity while training through regularization (\texttt{ViT-small-reg}) and data augmentation (\texttt{ViT-small-aug}) leads to improved robustness (see \cref{sec:implications} for details).} } 
    \label{fig:sens-plot}
\end{figure}

\cref{fig:sens-plot} compares the performance of two CNNs: ResNet-18 and DenseNet-121 with two ViTs: ViT-small and ViT-simple on various corruptions from the CIFAR-10-C dataset, at the end of training. We observe that the ViTs have lower sensitivity and better test performance on almost all corruptions compared to the CNNs, which have a higher sensitivity. 
Since the definition of sensitivity involves the addition of noise and ViTs have lower sensitivity, one can expect to be robust to 
various noise corruptions. However, 
the ViTs also have better test performance on several corruptions from weather and digital categories, which are significantly different from noise corruptions. This is consistent with the observations in \citet{mahmood2021robustness,bhojanapalli2021understanding}.

Next, we conduct an experiment to investigate the role of low sensitivity in the robustness of transformers. We add a regularization term while training the model to explicitly encourage it to have lower sensitivity. If this model is more robust, then we can disentangle the role of low sensitivity from the role of the architecture and establish a concrete connection between lower sensitivity and improved robustness. To add the regularization, we use the fact that sensitivity can be estimated efficiently via sampling and consider two methods. In the first method (augmentation), we augment the training set by injecting the images with Gaussian noise (mean $0$, variance $0.1$) while preserving the label, and train the ViT on the augmented training set. In the second method (regularization), we add a mean squared error term using the model outputs for the original image and the image with Gaussian noise (mean $0$, variance $1$) injected into a randomly selected patch. 

\begin{table}[t]
    \centering
    \caption{Comparison of two sharpness metrics at the end of training the ViT-small model on the CIFAR-10 dataset with and without the sensitivity regularization. Lower values correspond to flatter minima; see text for discussion.}
    \begin{tabular}{ccc}    
    \toprule 
    Setting &  ShOp  & ShPred  \\
    \midrule 
    ViT-small + vanilla training & $39.166$ &
    $0.5346$  \\
    ViT-small + sensitivity regularization & $9.025$ & $0.3982$ \\
    \bottomrule
    \end{tabular}
    \label{tab:sharpness}
\end{table}

\cref{fig:sens-plot} also shows the test performance of ViT-small trained with augmentation and regularization methods on various corruptions from CIFAR-10-C. We observe that ViTs trained with these methods exhibit lower sensitivity compared to vanilla training. This is accompanied by an improved test performance on various corruptions, particularly on the noise and blur categories. As encouraging lower sensitivity improves robustness, the inductive bias of transformers to learn functions of lower sensitivity could explain their better robustness 
(to common corruptions) compared to CNNs. 

\subsection{Lower Sensitivity leads to Flatter Minima} 

In this section, we investigate the connection between low sensitivity and flat minima. Consider a linear model $\f(\Tb;\xb) = \Tb^\top \xb$. Measuring sensitivity involves perturbing the input by some $\Delta \xb$. Prediction on the perturbed input 
is equivalent to perturbing the weight vector 
with $\Delta \Tb = \tfrac{\Tb^\top \Delta \xb}{\|\xb\|_2^2} \xb$, as
\begin{align*}
   \f(\Tb;\xb + \Delta \xb) = \Tb^\top (\xb + \Delta \xb) = \f(\Tb;\xb) + \Tb^\top \Delta \xb = \f(\Tb;\xb) + \Delta \Tb^\top \xb = \f(\Tb + \Delta \Tb;\xb).
\end{align*} 
This draws a natural connection between sensitivity, which is measured with perturbation in the input space, and flatness of minima, which is measured with perturbation in the weight space \citep{keskar2017on}. Below, we investigate whether such a connection extends to more complex architectures such as transformers. Given model $\f$ and train set $\D$, we consider two metrics to measure the flatness of the minimum, based on the model outputs and model predictions, respectively, 
\begin{align*}
\text{ShOp}&:=\!\!\!\underset{\xb\sim \D,\vct{\xi}\sim\calN(0,\sigma^2 \mat{I})}{\E}\!\!\!\!\!\! | \f(\Tb;\xb)-\f(\Tb+\vct{\xi};\xb)|, \,\, \text{ShPred}&:=\!\!\!\underset{\xb\sim \D,\vct{\xi}\sim\calN(0,\sigma^2 \mat{I})}{\E}\!\!\!\!\!\! \mathds{1}[ f(\Tb;\xb)\neq f(\Tb+\vct{\xi};\xb)], 
\end{align*}
where $f(\Tb;\xb) = \mathds{1}[\f(\Tb;\xb)\geq 0]$. Intuitively, for flatter minima, the model output and hence its prediction would remain relatively invariant to small perturbations in the model parameters.

\Cref{tab:sharpness} shows a comparison of these metrics for the ViT-small model trained with and without the sensitivity regularization at the end of training. Both metrics indicate that lower sensitivity corresponds to a flatter minimum. It is widely believed that flatter minima correlate with better generalization \citep{Jiang*2020Fantastic,keskar2017on,Neyshabur2017ExploringGI}, though 
they may not always be correlated \citep{Andriushchenko2023AML}. Our results indicate that low-sensitivity correlates with improved generalization and investigating this connection for other settings can be an interesting direction for future work.

\subsection{Sensitivity as a Progress Measure for Grokking} \label{sec:grokking}

\input{sections/grokking-fig}
In this section, we investigate if the sensitivity notion could serve as a progress measure for grokking \citep{nanda2023progress,chen2024sudden}. We train an one-layer Transformer model on the modular addition task $a + b \mod 113$. When evaluating sensitivity, we add a random Gaussian noise with $\sigma = 0.1$ to the number embeddings. As shown in \Cref{fig:sens-progress}, the test accuracy stays low from epoch 500 to 9,800 while the training accuracy saturates. However, sensitivity values continue to decrease smoothly starting from epoch 3000, and hence it provides a measure of the \emph{hidden progress} \citep{barak2022hidden} which the model makes even though the loss does not change, and indicates stages of grokking. In contrast, the weight norm is not a progress measure since it has the same flat curve as test accuracy during epoch 3,000 to 9,800. 

As discussed by \citet{nanda2023progress}, grokking occurs when the model learns to use Fourier features to solve the task. A further bump in sensitivity after the grokked phase at epoch 9800 suggests that the model initially learns less robust Fourier features. At this stage, a small random noise could slightly disrupt the model's performance. Over time, the Fourier basis becomes more robust. See \cref{app:grokking} for further discussion and results for more settings.

%% file: sections/grokking-fig.tex
\begin{figure}[t]
    \centering
    \subfigure[Train \& Test accuracy]{\includegraphics[width=0.32\linewidth]{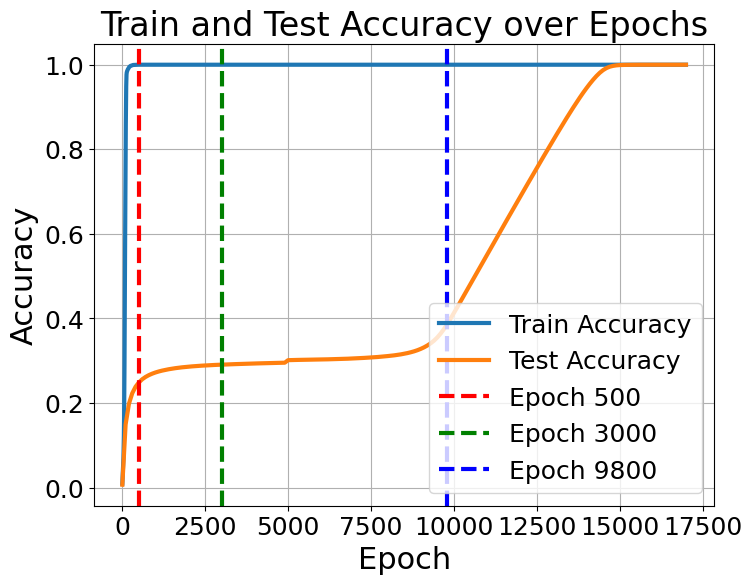}}
    \subfigure[Sensitivity]{\includegraphics[width=0.32\linewidth]{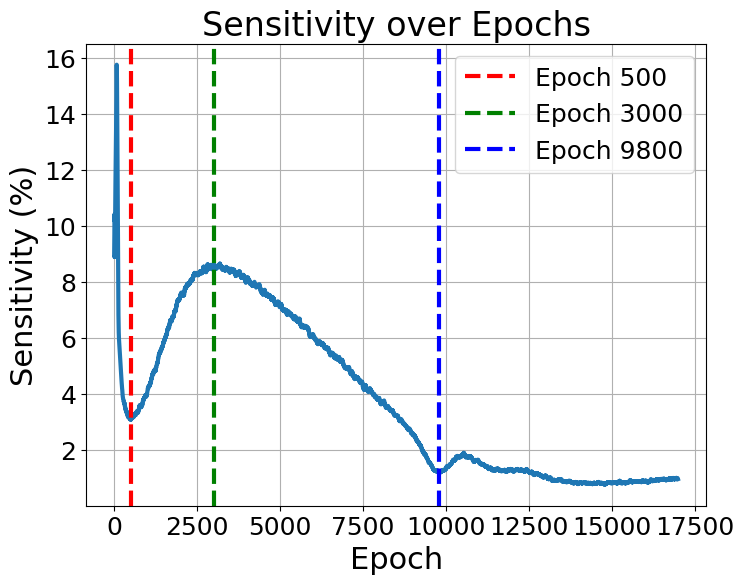}}
    \subfigure[Weight norm]{\includegraphics[width=0.32\linewidth]{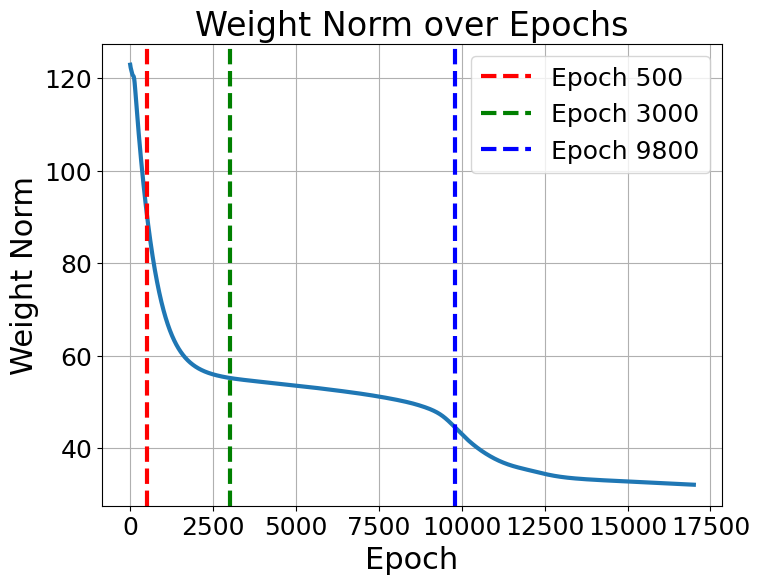}}
    \caption{Sensitivity measures progress on modular addition task $a + b \mod 113$ and indicates different stages of grokking. 
    }
    \label{fig:sens-progress}
\end{figure}

%% file: sections/conclusion.tex
\section{Conclusion}
In this work, we investigate how the notion of sensitivity, which has shown promise in understanding inductive biases of Transformers on Boolean functions in prior work, can be extended to more realistic settings involving real-valued data. Our results show that transformers learn functions that have low sensitivity to small token-wise input perturbations, compared to other architectures, across vision and language tasks. We corroborate these observations with theoretical results, showing that transformers exhibit spectral bias and lower sensitivity corresponds to better robustness. We also demonstrate three important implications of this low-sensitivity bias: it correlates with improved robustness, flatter minima in the loss landscape, and serves as a progress measure that offers insights about the training dynamics. Investigating sensitivity as a progress measure in more settings can be an interesting direction for future work.

%% file: sections/extra-results.tex
\section{Additional Experiments}
In this section, we include some additional results to supplement the main experimental results for synthetic data as well as the vision and language tasks.
\input{sections/mnist}
\input{sections/figure_extra}

\subsection{Sensitivity with Random Noise instead of Token-wise Noise} 
In this section, we consider changing the way we compute sensitivity, to see if the resulting metric also distinguishes transformers from other architectures. Instead of token-wise perturbations, we add Gaussian noise across the entire input with a smaller variance so that the transformer's sensitivity in this case is similar to the sensitivity with token-wise noise. 

\begin{table}[h!]
    \centering
    \caption{Comparison of sensitivity values measured with random and patch-wise noise for various model-dataset settings. Token-wise perturbations lead to a larger gap between the sensitivity of transformer-based models compared to other architectures.}
    \begin{tabular}{ccc}
    \toprule
         Model and dataset & Random noise & Token-wise noise \\
         \hline
 ResNet-18 on CIFAR-10 & 0.0172 & 0.0827 \\
 ViT-small on CIFAR-10 & 0.0082 & 0.0050 \\
 \hline
 LSTM on QQP & 0.11 & 0.09 \\
 RoBERTa on QQP & 0.05 & 0.03 \\
 \bottomrule
    \end{tabular}
    
    \label{tab:my_label}
\end{table}
In \cref{tab:my_label}, we compare the sensitivity values at the end of training for ResNet-18 and ViT-small on the CIFAR-10 dataset (variance $0.025$) and LSTM and RoBERTa on the QQP dataset (variance $0.5$). We find that for random perturbations, the difference between sensitivity values is much smaller for CIFAR-10 and similar for the QQP dataset, compared to patch-wise perturbations. These results suggest that measuring sensitivity with patch-wise noise is indeed the metric that we should consider since it distinguishes transformers from other architectures with a larger gap. 

\subsection{Vision Tasks} 
\label{app:add-expts}

\paragraph{Effect of Depth, Number of Heads and the Optimization Algorithm.} In \cref{fig:sens-opt}, we compare the sensitivity values of a ViT-small model trained on CIFAR-10 dataset with SGD and Adam optimization algorithms. Although the model trained with Adam has a slightly higher sensitivity, the sensitivity values for both the models are quite similar. This indicates that the low-sensitivity bias is quite robust to the choice of the optimization algorithm.

\begin{figure}[h!]
    \centering
    \subfigure[Sensitivity for varying depth]{\includegraphics[width=0.4\linewidth]{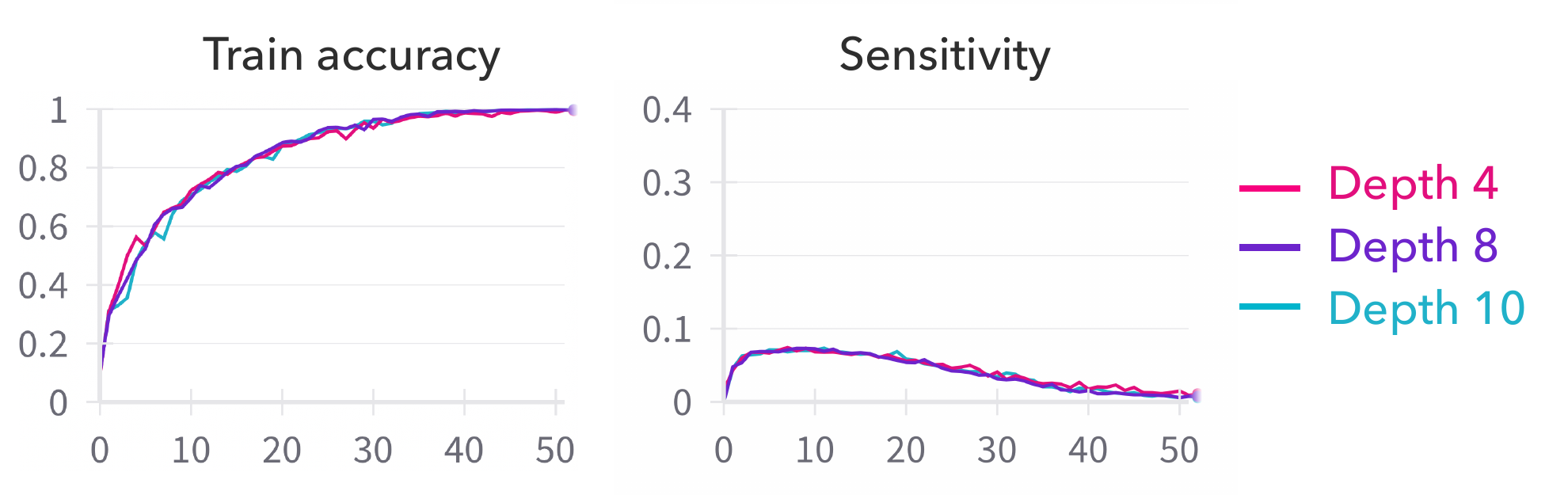}}\hspace{0.5cm}
    \subfigure[Sensitivity for varying number of heads]{\includegraphics[width=0.4\linewidth]{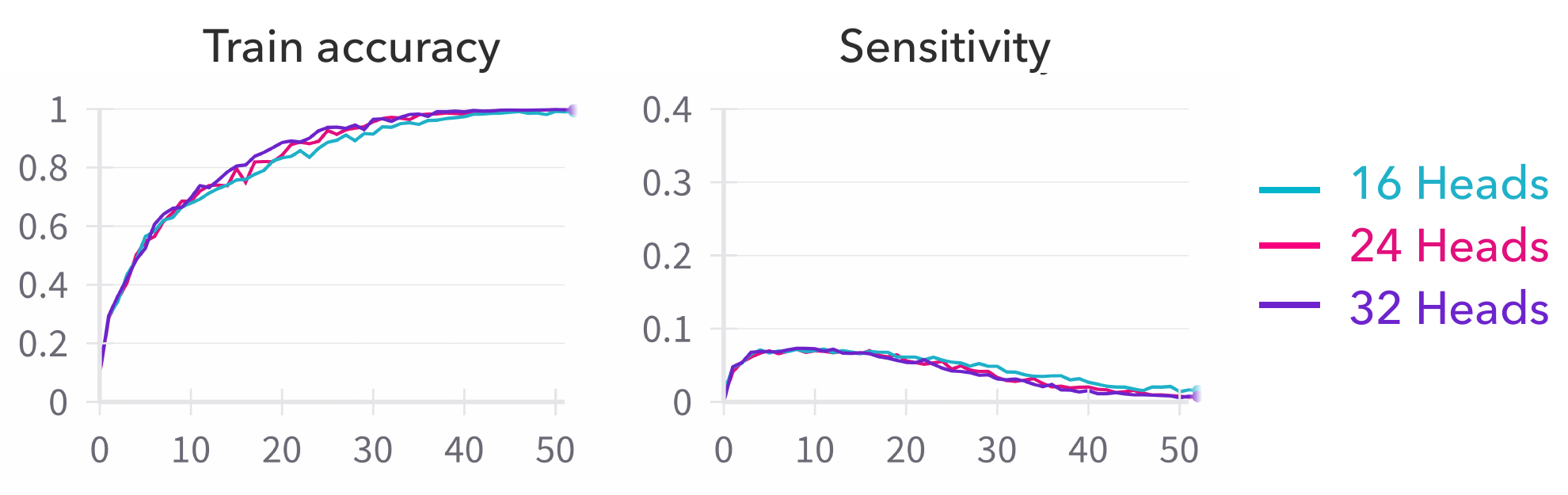}}
    \caption{\textbf{Sensitivity for Various Model Settings.} Comparison of train accuracies and sensitivity values on the CIFAR-10 dataset when varying the depth and number of heads of the ViT-small model. We observe that for the same train accuracy, the sensitivity values remain very similar for different model settings.}
    \label{fig:sens-depth-heads}
\end{figure}

In \cref{fig:sens-depth-heads}, we compare the sensitivity values of a ViT-small model with different depth and number of attention heads, when trained on the CIFAR-10 dataset. Note that for our main results, we use a model with depth $8$ and $32$ heads. We observe that the train accuracies and the sensitivity values remain the same across the different model settings. This indicates that the low-sensitivity bias is quite robust to the model setting.
\begin{figure}[h!]
    \centering
    \includegraphics[width=\linewidth]{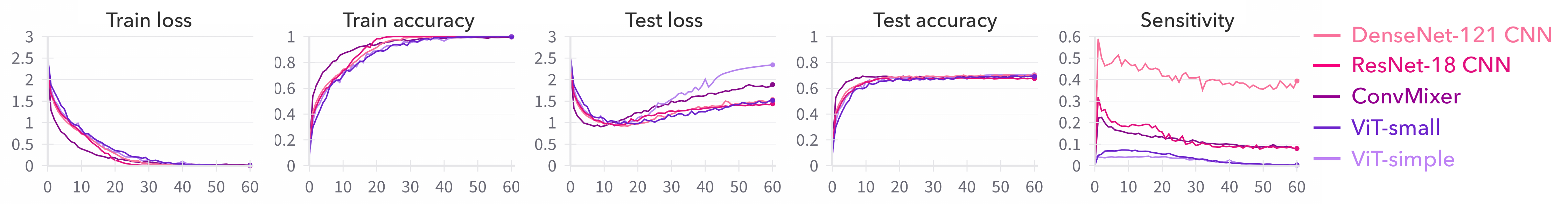}
    \caption{Comparison of the sensitivity of two CNNs, two ViTs, and ConvMixer trained on the CIFAR-10 dataset, as a function of training epochs. For a fair comparison, the figure also shows the train and test accuracies and loss values (cross-entropy loss). All models have similar accuracies but the ViTs have significantly lower sensitivity than the other models.}
    \label{fig:cifar-full}
\end{figure}

\begin{wrapfigure}[14]{R}{0.5\textwidth}
    \centering
    
    \includegraphics[width=0.8\linewidth]{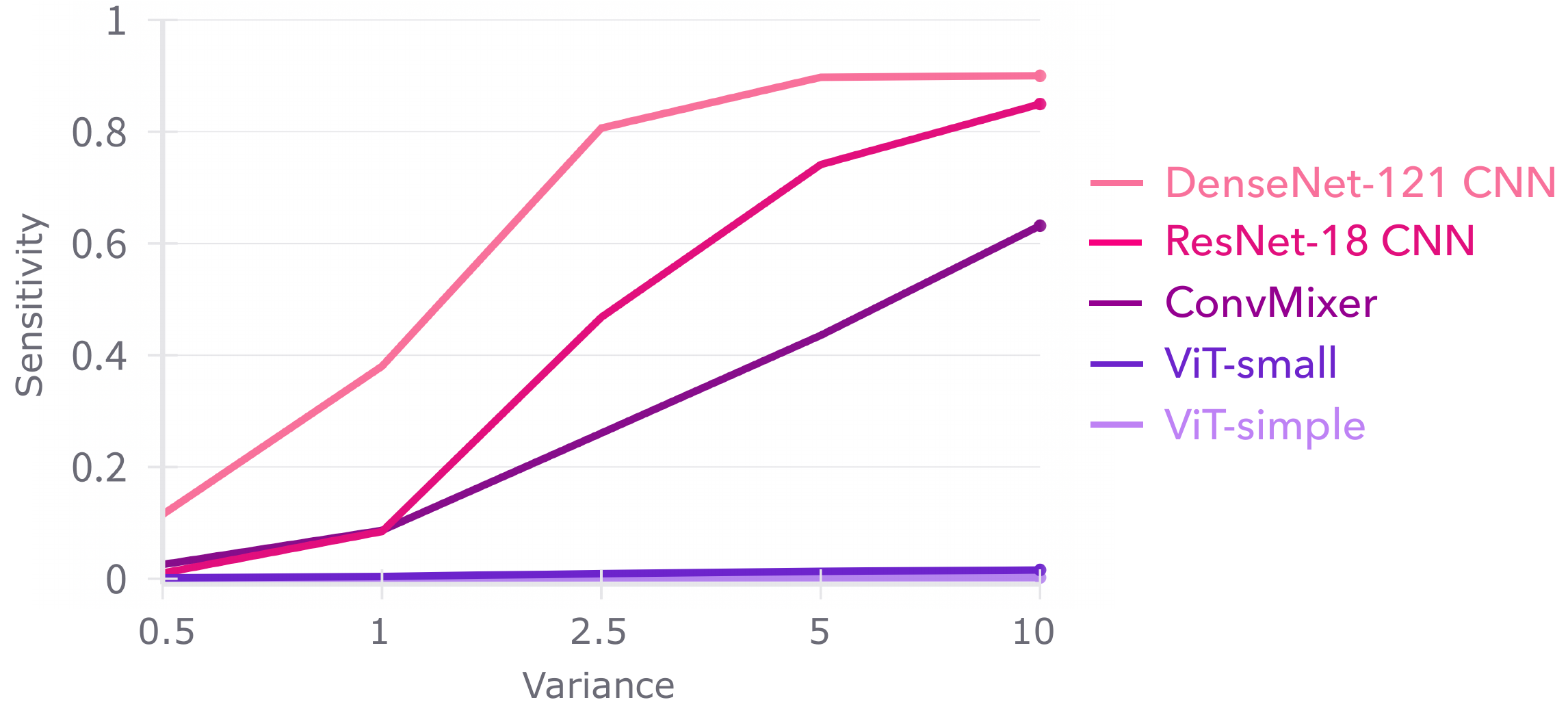}
    \caption{\textbf{Sensitivity for Different Variances.} Comparison of the sensitivity of two CNNs, two ViTs, and ConvMixer trained on the CIFAR-10 dataset, as a function of different variance levels, at the end of training. 
    The ViTs have significantly lower sensitivity at any variance and the difference grows as variance increases. 
    }
    
    \label{fig:cifar_vars}
\end{wrapfigure}

\paragraph{Effect of Variance.} In \cref{fig:cifar_vars}, we compare the effect of the variance $\sigma^2$ used while evaluating sensitivity for different models trained on the CIFAR-10 dataset. We observe that the ViTs have significantly lower sensitivity than the other models and the difference becomes starker as the variance level increases.

\paragraph{Results on SVHN Dataset.} \cref{fig:svhn} shows the training accuracy and sensitivity of a ResNet-18 and a ViT-small model trained on SVHN dataset \citep{svhn}. At the end of training, the sensitivity values are: $0.0516$ for ResNet-18 and $0.0147$ for ViT-small. Similar to the observations for CIFAR-10, we see that the ViT has a significantly lower sensitivity. 
\input{sections/figure_vision}

\begin{figure*}[h!]
    \centering    \includegraphics[width=0.9\linewidth]{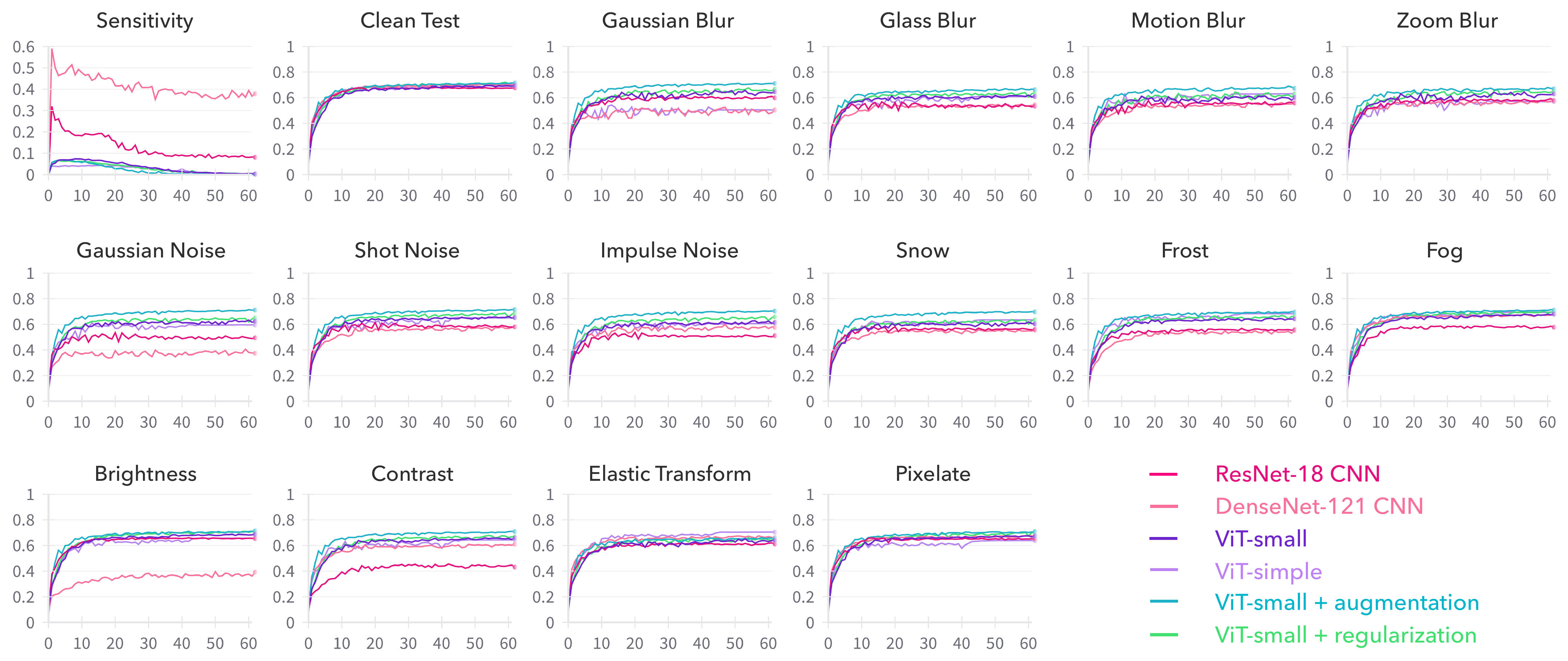}
    \caption{
    Comparison of the sensitivity, test accuracy on CIFAR-10 and test accuracies on various corruptions 
    from the CIFAR-10-C dataset (see \cref{sec:implications} for details) of two CNNs and two ViTs trained on the CIFAR-10 dataset, as a function of the training epochs. We also compare with ViT-small trained with data augmentation/regularization, which encourage low sensitivity (see \cref{sec:implications} for discussion).}
    \label{fig:cif10c}
    \
\end{figure*}
\begin{figure*}[h!]
    \centering
    \includegraphics[width=0.9\linewidth]{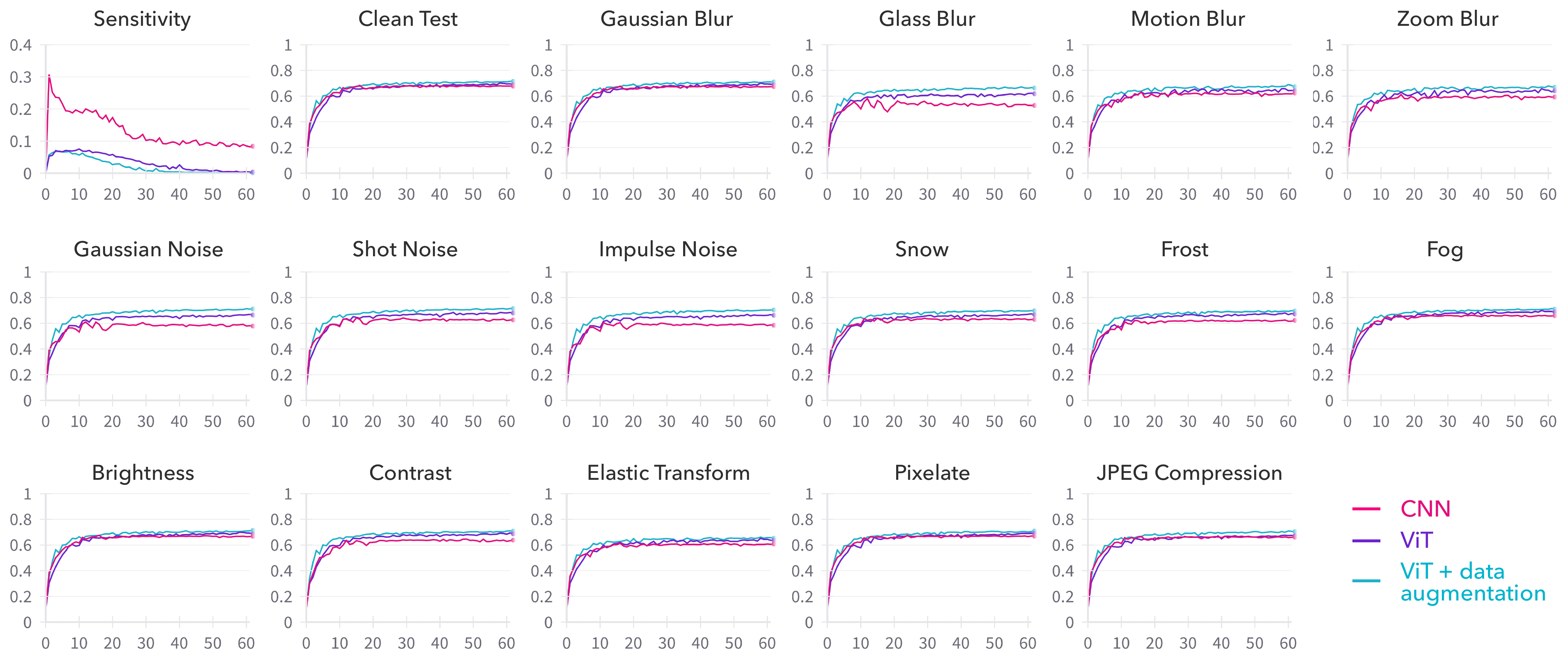}
    \caption{ 
    Comparison of the sensitivity, test accuracy on CIFAR-10, and test accuracies on various corruptions from the CIFAR-10-C dataset (see \cref{sec:implications} for details) of a ResNet-18 CNN and a ViT-small model trained on the CIFAR-10 dataset, as a function of the training epochs. We also compare with ViT-small trained with data augmentation, which acts as a regularizer to encourage low sensitivity (see \cref{sec:implications} for discussion). Here, we use severity level $1$, while in \Cref{fig:cif10c}, we considered severity level $2$. }
    \label{fig:cc0}
\end{figure*}

\paragraph{Additional Results on CIFAR-10-C.} \cref{fig:cif10c} and \cref{fig:cc0} show the test performance on various corruptions from the CIFAR-10-C dataset with severity level $2$ and $1$, respectively. We observe that CNNs have lower test accuracies on corrupted images compared to ViTs. Further, encouraging lower sensitivity in the ViT leads to better robustness. 


\subsection{Language Tasks} \label{app:add-nlp-expts}
\begin{figure}[h!]
    \centering
    \subfigure
    {\includegraphics[width=0.26\linewidth]{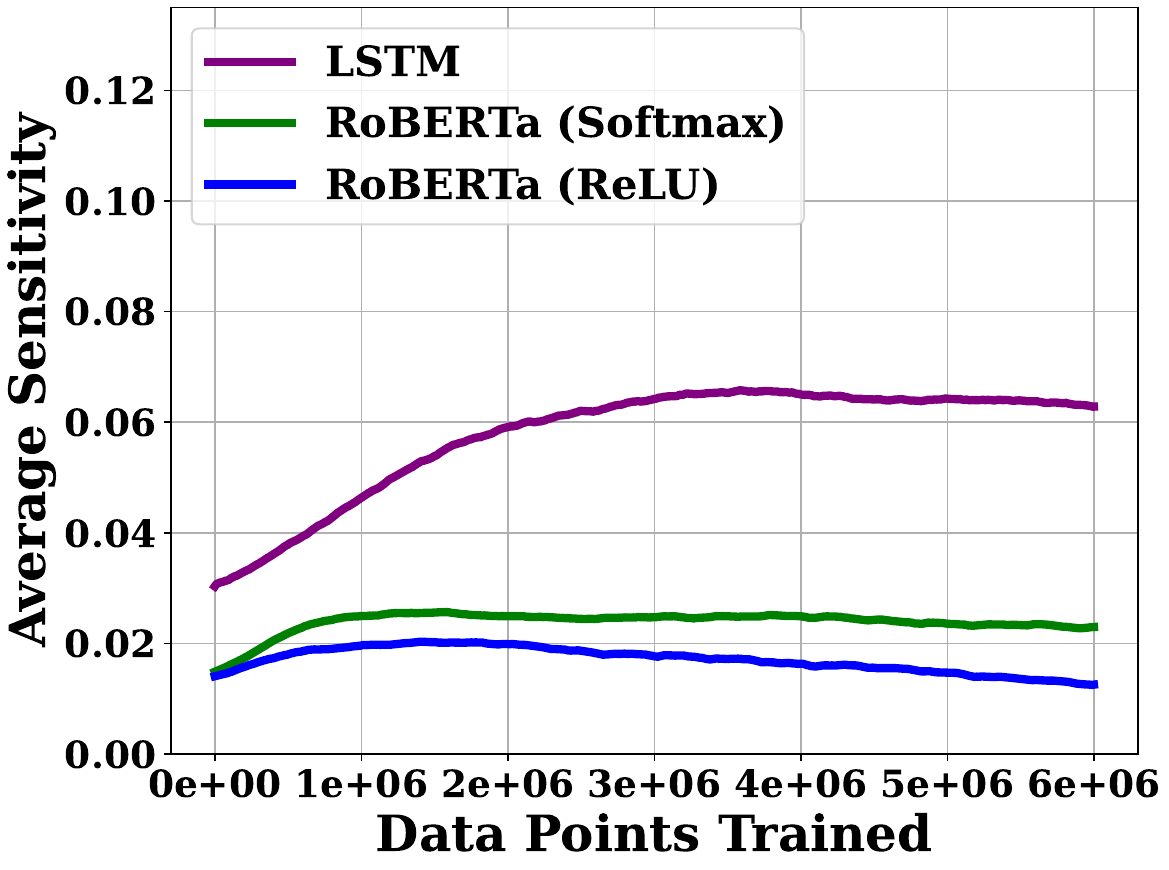}}\hspace{0.25cm}
    \subfigure
    {\includegraphics[width=0.28\linewidth]{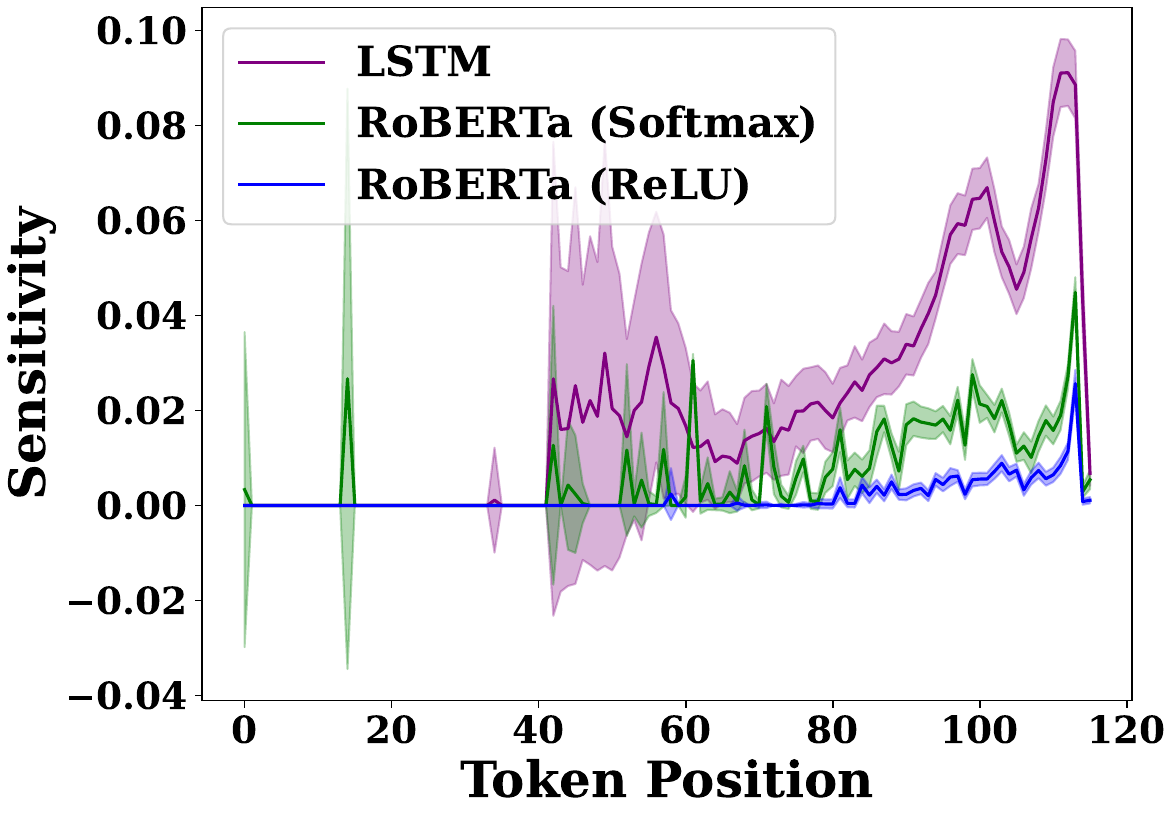}}
    \caption{\textbf{Sensitivity on the QQP Dataset with Variance $\sigma^2=4$.} Results with alternative variance yield observations that are consistent with the setup in the main text. (Left) LSTM has higher sensitivity than the RoBERTa models. (Right) Softmax activation for RoBERTa induces higher sensitivity towards the CLS token.
    }
    \
    \label{fig:sensitivity_nlp_var_4}
\end{figure}

\paragraph{Sensitivity Measured with Variance $\sigma^2 = 4$.} Alternative to the main experiments with $\sigma^2 = 15$, we also evaluate sensitivity with a different corruption strength $\sigma^2 = 4$ on the QQP dataset, as shown in \cref{fig:sensitivity_nlp_var_4}. We observe the same trends as in \Cref{fig:nlp-sensitivity,fig:nlp-sensitivity-position}: RoBERTa models have lower sensitivity than the LSTM and the LSTM is more sensitive to more recent tokens. These results indicate that the low-sensitivity bias is robust to the choice of corruption strength.

\begin{figure}[h!]
    \centering
    \subfigure[MRPC]
    {\includegraphics[width=0.27\linewidth]{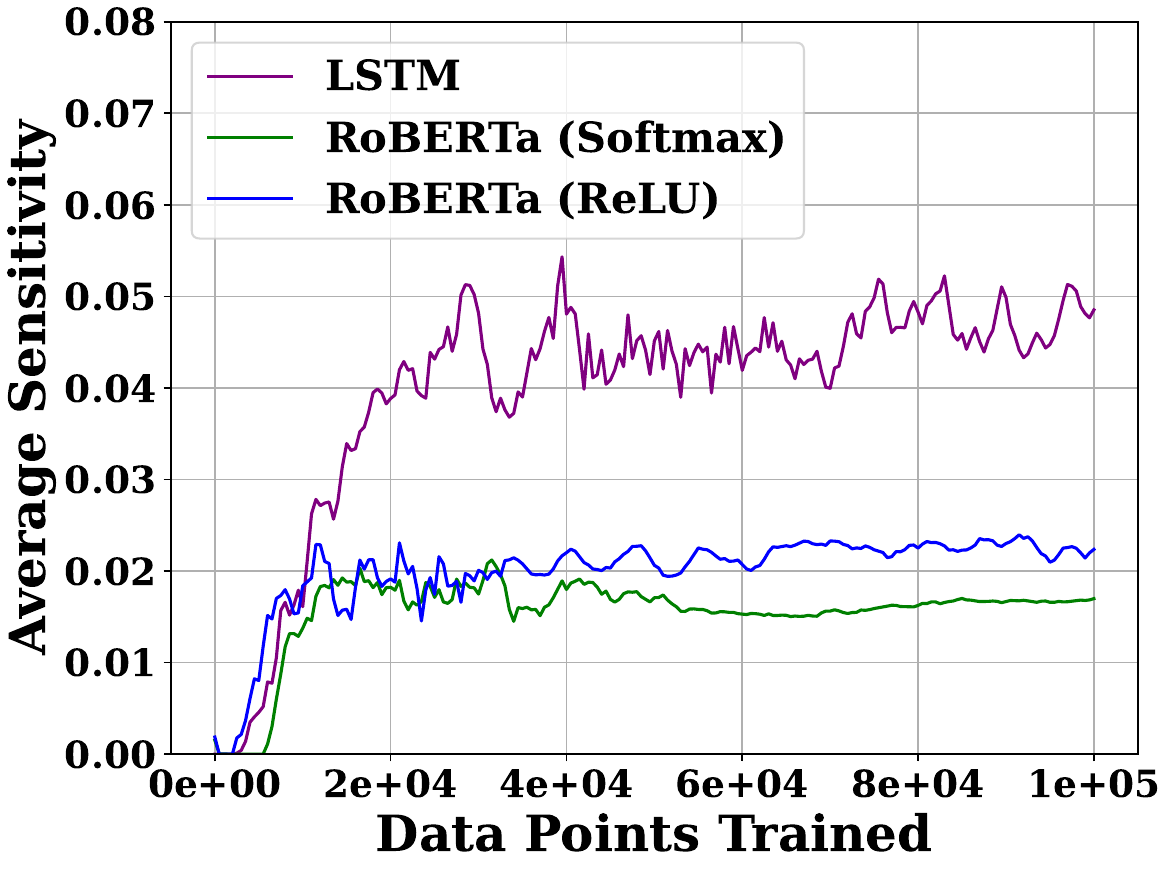}}\hspace{0.25cm}
    \subfigure[QQP]
    {\includegraphics[width=0.27\linewidth]{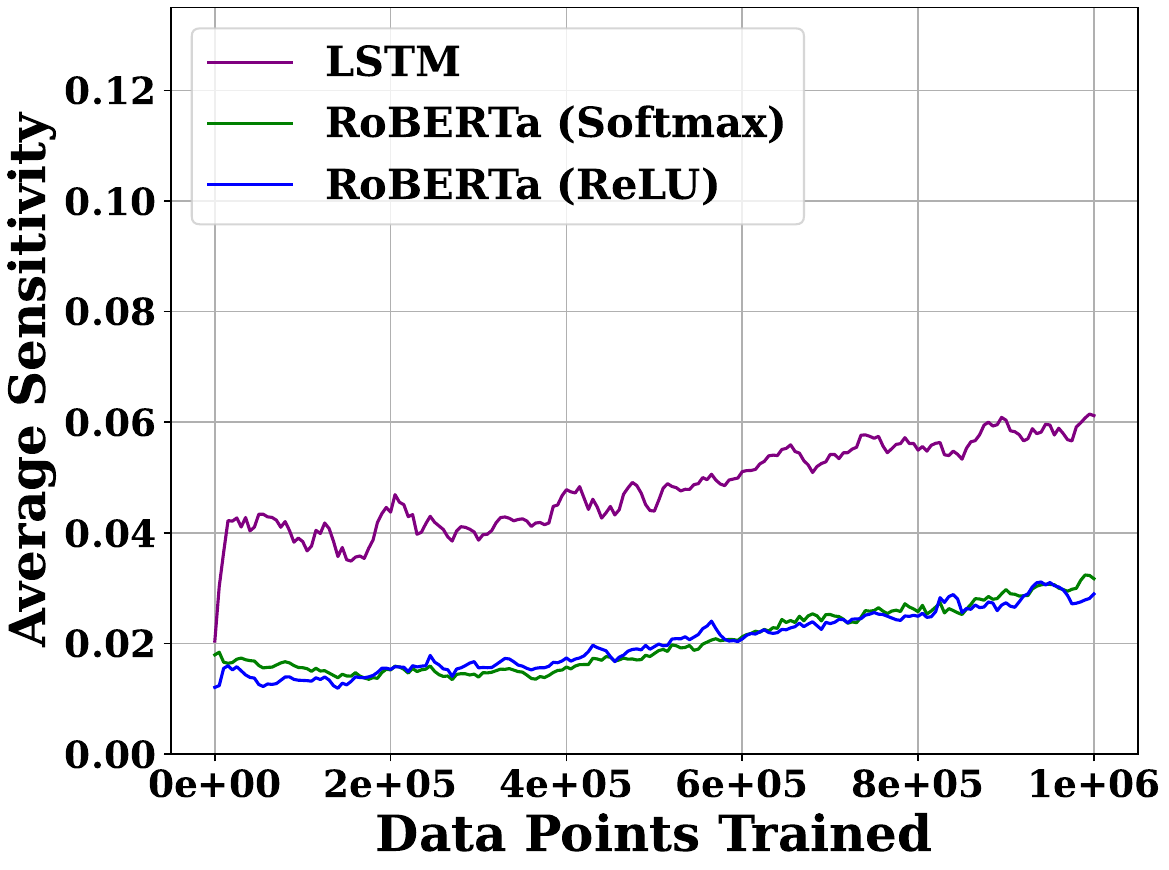}}
    \caption{\textbf{Sensitivity on the Validation Sets.} Similar to the observation in \Cref{fig:nlp-sensitivity}, the RoBERTa models have lower sensitivity than the LSTM for both the datasets. However, the difference between RoBERTa-ReLU and RoBERTa-softmax is less marginal on the validation set compared to the training set.}
    \label{fig:sensitivity_nlp_val}
    \
\end{figure}

\paragraph{Sensitivity Measured on the Validation Set.} We also test sensitivity on the validation set for the two language datasets. As seen in \Cref{fig:sensitivity_nlp_val}, the results on the validation set are consistent with those on the train set in \cref{fig:nlp-sensitivity}.

\begin{figure}[h!]
    \centering
    \subfigure
    {\includegraphics[width=0.26\linewidth]{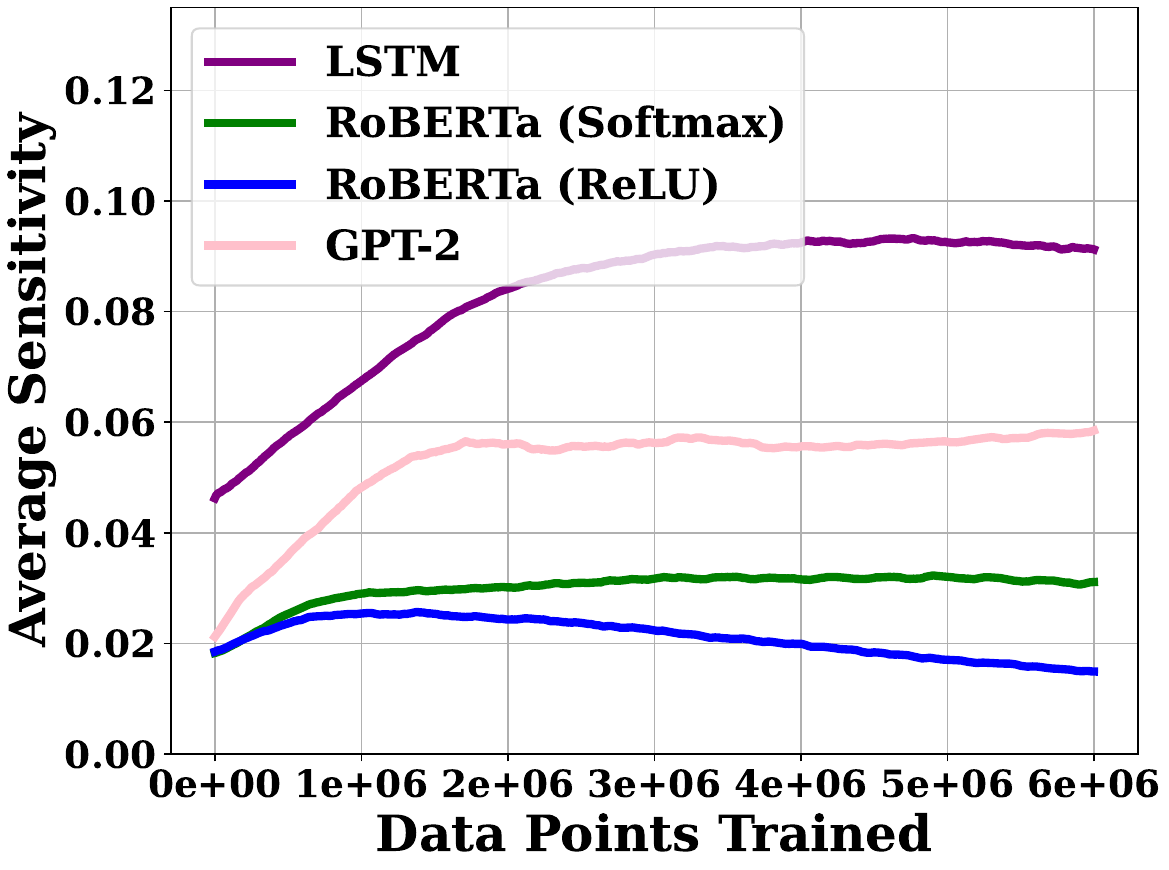}}\hspace{0.25cm}
    \subfigure
    {\includegraphics[width=0.27\linewidth]{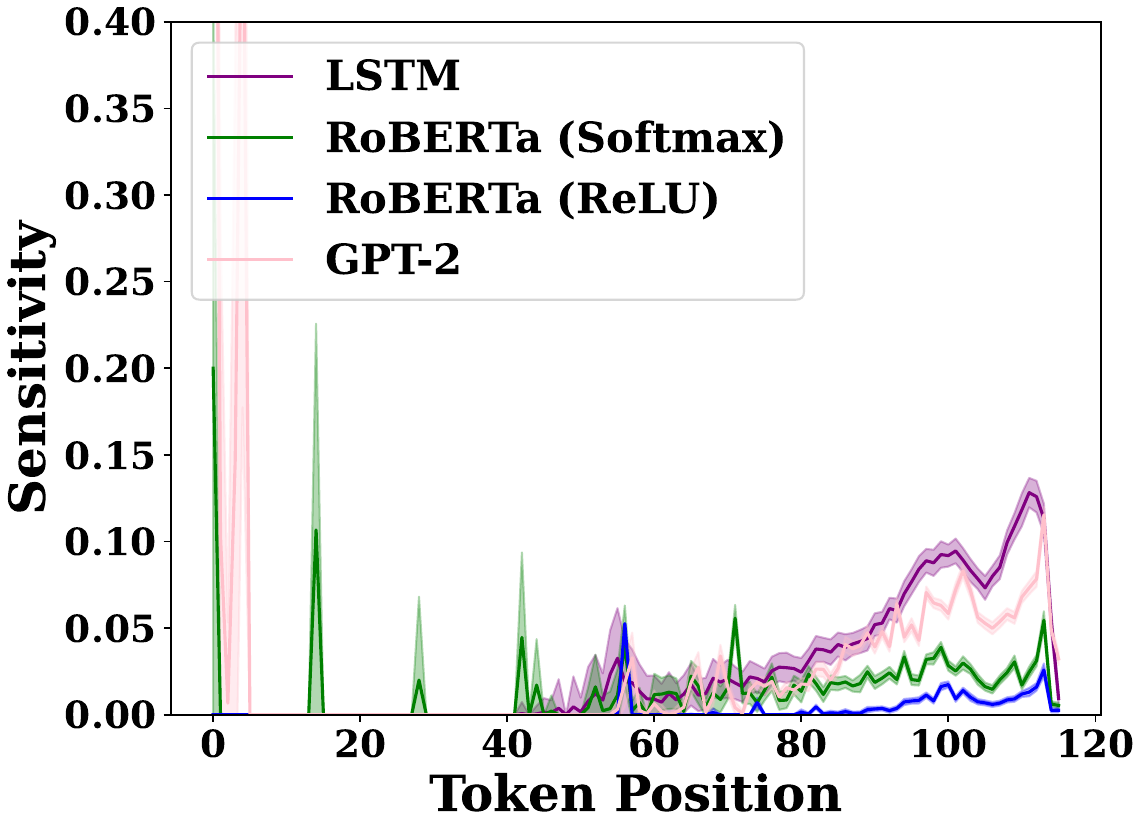}}
    \caption{\textbf{Sensitivity of GPT-2 on the QQP Dataset.} (Left) We find that the RoBERTa models tend to have lower sensitivity than GPT-2, and all Transformer models have lower sensitivity than LSTM. (Right) The sensitivity per token of GPT-2 is more similar to LSTMs, which is possibly due to their shared auto-regressive design.}
    \label{fig:sensitivity_nlp_gpt}
    \
\end{figure}

\paragraph{Sensitivity Measured with the GPT-2 Model.} Here, we ablate the effect of the transformer architecture on the sensitivity values. We compare a GPT-based model with the two BERT-based models used in our main experiments. The key difference lies in the construction of the attention masks: for GPT models, each token only observes the tokens that appear before it, whereas BERT models are bidirectional, therefore each token observes all the tokens in the sequence. In \cref{fig:sensitivity_nlp_gpt}, we observe that the GPT-2 model has higher sensitivity compared to the RoBERTa models, but the sensitivity is significantly lower than the LSTM. The GPT-2 model is also relatively more sensitive to more recent tokens compared to the RoBERTa models, while also being sensitive to some CLS tokens. 

\begin{figure}[h!]
    \centering
    \includegraphics[width=0.4\linewidth]{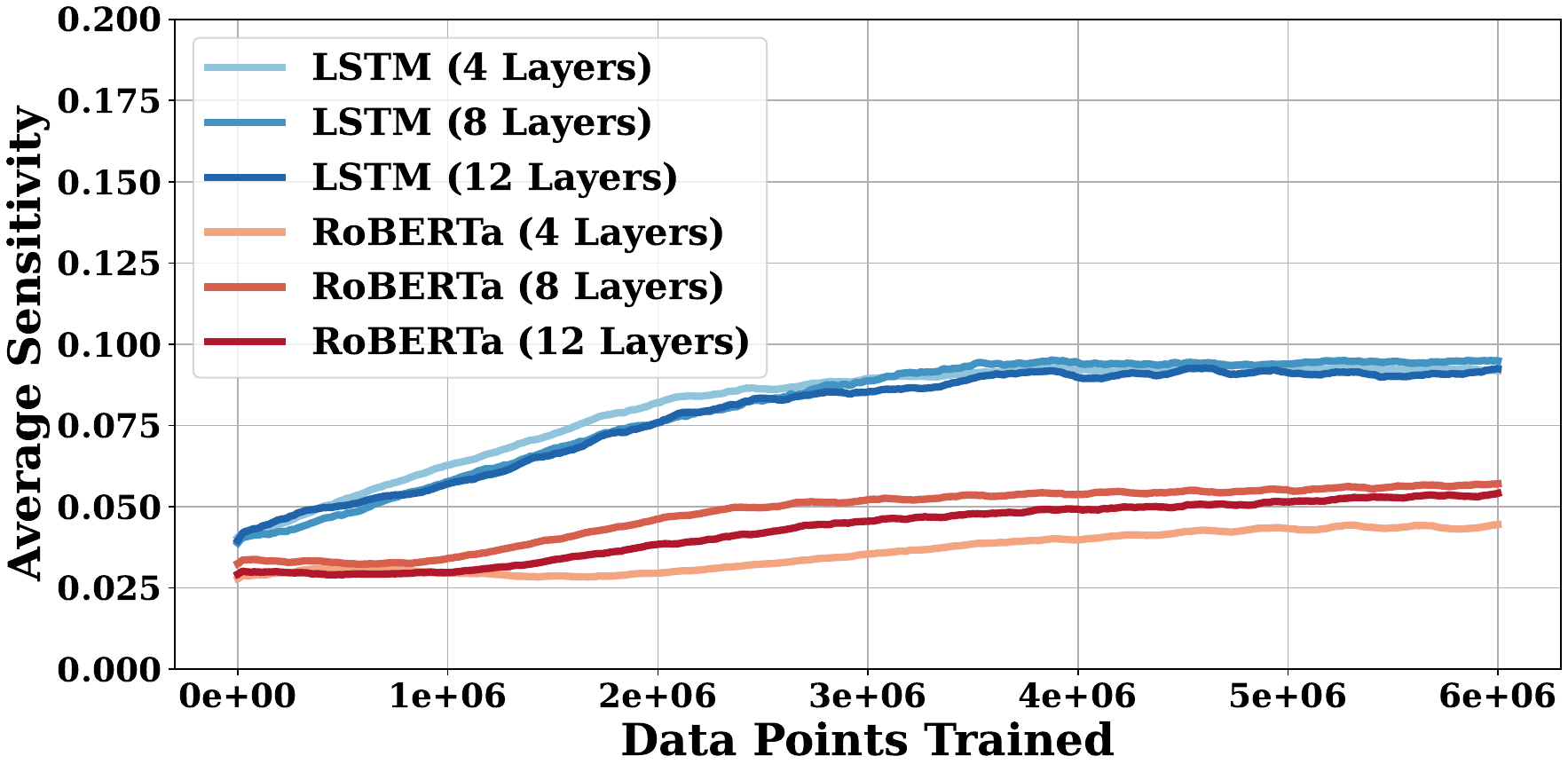}
    \caption{\textbf{Sensitivity for Different Model Depths}. We vary the model depths of LSTM and RoBERTa on the QQP datasets and observe that LSTM models tend to have the same sensitivity throughout the entire training. RoBERTa model with 4 layers has slightly lower sensitivity with its 8-layer or 12-layer variants. RoBERTa, regardless of depths, have lower sensitivity than LSTMs.}
    \label{fig:qqp_sensitivity_by_layers}
    \
\end{figure}

\subsection{Sensitivity as a Progress Measure for Grokking} \label{app:grokking}

Grokking is a phenomenon in which a neural network suddenly and drastically improves its generalization ability after a long period of training, even though it initially overfits to the training data \citep{nanda2023progress}. During grokking, the model transitions from memorizing the training data to learning a more general solution, allowing it to perform well on unseen data. This often happens after many training steps, during which time the test accuracy remains low despite perfect training accuracy. The transition is abrupt, making grokking seem like an emergent behavior where the model, after much training, "figures out" the correct approach to the task.

\begin{figure}[h!]
    \centering
    \subfigure[Train \& Test accuracy]{\includegraphics[width=0.32\linewidth]{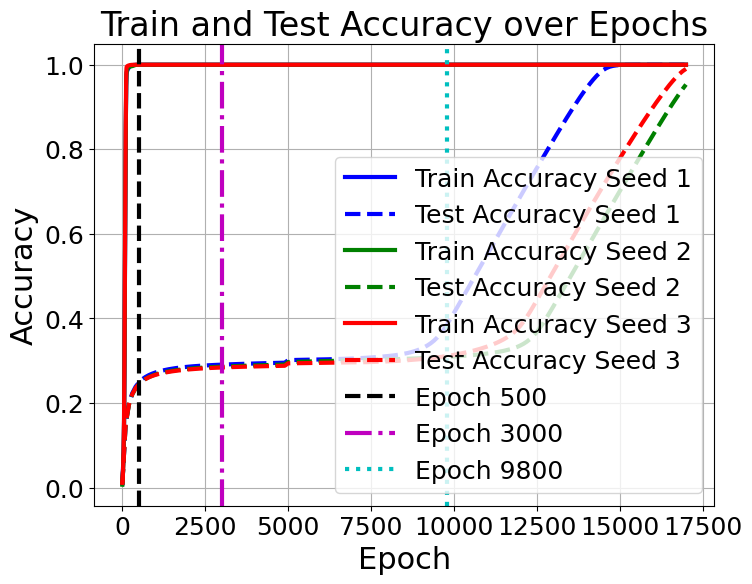}}
    \subfigure[Sensitivity]{\includegraphics[width=0.32\linewidth]{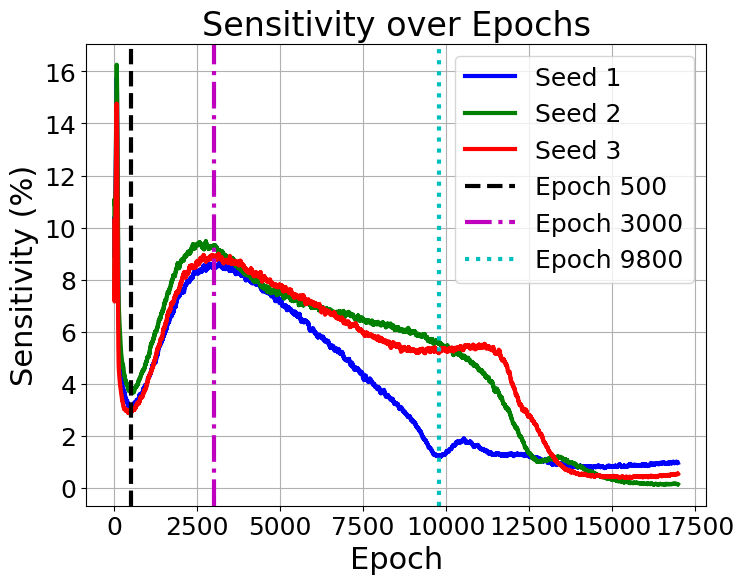}}
    \subfigure[Weight norm]{\includegraphics[width=0.32\linewidth]{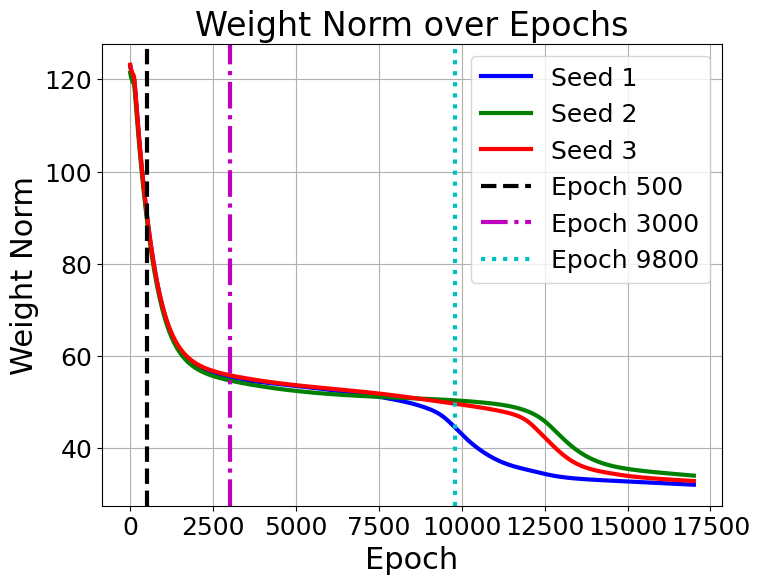}}
    \
    \caption{Sensitivity measures progress on modulo addition task $a + b \mod 113$ and indicates different stages of grokking. 
    }
    \label{fig:sens-progress-appendix}
\end{figure}

In more technical terms, this shift occurs as the network amplifies structured mechanisms that enable generalization and removes components that only lead to memorization. Grokking has been observed in models trained with regularization techniques like weight decay on algorithmic tasks, where the model learns an underlying structure that generalizes well beyond the training data. In \citep{nanda2023progress}, the authors demonstrate that for the modular addition task, the model initially memorizes the training data, leading to low test accuracy. Later, the model discovers how to use trigonometric functions to solve the task. However, the emergence of grokking is difficult to measure in practice, and it is not guaranteed that grokking will occur or that the model will find the correct approach to the task. Nevertheless, as shown in Figure \ref{fig:sens-progress-appendix}, we can clearly observe a significant change in sensitivity between epochs 3000 and 9800, during the saturation of the training loss. This suggests that the model is discovering a more robust solution to the task. We propose that sensitivity can serve as a useful metric for assessing whether the model is learning to grok.

Similar to \citet{nanda2023progress}'s classification on phases: memorization, circuit formation, and cleanup, we claim that sensitivity also provides a progress measure, with an extra phase of noise reduction:

\textbf{Memorization: } From epoch 0 to 500, sensitivity drops significantly while training accuracy saturates to 100 and test accuracy increases but cannot saturate.  

\textbf{Circuit Formation: } From epoch 500 to 3,000, the sensitivity goes up again but test accuracy remains low and flat. The dramatic fall in the weight norms suggests that circuit formation likely happens due to weight decay. 

\textbf{Clean up: } From epoch 3,000 to 9,800, sensitivity starts to decrease, and the progress measure indicates that the model starts to learn to use Fourier features. 

\textbf{Noise Reduction: } From epoch 9,800 onwards, sensitivity has an upward and then downward trend, and this is not caught by other mechanistic interpretability measures in \citet{nanda2023progress}. This trend in sensitivity is due to further Fourier noise reduction, where initially, slightly perturbations to the number embeddings could drift the model performance where later on, model learns a more robust Fourier basis.

\subsection{Sensitivity as a Progress Measure for Learning Sparse Parities}
\label{app:parity}
In this section, we investigate if sensitivity also acts as a progress measure when training transformers on the sparse parity task, with input $\xb\in\{\pm1\}^d$ and label $y=\textstyle\prod_{i\in S}x_i$, where $S\subset[d]$ with sparsity level $p:=|S|<d$. 

\begin{figure}[h!]
    \centering
    \includegraphics[width=0.8\linewidth]{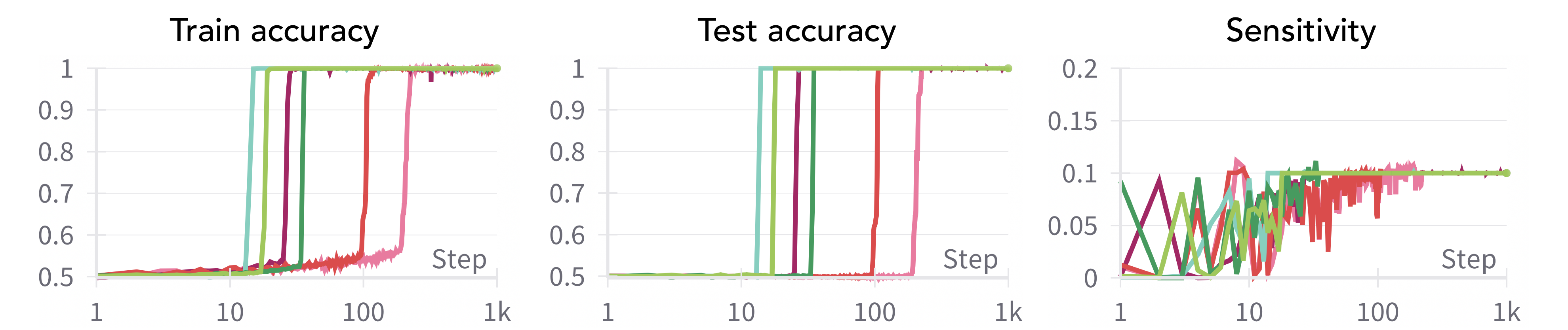}
    \caption{Train accuracy, test accuracy, and sensitivity as a function of training epochs, when training a two-layer transformer on the sparse parity task with dimension $40$ and sparsity level $4$. Sensitivity increases while train and test accuracy are close to $50\%$, and converges to $0.1$ when the model learns the sparse parity function, which coincides with train and test accuracy going to $100\%$.}
    \label{fig:sparse-parity}
\end{figure}

We train a two-layer transformer model on this task with $4$ heads using Adam optimizer for $1000$ epochs. We evaluate the sensitivity metric based on \cref{eq:sens-bool}, using $10^5$ samples to estimate the expectation over the Boolean cube. \cref{fig:sparse-parity} shows the results on six different settings in terms of (number of train samples, batch size, embedding dimension, learning rate, seed): 
\begin{center}
    $\circ$ Sea green: $(50k, 250, 32, 0.001, 123)$ \quad $\circ$  Parrot green: $(50k, 250, 32, 0.001, 42)$\\ $\circ$
    Dark green: $(50k, 250, 32, 0.001, 0)$\quad$\circ$
    Maroon: $(10k, 100, 64, 0.001, 42)$\\ 
    $\circ$ Coral: $(5k, 25, 32, 0.0001, 42)$\quad $\circ$
    Pink: $(5k, 50, 32, 0.0001, 42)$.
\end{center}
We observe that in all the settings, the train and test accuracy remain close to $50\%$ for the initial few epochs, while sensitivity increases, which indicates progress in learning the sparse parity task. When the model learns the sparse parity function, the sensitivity converges to $\tfrac{p}{d}=0.1$, which coincides with train and test accuracy going to $100\%$.

%% file: sections/mnist.tex
\subsection{Synthetic Data and the MNIST Dataset}\label{app:add-expts-synth}
In this section, we present some additional results for the low-sensitivity bias of a single-layer self-attention model (\cref{eq:attn-model}) on the synthetic dataset generated based on \cref{def:gen_data},  visualized in \cref{fig:vocab}. Similar to the results in \cref{sec:synth}, we consider three data settings where using the sparse token leads to a function with lower sensitivity (\cref{fig:synth-joint2}, top row) and three settings where using the frequent token leads to lower sensitivity (\cref{fig:synth-joint2}, bottom row). The exact data settings and a comparison of the sensitivity values for each setting are shown in \cref{tab:sens-synth2}. These results yield similar conclusions as in \cref{sec:synth}: in both cases, the model uses tokens which leads to a lower sensitivity function. 
\input{sections/figure_synth_full}

Continuing from the synthetic data, we now consider a slightly more complicated dataset, namely MNIST \citep{LeCun2005TheMD}. The MNIST dataset consists of $70k$ black-and-white images of handwritten digits of resolution $28 \times 28$. There are $60k$ images in the training set and $10k$ images in the test set. It is released under the CC BY-SA 3.0 license. We 
compare the sensitivity of a ViT-small model with an MLP on a binary digit classification task ($<5$ or $\geq 5$). In our experiments, each image is divided into $T = 16$ patches of size $7 \times 7$ for the ViT-small model. For the MLP, the inputs are vectorized as usual. With this setting, we measure the sensitivity of the two models using patch token replacement as per \Cref{def:sensitivity}. As shown in \Cref{fig:mnist}, when achieving the same training accuracy, the ViT shows lower sensitivity compared to the MLP. 

%% file: sections/figure_synth_full.tex
\begin{figure*}[h!]
    \centering
    \begin{minipage}[b]{\linewidth}
    \centering
\begin{minipage}[b]{\linewidth}
\centering
\captionof{table}{Comparison of sensitivity values for models that use only sparse or frequent tokens for the settings considered in \cref{fig:synth-joint2}.}
\resizebox{0.8\linewidth}{!}{%
\begin{tabular}{lccccccc}    
\hline 
& \multicolumn{3}{c}{Top row in \cref{fig:synth-joint} $(n_s=3, n_d=1)$} & \multicolumn{3}{c}{ Bottom row in \cref{fig:synth-joint} $(n_s=1, n_d=7)$} \\
Data Setting $(n_f, m)$ & \textcolor[HTML]{E08382}{$(3,6)$} & \textcolor[HTML]{E20F7D}{$(5,16)$} & \textcolor[HTML]{6D24CC}{$(7,28)$} & \textcolor[HTML]{E08382}{$(7,10)$} & \textcolor[HTML]{E20F7D}{$(17,20)$} & \textcolor[HTML]{6D24CC}{$(32,36)$} \\
\hline 
Using sparse tokens & $0$ & $0$ & $0$ & $0.0339$ & $0.0339$ & $0.0339$\\
Using frequent tokens & $0.1315$ & $0.2878$ & $0.4502$ & $0$ & $0$ & $0$\\
\hline
\end{tabular}}

    \label{tab:sens-synth2}
\end{minipage}
        \begin{minipage}[b]{\linewidth}
            \centering
\includegraphics[width=0.8\linewidth]{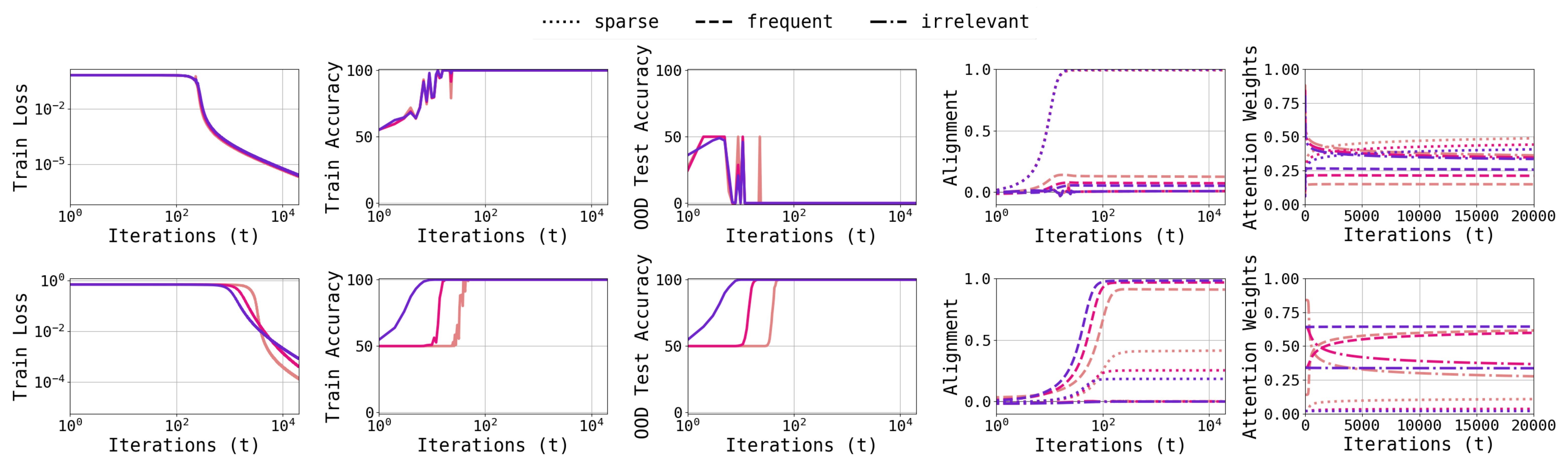}
    \caption{Train and test dynamics for a single-layer self-attention model (\cref{eq:attn-model}) using the synthetic data visualized in \cref{fig:vocab}; see \cref{sec:synth} for details. The top row corresponds to the cases where the predictor that uses sparse tokens has lower sensitivity, while the bottom row corresponds to the cases where using the frequent tokens leads to lower sensitivity. The precise data settings for this figure, as well as a comparison of sensitivity values, are shown in \cref{tab:sens-synth2}.}
    \label{fig:synth-joint2}
    \end{minipage}   
\end{minipage}
\end{figure*}

%% file: sections/figure_extra.tex
\begin{figure*}[h!]
\centering
    \begin{minipage}[b]{0.43\linewidth}
            \centering
\includegraphics[width=0.9\linewidth]{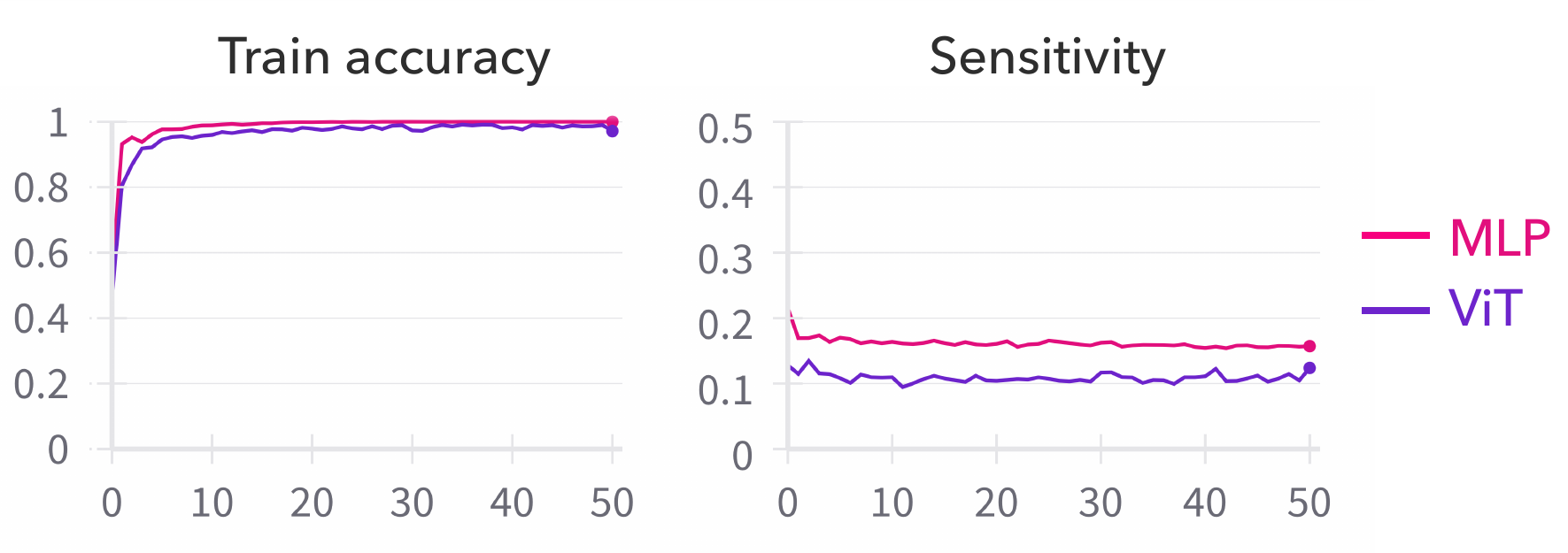}
    \caption{\textbf{Sensitivity on MNIST.} ViT and MLP get similar accuracy, but the ViT has lower sensitivity.}
    \label{fig:mnist}
    \end{minipage}
    \hfill
    \begin{minipage}[b]{0.55\linewidth}
            \centering
\includegraphics[width=0.7\linewidth]{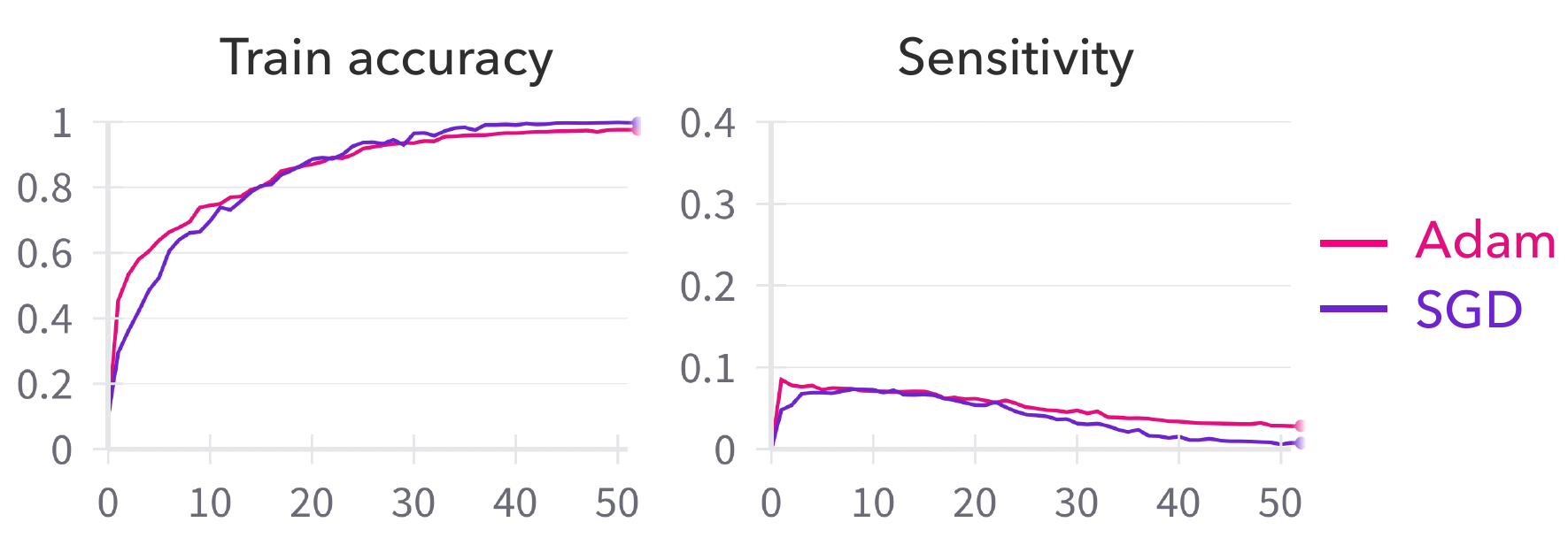}
    \caption{\textbf{Sensitivity using SGD and Adam.} Comparison of train accuracies and sensitivity values of the ViT-small model trained on the CIFAR-10 dataset using SGD and Adam optimizers.}
    \label{fig:sens-opt}
    \end{minipage}
\end{figure*}

%% file: sections/figure_vision.tex
\begin{figure*}[h!]
\centering
    \begin{minipage}[b]{0.54\linewidth}
            \centering
     \includegraphics[width=0.78\linewidth]{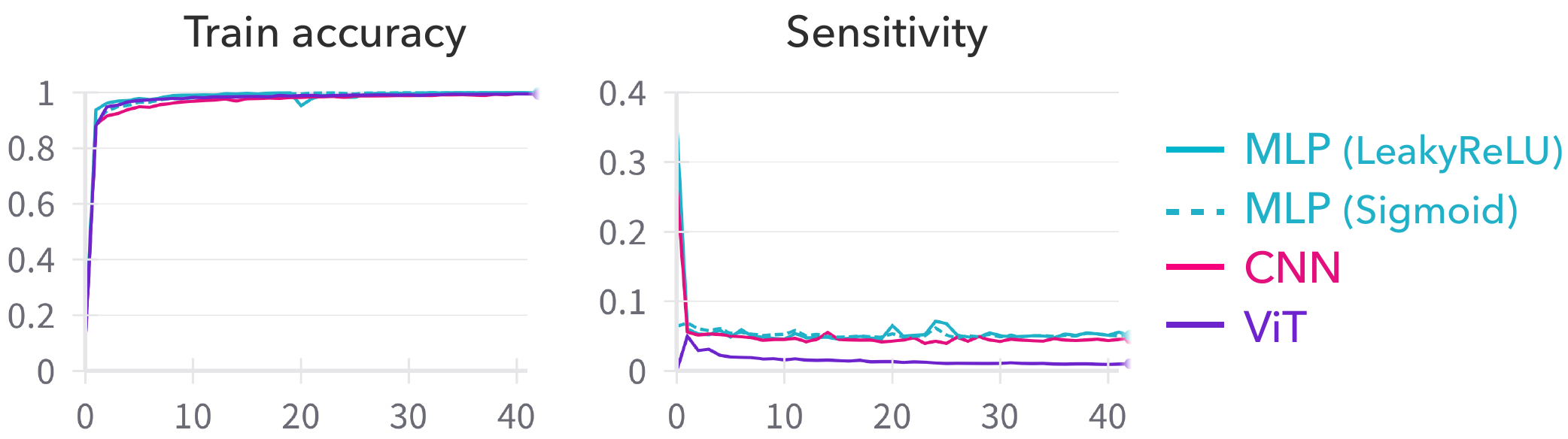}
    \caption{\textbf{Sensitivity on Fashion-MNIST.} Comparison of sensitivity of a ViT with a CNN, an MLP with LeakyReLU activation and an MLP with sigmoid activation, as a function of training epochs. All the models have similar accuracies but the ViT has significantly lower sensitivity.}
    \label{fig:fm}
    \end{minipage}
    \hfill
    \begin{minipage}[b]{0.45\linewidth}
            \centering
     \includegraphics[width=0.78\linewidth]{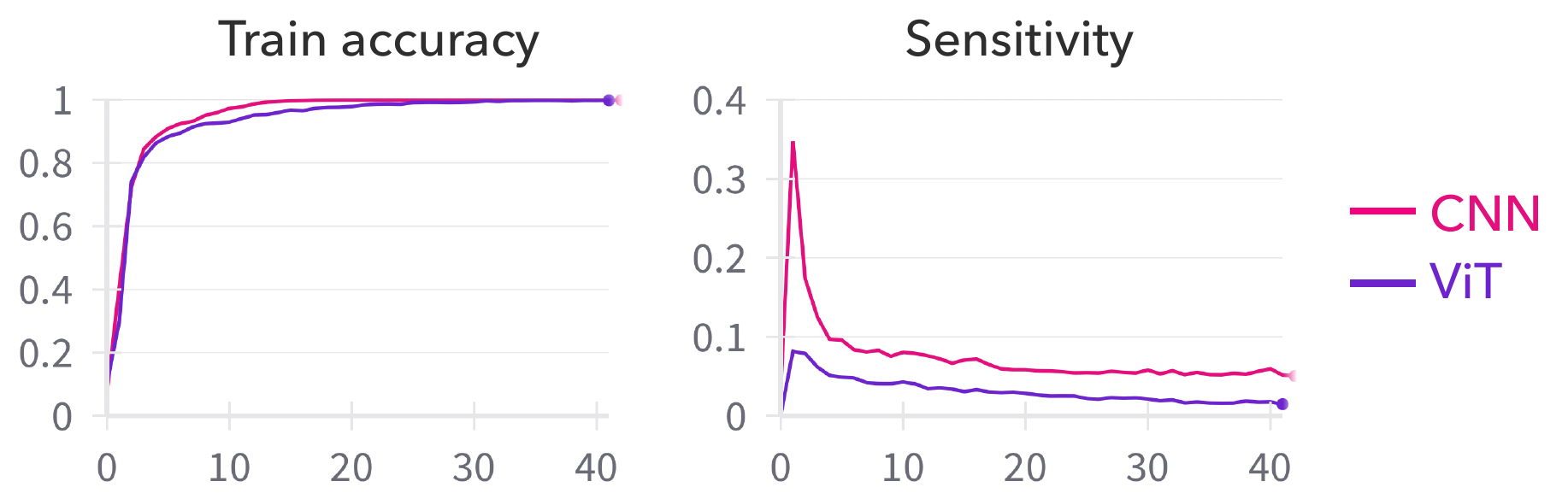}
    \caption{\textbf{Sensitivity on SVHN.} Comparison of sensitivity of a ResNet-18 CNN and a ViT-small trained on SVHN dataset, as a function of training epochs. Both the models have similar accuracies but the ViT has significantly lower sensitivity.}
    \label{fig:svhn}
    \end{minipage}
\end{figure*}

%% file: sections/proof.tex
\section{Proofs for \cref{sec:th-spec}}
\label{app:proof}

We give a brief overview of the CK and NTK here and refer the reader to \citep{lee2018deep,yang2020finegrained} for more details. 

Consider a model with $L$ layers and widths $\{d_l \}_{l=1}^L$ and an input $\x$. Let $g^l(\x)$ denote the output of the $l^\text{th}$ layer scaled by $d_l^{-1/2}$. Suppose we randomly initialize weights from the Gaussian distribution $\calN(0,1)$. It can be shown that in the infinite width limit when $\min_{l\in[L]} d^l \rightarrow \infty$, each element of $g^l(\x)$ is a Gaussian process (GP) with zero mean and kernel function $K^l$. The kernel $K^L$ corresponding to the last layer of the model is the CK. In other words, it is the kernel induced by the embedding $\x \mapsto g^{L-1}(\x)$ when the model is initialized randomly. On the other hand, NTK corresponds to training the entire model instead of just the last layer. Intuitively, when the model parameters $\Tb$ stay close to initialization $\Tb_0$, the residual $g^L(\x; \Tb)-g^L(\x; \Tb_0)$ behaves like a linear model with features given by the gradient at random initialization, $\nabla_\Tb g^L(\x,\Tb_0)$, and the NTK is the kernel of this linear model. The spectra of these kernels provide insights about the implicit prior of a randomly initialized model as well as the implicit bias of training using gradient descent \citep{yang2020finegrained}. The closer these spectra are to the spectrum of the target function, the better we can expect training using gradient descent to generalize.

\subsection{Proof of \cref{corr:wsb}}
\citep{hron2020infinite} show that the self-attention layer 
with linear attention and $d^{-1/2}$ scaling converges in distribution to $\mathcal{GP}(0,K)$ in the infinite width limit, \textit{i.e.} when the number of heads $d^H$ become large. For any layer $l\in [L]$, let $\tilde{K}^l$ denote the kernel induced by the intermediate transformation when applying some nonlinearity $\phi$ to the output of the previous layer $l-1$. Let $f^l_{\cdot j} := \{f^l_{i,j}(\x): \x \in \mathcal{X},i \in [\Tau]\}$, where $\mathcal{X}$ denotes the input space of $\x$. They show the following result for NNs with at least one linear attention layer, in the infinite width limit.
\begin{theorem}[Theorem 3 in \citep{hron2020infinite}]
Let $l\in[L]$, and $\phi$ be such that $|\phi(x)| \leq c + m|x|$ for some $c, m \in \R^+$. Assume
$g^{l-1}$ converges in distribution to $g^{l-1} \sim GP(0,K^{l-1})$,
such that $g_{\cdot j}^{l-1}$ and $g_{\cdot k}^{l-1}$ are independent for any $j \neq k$. Then as $\min\{d^{l,H},d^{l}\} \rightarrow \infty$, $g^l$ converges in distribution to $g^l \sim GP(0,K^l)$ with $g_{\cdot k}^{l}$ and $g_{\cdot \ell}^{l}$ independent for any $k \neq  \ell$, and
\begin{align*}
    K^l (\x,\x') = \E[g^l (\x)g^l (\x')]= \sum\limits_{i,j=1}^{\tilde{d}} (\tilde{K}^l_{ij}(\x,\x'))^2 \tilde{K}^l_{ab} (\x, \x'). 
\end{align*}
\end{theorem}

Similar results are also known for several non-linearities and other layers such as convolutional, dense, average pooling  \citep{lee2018deep, matthews2018gaussian, garriga-alonso2018deep,  novak2019bayesian,yang2021tensor}, as well as residual, positional encoding and layer normalization \citep{hron2020infinite,yang2021tensor}. 

Consequently, any model composed of these layers, such as a transformer with linear attention, also converges to a Gaussian process. This follows using an induction-based argument. It can easily be shown that the induced kernel takes the form 
\begin{align*}
 K(\x,\y)=\Psi\left(\frac{\inpb{\x}{\y}}{\norm{\x}\norm{\y}},\frac{\norm{\x}^2}{d},\frac{\norm{\y}^2}{d}\right),
\end{align*}
for some function $\Psi:\R^3\rightarrow \R$. In addition, since $\x,\y\in \cube{d}$, they have the same norm, and $\Psi$ can be treated as a univariate function that only depends on $c=d^{-1}\inpb{\x}{\y}$, \textit{i.e.} $\Psi\left(c,1,1\right)=\Psi(c)$.

Using this property and the following result, it follows that the kernel induced by a transformer with linear attention is diagonalized by the Fourier basis $\{\chi_U\}_{U\subseteq[d]}$. 

\begin{theorem}[Theorem 3.2 in \citep{yang2020finegrained}]
On the $d$-dimensional boolean cube $\cube{d}$, for every $U\subseteq [d]$, $\chi_U$ is an eigenfunction of $K$ with eigenvalue
\begin{align*}
        \mu_{|U|}:= \mathop{\E}_{\x\sim\cube{d}}\left[x^UK(\x,\mathbf{1})\right]= \mathop{\E}_{\x\sim\cube{d}}\left[x^U \Psi\left(d^{-1}\textstyle\sum\limits_ix_i\right)\right],
    \end{align*}
where $\mathbf{1} := (1,\dots,1) \in \cube{d}$. This definition of $\mu_{|U|}$ does not depend on the choice $S$, only on the cardinality of $S$. These are all of the eigenfunctions of $K$ by dimensionality considerations.
\end{theorem}

Further, using the following result, it follows that transformers (with linear attention) exhibit weak spectral simplicity bias.

\begin{theorem}[Theorem 4.1 in \citep{yang2020finegrained}] Let $K$ be the CK or NTK of an MLP on a boolean cube $\cube{d}$. Then the eigenvalues $\mu_k, k = 0, \dots , d$, satisfy
\begin{align*}
        \mu_0&\geq  \mu_2\geq \dots\geq \mu_{2k} \geq \dots ,\\
\mu_1&\geq  \mu_3\geq  \dots\geq\mu_{2k+1} \geq \dots .
    \end{align*}
\end{theorem}
\subsection{Proof of \cref{prop:ns}}
First, we introduce the concept of noise stability $Q_\rho(f)$, which measures the correlation between the outputs of a function $f$ for $\rho$-correlated pair $(\xb,\xb')$, as $Q_\rho(f):=\underset{(\xb,\xb')_\rho}\E f(\xb)f(\xb')$. Note that $Q_\rho(f)$ is related to $R_\rho(f)\!:=\!\underset{(\xb,\xb')_\rho}{\prob}[f(\xb)\neq f(\xb')]$ as $Q_\rho(f)=1-2R_\rho(f)$ \citep{o'donnell_2014}. 

Using the following result, we can relate noise stability to the Fourier weight of the function $f$ at different degrees $i\in[d]$.
\begin{theorem}[Theorem 2.49 in \citep{o'donnell_2014}] For function $f\!:\!\cube{d}\!\rightarrow\!\{\pm 1\}$, the noise stability for $\rho$-correlated pair $(\x,\x')$ satisfies $Q_\rho(f)\!=\!\sum_{U\subseteq [d]}\rho^{|U|}\hat{f}(U)^2\!=\!\sum_{i=0}^d\rho^iW^i[f]$, where $W^i[f]\!:=\!\!\sum_{U\subseteq[d],|U|=i}\!\!\hat{f}(U)^2$.
\end{theorem}
Clearly, $Q_\rho(f)\leq 1$ since the minimum degree of $f$ is $0$. Next, we use the following important result, which upper bounds the degree of $f$ in terms of its maximum sensitivity, $S_{\max}(f)\!:=\!\max_{\x\in\cube{d}}S(f,\x)$.
\begin{theorem}[Theorem 1.4 in \citep{huang2019induced}] For function $f\!:\!\cube{d}\!\rightarrow\! \{\pm 1\}$, the degree $D(f)$ of the multilinear polynomial which represents $f$ satisfies $D(f)\!\leq\! (S_{\max}(f))^2$.
\end{theorem}
Using this, and the fact that $Q_\rho(f)$ is minimized when the Fourier weight is concentrated on the highest degree term, we get the lower bound $Q_\rho(f)\geq \rho^{(S_{\max}(f))^2}$, since $\rho<1$ and $S_{\max}(f)\in [1,d]$.

Using the relation between $Q_\rho(f)$ and $R_\rho(f)$ then finishes the proof.

%% file: sections/rel-work-app.tex
\section{Related Work}
\label{app:rel-work}
\paragraph{Implicit Biases of Gradient Methods.} Several works study the implicit bias of gradient-based methods for linear predictors and MLPs. Pioneering work by \citet{soudry2018implicit,ji2018risk} revealed that linear models trained with gradient descent to minimize an exponentially-tailed loss on linearly separable data converge (in direction) to the max-margin classifier. Following this, \citet{nacson2019convergence, telgarskyngd2021, tel2021dualacc} derived fast convergence rates for gradient-based methods in this setting. Recent works show that MLPs trained with gradient flow/descent converge to a KKT point of the corresponding max-margin problem in the parameter space, in both finite \citep{tel2020directional, lyu2020Gradient} and infinite width \citep{chizat2020implicit} regimes. Further, \citet{phuongortho-separable2021, frei2022implicit, qqgimplicit2023} have also studied ReLU/Leaky-ReLU networks trained with gradient descent on nearly orthogonal data. \citet{li2022implicit} show that the training path in over-parameterized models can be interpreted as mirror descent applied to an alternative objective.
In regression problems, when minimizing the mean squared error, the bias manifests in the form of rank minimization \citep{aroraimplicitdeep2019,li2021resolving-deepmatrix-ib}. Additionally, the implicit bias of other optimization algorithms, such as stochastic gradient descent and adaptive methods, has also been explored in various studies \citep{blanc2020implicit,haochen2021shape}; see the recent survey \citep{vardi2022implicit} for a detailed summary. 

\paragraph{Robustness.}
Several research efforts have been made to investigate the robustness of Transformers. \citet{shao2021adversarial} showed that Transformers exhibit greater resistance to adversarial attacks compared to other models. Additionally, \citet{mahmood2021robustness} highlighted the notably low transferability of adversarial examples between CNNs and ViTs. Subsequent research \citep{shen2023improving, bhojanapalli2021understanding,paul2022vision} expanded this robustness examination to improve transformer-based language models.
 \citet{shi2020robustness} introduced the concept of robustness verification in Transformers. Various robust training methods have been suggested to enhance the robustness guarantees of models, often influenced by or stemming from their respective verification techniques. \citet{shi2021fast} expedited the certified robust training process through the use of interval-bound propagation. \citet{wang2021macrobert} employed randomized smoothing to train BERT, aiming to maximize its certified robust space. Recent work of \dfedit{\citet{bombari2024understanding} shows that randomly-initialized attention layers tend to have higher word-level sensitivity than fully connected layers.} \bvedit{In contrast to our work, they consider word sensitivity, which has been experimentally shown to be similar for transformers and LSTMs \citep{bhattamishra2023simplicity}.}

\paragraph{Spurious Correlations.} 
A common pitfall to the generalization of neural networks is the presence of spurious correlations \citep{Sagawa2020Distributionally}. 
For example, \citet{geirhos2018imagenettrained} observed that trained CNNs are biased towards textures rather than shapes to make predictions for object recognition tasks. Such biases make NNs vulnerable to adversarial attacks. \citet{gururangan-etal-2018-annotation} attribute the reliance of NNs on spurious features to confounding factors in data collection while \citet{shah2020pitfalls} attribute it 
to a \textit{simplicity bias}. Several works have studied the underlying causes of simplicity bias \citep{chiang2021benefit,nagarajan2021understanding,morwani2023simplicity,Huh2021TheLS,Lyu2021GradientDO} and multiple methods have been developed to mitigate this bias and improve generalization \citep{Pezeshki2020GradientSA,Kirichenko2022LastLR,vasudeva2023mitigating,tiwari2023overcoming}. 

\paragraph{Data Augmentation.} The essence of data augmentation is to impose some notion of regularization. The simplest design of data augmentation dates back to \citet{Robbins1951ASA} where image manipulation, e.g., flip, crop, and rotate, was introduced. \citet{Bishop1995TrainingWN} proved that training with Gaussian noise is equivalent to Tikhonov regularization. We also note this observation is in parallel to our proposition in \Cref{sec:implications} that training with Gaussian noise promotes low sensitivity. Recently, mixup-based augmentation methods have been proposed to improve model robustness by merging two images as well as their labels \citep{zhang2018mixup}. Several works also use a combination of existing augmentation techniques \citep{Cubuk2019AutoAugmentLA,Lim2019FastA}.  A common belief is that data augmentation can improve model robustness \citep{Rebuffi2021DataAC}, and this work bridges the method (augmentation) and the outcome (robustness) with an explanation --- simplicity bias towards low sensitivity. 

%% file: sections/exp-settings.tex
\section{Details of Experimental Settings}
\label{app:expt-details}
We use PyTorch \citep{paszke2019pytorch} as our code framework and as our implementation of LSTMs. PyTorch is licensed under the Modified BSD license.

\paragraph{Experimental Settings for Synthetic Data Experiments.}
We use standard SGD training with batch size $100$. We consider $\Tau=50$ and train with $1000$ samples and test on $500$ samples generated as per \cref{def:gen_data}. 
\paragraph{Datasets, Model Architectures and Experimental Settings for Vision Tasks.} We consider the following datasets:

\noindent \emph{Fashion-MNIST.} Fashion-MNIST \citep{fm} consists of $28\times 28$ grayscale images of Zalando's articles. This is a $10$-class classification task with $60k$ training and $10k$ test images. It is released under the MIT license. \\

\noindent \emph{CIFAR-10.} The CIFAR-10 dataset \citep{cifar} is a well-known object recognition dataset. It consists of $32\times32$ color images in $10$ classes, with $6k$ images per class. There are $50k$ training and $10k$ test images. It is released under the MIT license.\\

\noindent \emph{SVHN.} Street View House Numbers (SVHN) \citep{svhn} is a real-world image dataset used as a digit classification benchmark. It contains $32\times32$ RGB images of printed digits ($0$ to $9$) cropped from Google Street View images of house number plates. 
There are $60k$ images in the train set and $10k$ images in the test set. It is released under the CC BY 4.0 license.\\

\noindent \emph{ImageNet-1k.} The ImageNet-1k dataset \citep{imagenet15russakovsky}, also known as the ILSVRC (ImageNet Large Scale Visual Recognition Challenge) dataset, is a widely used benchmark dataset in computer vision for tasks such as image classification, object detection, and image segmentation. There are $1000$ different classes, and approximately $1.28$ million training images of size $224\times 224$. It is released under a non-commercial research use license. \\

For all the datasets, we use the ViT-small architecture implementation available at \url{https://github.com/lucidrains/vit-pytorch}. For the ResNet-18 model used in the experiments on CIFAR-10 and SVHN datasets, we use the implementation available at \url{https://github.com/kuangliu/pytorch-cifar}. Additionally, for the DenseNet-121 model, ConvMixer model and ViT-simple model used in the experiments on CIFAR-10, we use the implementations available at \url{https://github.com/huyvnphan/PyTorch_CIFAR10}, \url{https://github.com/locuslab/convmixer} and \url{https://github.com/lucidrains/vit-pytorch}, respectively. All of these models are released under the MIT license. 

All the models are trained with SGD using batch size $50$ for MNIST and $100$ for the other datasets. We use patch size $7$ for MNIST and $4$ for the other datasets. We estimate the expectation over $\calP$ in \cref{def:sensitivity} by replacing every patch with a noisy patch $5$ times, and sample about $30\%$ of the training data to evaluate sensitivity.

For the MNIST experiments, we consider a 1-hidden-layer MLP with $100$ hidden units and LeakyReLU activation. We set depth as $2$, number of heads as $1$ and the hidden units in the MLP as $128$ for the ViT. We train both models with a learning rate of $0.01$. 

For Fashion-MNIST, we set depth as $2$, number of heads as $8$ and the hidden units in the MLP as $256$ for the ViT. We consider a 2-hidden layer MLP with $512$ and $128$ hidden units, respectively. The CNN consists of two 2D convolutional layers with $32$ output channels and kernel size $3$ followed by a 2D MaxPool layer with both kernel size and stride as $2$ and two fully connected layers with $128$ hidden units. We use LeakyReLU activation for the CNN. 
We use learning rates of $0.1$ for the MLP with LeakyReLU, $0.5$ for the MLP with sigmoid, $0.005$ for the CNN and $0.1$ for the ViT. 

For CIFAR-10, we set depth as $8$, number of heads as $32$ and the hidden units in the MLP as $256$ for the ViT-small model, while these values are set as $6$, $16$, and $512$ for the ViT-simple model. For the ConvMixer model, we use depth $6$, embedding dimension $128$ and kernel size $3$. The learning rate is set as $0.1$ for ViT-small, $0.2$ for ViT-simple, $0.06$ for ConvMixer, $0.001$ for ResNet-18 and $0.005$ for DenseNet-121. 

For SVHN, most of the settings are the same as the CIFAR-10 experiments, except we set the hidden units in the MLP as $512$ for the ViT-small model and the learning rate is set as $0.0015$ for ResNet-18. 

\dfedit{ 
For ImageNet-1k, we sample $20k$ samples from the training set and compare the sensitivity values of pre-trained ConvNext (\texttt{ConvNextV2-Tiny}) and ViT/L-16 with $\sigma^2=15$. 
Both achieved the same training and validation accuracy of $85\%$, and it ensures our sensitivity comparison is fair. 
}

\paragraph{Datasets, Model Architectures and Experimental Settings for Language Tasks.} We \bvedit{consider the following} two binary classification datasets, 
which are relatively easy to learn without pretraining \citep{kovaleva-etal-2019-revealing}.\\ 

\noindent \emph{MRPC.} Microsoft Research Paraphrase Corpus (MRPC) \citep{dolan-brockett-2005-automatically} is a corpus that consists of $5801$ sentence pairs. Each pair is labeled if it is a paraphrase or not by human annotators. It has $4076$ training examples and $1725$ validation examples. It is released under the ODC-By or the Microsoft Research license.\\

\noindent\emph{QQP.} Quora Question Pairs (QQP) \citep{qqp-data} dataset is a corpus that consists of over $400k$ question pairs. Each question pair is annotated with a binary value indicating whether the two questions are paraphrases of each other. It has $364k$ training examples and $40k$ validation examples. It is released under the CC BY-SA 2.5 license.

  For both RoBERTa models and LSTM models, we keep the same number of layers: $4$ layers. We set number of heads as $8$ RoBERTa. We use the AdamW optimizer with a learning rate of $0.0001$ and weight decay of $0.0001$ for all the tasks. We also use a dropout rate of $0.1$. We use a batch size of $32$ for all the experiments. The used RoBERTa model is released on Huggingface \href{https://huggingface.co/FacebookAI/roberta-base}{https://huggingface.co/FacebookAI/roberta-base} with MIT license.

\paragraph{Experimental Settings for \cref{sec:implications}.} We set the learning rate as $0.16$ and $0.2$ when training the ViT-small with regularization and augmentation, respectively. We use a regularization strength of $0.25$. The remaining settings are the same as for the other experiments. For computing the sharpness metrics, we approximate the expectation over the Gaussian noise by averaging over $5$ repeats and set $\sigma$ as $0.005$. 

\paragraph{Experimental Settings for \cref{app:add-expts}.} For the experiment with the Adam optimizer, we employ a learning rate scheduler to ensure that the accuracy on the train set is similar to the model trained with SGD. The initial learning rate is $0.002$ and after every $8$ epochs, it is scaled by a factor of $0.5$.

For the remaining experiments in this section, we consider the same settings as for the respective main experiment.

\paragraph{Compute Details.} 
Experiments with synthetic data were run on Google Colab. Experiments on vision and language tasks were run on internal clusters using NVIDIA RTX A6000 GPUs with 48GB of VRAM. For the experiments on vision data, we use two GPUs and the runtime for each setting is about $17$ hours. Experiments on language tasks use one GPU and the runtime for each experiment is about $24$ hours.

%% file: sections/limitations.tex
\section{Limitations}
\label{sec:limitations}

In our theoretical results, we show that transformers exhibit weak spectral bias, similar to other NN architectures. An important direction for future work is to distinguish transformers from other architectures and show that they exhibit a stronger spectral bias.

Additionally, this work focuses on the inductive bias of the transformer architecture. However, other factors such as the data used while pre-training can also effect the biases these models exhibit on downstream tasks. It would be interesting to explore this effect in the future.

Similarly, the choice of the optimization algorithm used for training can also have an effect. In our experiments on the CIFAR-10 dataset (\cref{fig:sens-opt}), we see that SGD and Adam are very similar. However, conducting a more thorough comparison, e.g., by considering second-order optimization methods, can be an important direction for future work.

%% file: main.bbl
\begin{thebibliography}{124}
\providecommand{\natexlab}[1]{#1}
\providecommand{\url}[1]{\texttt{#1}}
\expandafter\ifx\csname urlstyle\endcsname\relax
  \providecommand{\doi}[1]{doi: #1}\else
  \providecommand{\doi}{doi: \begingroup \urlstyle{rm}\Url}\fi

\bibitem[Aky{\"u}rek et~al.(2023)Aky{\"u}rek, Schuurmans, Andreas, Ma, and Zhou]{Akyrek2022WhatLA}
Ekin Aky{\"u}rek, Dale Schuurmans, Jacob Andreas, Tengyu Ma, and Denny Zhou.
\newblock What learning algorithm is in-context learning? investigations with linear models.
\newblock In \emph{The Eleventh International Conference on Learning Representations}, 2023.

\bibitem[Andriushchenko et~al.(2023)Andriushchenko, Croce, Mueller, Hein, and Flammarion]{Andriushchenko2023AML}
Maksym Andriushchenko, Francesco Croce, Maximilian Mueller, Matthias Hein, and Nicolas Flammarion.
\newblock A modern look at the relationship between sharpness and generalization.
\newblock In \emph{International Conference on Machine Learning}, 2023.
\newblock URL \url{https://api.semanticscholar.org/CorpusID:256846369}.

\bibitem[Arora et~al.(2019)Arora, Cohen, Hu, and Luo]{aroraimplicitdeep2019}
Sanjeev Arora, Nadav Cohen, Wei Hu, and Yuping Luo.
\newblock Implicit regularization in deep matrix factorization.
\newblock In \emph{Advances in Neural Information Processing Systems}, volume~32, 2019.

\bibitem[Arpit et~al.(2017)Arpit, Jastrz{\k{e}}bski, Ballas, Krueger, Bengio, Kanwal, Maharaj, Fischer, Courville, Bengio, et~al.]{arpit2017closer}
Devansh Arpit, Stanis{\l}aw Jastrz{\k{e}}bski, Nicolas Ballas, David Krueger, Emmanuel Bengio, Maxinder~S Kanwal, Tegan Maharaj, Asja Fischer, Aaron Courville, Yoshua Bengio, et~al.
\newblock A closer look at memorization in deep networks.
\newblock In \emph{International conference on machine learning}, pp.\  233--242. PMLR, 2017.

\bibitem[Ba et~al.(2016)Ba, Kiros, and Hinton]{ba2016layer}
Jimmy~Lei Ba, Jamie~Ryan Kiros, and Geoffrey~E. Hinton.
\newblock Layer normalization, 2016.

\bibitem[Bai et~al.(2023)Bai, Chen, Wang, Xiong, and Mei]{bai2023transformers}
Yu~Bai, Fan Chen, Huan Wang, Caiming Xiong, and Song Mei.
\newblock Transformers as statisticians: Provable in-context learning with in-context algorithm selection.
\newblock \emph{arXiv preprint arXiv:2306.04637}, 2023.

\bibitem[Barak et~al.(2022)Barak, Edelman, Goel, Kakade, Malach, and Zhang]{barak2022hidden}
Boaz Barak, Benjamin Edelman, Surbhi Goel, Sham Kakade, Eran Malach, and Cyril Zhang.
\newblock Hidden progress in deep learning: Sgd learns parities near the computational limit.
\newblock \emph{Advances in Neural Information Processing Systems}, 35:\penalty0 21750--21764, 2022.

\bibitem[Basri et~al.(2019)Basri, Jacobs, Kasten, and Kritchman]{basri2019convergence}
Ronen Basri, David Jacobs, Yoni Kasten, and Shira Kritchman.
\newblock The convergence rate of neural networks for learned functions of different frequencies, 2019.

\bibitem[Beyer et~al.(2022)Beyer, Zhai, and Kolesnikov]{beyer2022better}
Lucas Beyer, Xiaohua Zhai, and Alexander Kolesnikov.
\newblock Better plain vit baselines for imagenet-1k, 2022.

\bibitem[Bhattamishra et~al.(2023{\natexlab{a}})Bhattamishra, Patel, Blunsom, and Kanade]{bhattamishra2023understanding}
Satwik Bhattamishra, Arkil Patel, Phil Blunsom, and Varun Kanade.
\newblock Understanding in-context learning in transformers and llms by learning to learn discrete functions, 2023{\natexlab{a}}.

\bibitem[Bhattamishra et~al.(2023{\natexlab{b}})Bhattamishra, Patel, Kanade, and Blunsom]{bhattamishra2023simplicity}
Satwik Bhattamishra, Arkil Patel, Varun Kanade, and Phil Blunsom.
\newblock Simplicity bias in transformers and their ability to learn sparse {B}oolean functions.
\newblock In \emph{Proceedings of the 61st Annual Meeting of the Association for Computational Linguistics}, 2023{\natexlab{b}}.

\bibitem[Bhojanapalli et~al.(2021)Bhojanapalli, Chakrabarti, Glasner, Li, Unterthiner, and Veit]{bhojanapalli2021understanding}
Srinadh Bhojanapalli, Ayan Chakrabarti, Daniel Glasner, Daliang Li, Thomas Unterthiner, and Andreas Veit.
\newblock Understanding robustness of transformers for image classification.
\newblock In \emph{Proceedings of the IEEE/CVF international conference on computer vision}, pp.\  10231--10241, 2021.

\bibitem[Bietti \& Mairal(2019)Bietti and Mairal]{bietti-ntk-2019}
Alberto Bietti and Julien Mairal.
\newblock \emph{On the inductive bias of neural tangent kernels}.
\newblock Curran Associates Inc., Red Hook, NY, USA, 2019.

\bibitem[Bishop(1995)]{Bishop1995TrainingWN}
Christopher~M. Bishop.
\newblock Training with noise is equivalent to tikhonov regularization.
\newblock \emph{Neural Computation}, 7:\penalty0 108--116, 1995.

\bibitem[Blanc et~al.(2020)Blanc, Gupta, Valiant, and Valiant]{blanc2020implicit}
Guy Blanc, Neha Gupta, Gregory Valiant, and Paul Valiant.
\newblock Implicit regularization for deep neural networks driven by an ornstein-uhlenbeck like process.
\newblock In \emph{Conference on learning theory}, pp.\  483--513. PMLR, 2020.

\bibitem[Bombari \& Mondelli(2024)Bombari and Mondelli]{bombari2024understanding}
Simone Bombari and Marco Mondelli.
\newblock Towards understanding the word sensitivity of attention layers: A study via random features, 2024.

\bibitem[Brown et~al.(2020)Brown, Mann, Ryder, Subbiah, Kaplan, Dhariwal, Neelakantan, Shyam, Sastry, Askell, Agarwal, Herbert-Voss, Krueger, Henighan, Child, Ramesh, Ziegler, Wu, Winter, Hesse, Chen, Sigler, Litwin, Gray, Chess, Clark, Berner, McCandlish, Radford, Sutskever, and Amodei]{brown2020lm}
Tom Brown, Benjamin Mann, Nick Ryder, Melanie Subbiah, Jared~D Kaplan, Prafulla Dhariwal, Arvind Neelakantan, Pranav Shyam, Girish Sastry, Amanda Askell, Sandhini Agarwal, Ariel Herbert-Voss, Gretchen Krueger, Tom Henighan, Rewon Child, Aditya Ramesh, Daniel Ziegler, Jeffrey Wu, Clemens Winter, Chris Hesse, Mark Chen, Eric Sigler, Mateusz Litwin, Scott Gray, Benjamin Chess, Jack Clark, Christopher Berner, Sam McCandlish, Alec Radford, Ilya Sutskever, and Dario Amodei.
\newblock Language models are few-shot learners.
\newblock In H.~Larochelle, M.~Ranzato, R.~Hadsell, M.F. Balcan, and H.~Lin (eds.), \emph{Advances in Neural Information Processing Systems}, volume~33, pp.\  1877--1901. Curran Associates, Inc., 2020.

\bibitem[Cao et~al.(2021)Cao, Fang, Wu, Zhou, and Gu]{cao2021towards}
Yuan Cao, Zhiying Fang, Yue Wu, Ding-Xuan Zhou, and Quanquan Gu.
\newblock Towards understanding the spectral bias of deep learning.
\newblock In \emph{IJCAI}, 2021.

\bibitem[Chen et~al.(2024)Chen, Shwartz-Ziv, Cho, Leavitt, and Saphra]{chen2024sudden}
Angelica Chen, Ravid Shwartz-Ziv, Kyunghyun Cho, Matthew~L Leavitt, and Naomi Saphra.
\newblock Sudden drops in the loss: Syntax acquisition, phase transitions, and simplicity bias in {MLM}s.
\newblock In \emph{The Twelfth International Conference on Learning Representations}, 2024.
\newblock URL \url{https://openreview.net/forum?id=MO5PiKHELW}.

\bibitem[Chiang(2021)]{chiang2021benefit}
Ting-Rui Chiang.
\newblock On a benefit of mask language modeling: Robustness to simplicity bias, 2021.

\bibitem[Chizat \& Bach(2020)Chizat and Bach]{chizat2020implicit}
Lenaic Chizat and Francis Bach.
\newblock Implicit bias of gradient descent for wide two-layer neural networks trained with the logistic loss.
\newblock In \emph{Conference on Learning Theory}, pp.\  1305--1338. PMLR, 2020.

\bibitem[Conmy et~al.(2023)Conmy, Mavor-Parker, Lynch, Heimersheim, and Garriga-Alonso]{Conmy2023TowardsAC}
Arthur Conmy, Augustine~N. Mavor-Parker, Aengus Lynch, Stefan Heimersheim, and Adri{\`a} Garriga-Alonso.
\newblock Towards automated circuit discovery for mechanistic interpretability.
\newblock In \emph{Neural Information Processing Systems}, 2023.

\bibitem[Cubuk et~al.(2019)Cubuk, Zoph, Man{\'e}, Vasudevan, and Le]{Cubuk2019AutoAugmentLA}
Ekin~Dogus Cubuk, Barret Zoph, Dandelion Man{\'e}, Vijay Vasudevan, and Quoc~V. Le.
\newblock Autoaugment: Learning augmentation strategies from data.
\newblock \emph{2019 IEEE/CVF Conference on Computer Vision and Pattern Recognition (CVPR)}, pp.\  113--123, 2019.

\bibitem[de~G.~Matthews et~al.(2018)de~G.~Matthews, Rowland, Hron, Turner, and Ghahramani]{matthews2018gaussian}
Alexander~G. de~G.~Matthews, Mark Rowland, Jiri Hron, Richard~E. Turner, and Zoubin Ghahramani.
\newblock Gaussian process behaviour in wide deep neural networks, 2018.

\bibitem[Dolan \& Brockett(2005)Dolan and Brockett]{dolan-brockett-2005-automatically}
William~B. Dolan and Chris Brockett.
\newblock Automatically constructing a corpus of sentential paraphrases.
\newblock In \emph{Proceedings of the Third International Workshop on Paraphrasing ({IWP}2005)}, 2005.
\newblock URL \url{https://aclanthology.org/I05-5002}.

\bibitem[Dosovitskiy et~al.(2021)Dosovitskiy, Beyer, Kolesnikov, Weissenborn, Zhai, Unterthiner, Dehghani, Minderer, Heigold, Gelly, Uszkoreit, and Houlsby]{dosovitskiy2021an}
Alexey Dosovitskiy, Lucas Beyer, Alexander Kolesnikov, Dirk Weissenborn, Xiaohua Zhai, Thomas Unterthiner, Mostafa Dehghani, Matthias Minderer, Georg Heigold, Sylvain Gelly, Jakob Uszkoreit, and Neil Houlsby.
\newblock An image is worth 16x16 words: Transformers for image recognition at scale.
\newblock In \emph{International Conference on Learning Representations}, 2021.
\newblock URL \url{https://openreview.net/forum?id=YicbFdNTTy}.

\bibitem[Frei et~al.(2022)Frei, Vardi, Bartlett, Srebro, and Hu]{frei2022implicit}
Spencer Frei, Gal Vardi, Peter~L Bartlett, Nathan Srebro, and Wei Hu.
\newblock Implicit bias in leaky relu networks trained on high-dimensional data.
\newblock \emph{arXiv preprint arXiv:2210.07082}, 2022.

\bibitem[Fu et~al.(2023)Fu, Chen, Jia, and Sharan]{fu2023transformers}
Deqing Fu, Tian-Qi Chen, Robin Jia, and Vatsal Sharan.
\newblock Transformers learn higher-order optimization methods for in-context learning: A study with linear models.
\newblock \emph{arXiv}, abs/2310.17086, 2023.

\bibitem[Garg et~al.(2022)Garg, Tsipras, Liang, and Valiant]{Garg2022WhatCT}
Shivam Garg, Dimitris Tsipras, Percy Liang, and Gregory Valiant.
\newblock What can transformers learn in-context? a case study of simple function classes.
\newblock \emph{ArXiv}, abs/2208.01066, 2022.

\bibitem[Garriga-Alonso et~al.(2019)Garriga-Alonso, Rasmussen, and Aitchison]{garriga-alonso2018deep}
Adrià Garriga-Alonso, Carl~Edward Rasmussen, and Laurence Aitchison.
\newblock Deep convolutional networks as shallow gaussian processes.
\newblock In \emph{International Conference on Learning Representations}, 2019.
\newblock URL \url{https://openreview.net/forum?id=Bklfsi0cKm}.

\bibitem[Geirhos et~al.(2019)Geirhos, Rubisch, Michaelis, Bethge, Wichmann, and Brendel]{geirhos2018imagenettrained}
Robert Geirhos, Patricia Rubisch, Claudio Michaelis, Matthias Bethge, Felix~A. Wichmann, and Wieland Brendel.
\newblock Imagenet-trained {CNN}s are biased towards texture; increasing shape bias improves accuracy and robustness.
\newblock In \emph{International Conference on Learning Representations}, 2019.
\newblock URL \url{https://openreview.net/forum?id=Bygh9j09KX}.

\bibitem[Geirhos et~al.(2020)Geirhos, Jacobsen, Michaelis, Zemel, Brendel, Bethge, and Wichmann]{geirhos2020shortcut}
Robert Geirhos, J{\"o}rn-Henrik Jacobsen, Claudio Michaelis, Richard Zemel, Wieland Brendel, Matthias Bethge, and Felix~A Wichmann.
\newblock Shortcut learning in deep neural networks.
\newblock \emph{Nature Machine Intelligence}, 2\penalty0 (11):\penalty0 665--673, 2020.

\bibitem[Ghosal et~al.(2022)Ghosal, Ming, and Li]{ghosal2022vision}
Soumya~Suvra Ghosal, Yifei Ming, and Yixuan Li.
\newblock Are vision transformers robust to spurious correlations?, 2022.

\bibitem[Gopalan et~al.(2016)Gopalan, Nisan, Servedio, Talwar, and Wigderson]{gopalan2016smooth}
Parikshit Gopalan, Noam Nisan, Rocco~A Servedio, Kunal Talwar, and Avi Wigderson.
\newblock Smooth boolean functions are easy: Efficient algorithms for low-sensitivity functions.
\newblock In \emph{Proceedings of the 2016 ACM Conference on Innovations in Theoretical Computer Science}, pp.\  59--70, 2016.

\bibitem[Guo et~al.(2024)Guo, Hu, Mei, Wang, Xiong, Savarese, and Bai]{guo2024how}
Tianyu Guo, Wei Hu, Song Mei, Huan Wang, Caiming Xiong, Silvio Savarese, and Yu~Bai.
\newblock How do transformers learn in-context beyond simple functions? a case study on learning with representations.
\newblock In \emph{The Twelfth International Conference on Learning Representations}, 2024.
\newblock URL \url{https://openreview.net/forum?id=ikwEDva1JZ}.

\bibitem[Gururangan et~al.(2018)Gururangan, Swayamdipta, Levy, Schwartz, Bowman, and Smith]{gururangan-etal-2018-annotation}
Suchin Gururangan, Swabha Swayamdipta, Omer Levy, Roy Schwartz, Samuel Bowman, and Noah~A. Smith.
\newblock Annotation artifacts in natural language inference data.
\newblock In Marilyn Walker, Heng Ji, and Amanda Stent (eds.), \emph{Proceedings of the 2018 Conference of the North {A}merican Chapter of the Association for Computational Linguistics: Human Language Technologies, Volume 2 (Short Papers)}, pp.\  107--112, New Orleans, Louisiana, June 2018. Association for Computational Linguistics.
\newblock \doi{10.18653/v1/N18-2017}.
\newblock URL \url{https://aclanthology.org/N18-2017}.

\bibitem[Hahn \& Rofin(2024)Hahn and Rofin]{hahn-rofin-2024-sensitive}
Michael Hahn and Mark Rofin.
\newblock Why are sensitive functions hard for transformers?
\newblock In Lun-Wei Ku, Andre Martins, and Vivek Srikumar (eds.), \emph{Proceedings of the 62nd Annual Meeting of the Association for Computational Linguistics (Volume 1: Long Papers)}, pp.\  14973--15008, Bangkok, Thailand, August 2024. Association for Computational Linguistics.
\newblock \doi{10.18653/v1/2024.acl-long.800}.
\newblock URL \url{https://aclanthology.org/2024.acl-long.800}.

\bibitem[Hanna et~al.(2023)Hanna, Liu, and Variengien]{hanna2023how}
Michael Hanna, Ollie Liu, and Alexandre Variengien.
\newblock How does {GPT}-2 compute greater-than?: Interpreting mathematical abilities in a pre-trained language model.
\newblock In \emph{Thirty-seventh Conference on Neural Information Processing Systems}, 2023.
\newblock URL \url{https://openreview.net/forum?id=p4PckNQR8k}.

\bibitem[HaoChen et~al.(2021)HaoChen, Wei, Lee, and Ma]{haochen2021shape}
Jeff~Z HaoChen, Colin Wei, Jason Lee, and Tengyu Ma.
\newblock Shape matters: Understanding the implicit bias of the noise covariance.
\newblock In \emph{Conference on Learning Theory}, pp.\  2315--2357. PMLR, 2021.

\bibitem[Hassid et~al.(2022)Hassid, Peng, Rotem, Kasai, Montero, Smith, and Schwartz]{hassid2022does}
Michael Hassid, Hao Peng, Daniel Rotem, Jungo Kasai, Ivan Montero, Noah~A. Smith, and Roy Schwartz.
\newblock How much does attention actually attend? questioning the importance of attention in pretrained transformers, 2022.

\bibitem[Hatami et~al.(2011)Hatami, Kulkarni, and Pankratov]{hatami2011variations}
Pooya Hatami, Raghav Kulkarni, and Denis Pankratov.
\newblock Variations on the sensitivity conjecture.
\newblock \emph{Theory of Computing}, pp.\  1--27, 2011.

\bibitem[He et~al.(2016)He, Zhang, Ren, and Sun]{he2016residual}
Kaiming He, Xiangyu Zhang, Shaoqing Ren, and Jian Sun.
\newblock {Deep Residual Learning for Image Recognition}.
\newblock In \emph{Proceedings of 2016 IEEE Conference on Computer Vision and Pattern Recognition}, CVPR '16, pp.\  770--778. IEEE, June 2016.
\newblock \doi{10.1109/CVPR.2016.90}.
\newblock URL \url{http://ieeexplore.ieee.org/document/7780459}.

\bibitem[Hendrycks \& Dietterich(2019)Hendrycks and Dietterich]{hendrycks2018benchmarking}
Dan Hendrycks and Thomas Dietterich.
\newblock Benchmarking neural network robustness to common corruptions and perturbations.
\newblock In \emph{International Conference on Learning Representations}, 2019.
\newblock URL \url{https://openreview.net/forum?id=HJz6tiCqYm}.

\bibitem[Hochreiter \& Schmidhuber(1997)Hochreiter and Schmidhuber]{hochreiter1997lstm}
Sepp Hochreiter and Jürgen Schmidhuber.
\newblock {Long Short-Term Memory}.
\newblock \emph{Neural Computation}, 9\penalty0 (8):\penalty0 1735--1780, 11 1997.
\newblock ISSN 0899-7667.
\newblock \doi{10.1162/neco.1997.9.8.1735}.
\newblock URL \url{https://doi.org/10.1162/neco.1997.9.8.1735}.

\bibitem[Hron et~al.(2020)Hron, Bahri, Sohl-Dickstein, and Novak]{hron2020infinite}
Jiri Hron, Yasaman Bahri, Jascha Sohl-Dickstein, and Roman Novak.
\newblock Infinite attention: Nngp and ntk for deep attention networks.
\newblock In \emph{Proceedings of the 37th International Conference on Machine Learning}, ICML'20. JMLR.org, 2020.

\bibitem[Huang et~al.(2016)Huang, Liu, and Weinberger]{Huang2016DenselyCC}
Gao Huang, Zhuang Liu, and Kilian~Q. Weinberger.
\newblock Densely connected convolutional networks.
\newblock \emph{2017 IEEE Conference on Computer Vision and Pattern Recognition (CVPR)}, pp.\  2261--2269, 2016.
\newblock URL \url{https://api.semanticscholar.org/CorpusID:9433631}.

\bibitem[Huang(2019)]{huang2019induced}
Hao Huang.
\newblock Induced subgraphs of hypercubes and a proof of the sensitivity conjecture.
\newblock \emph{Annals of Mathematics}, 190\penalty0 (3):\penalty0 949--955, 2019.

\bibitem[Huh et~al.(2021)Huh, Mobahi, Zhang, Cheung, Agrawal, and Isola]{Huh2021TheLS}
Minyoung Huh, Hossein Mobahi, Richard Zhang, Brian Cheung, Pulkit Agrawal, and Phillip Isola.
\newblock The low-rank simplicity bias in deep networks.
\newblock \emph{Trans. Mach. Learn. Res.}, 2023, 2021.

\bibitem[Iyer et~al.(2017)Iyer, Dandekar, and Csernai]{qqp-data}
Shankar Iyer, Nikhil Dandekar, and Kornel Csernai.
\newblock First {Q}uora {D}ataset {R}elease: {Q}uestion {P}airs, 2017.
\newblock URL \url{https://quoradata.quora.com/First-Quora-Dataset-Release-Question-Pairs}.
\newblock Online.

\bibitem[Jawahar et~al.(2019)Jawahar, Sagot, and Seddah]{jawahar-etal-2019-bert}
Ganesh Jawahar, Beno{\^\i}t Sagot, and Djam{\'e} Seddah.
\newblock What does {BERT} learn about the structure of language?
\newblock In Anna Korhonen, David Traum, and Llu{\'\i}s M{\`a}rquez (eds.), \emph{Proceedings of the 57th Annual Meeting of the Association for Computational Linguistics}, pp.\  3651--3657, Florence, Italy, July 2019. Association for Computational Linguistics.
\newblock \doi{10.18653/v1/P19-1356}.
\newblock URL \url{https://aclanthology.org/P19-1356}.

\bibitem[Ji \& Telgarsky(2018)Ji and Telgarsky]{ji2018risk}
Ziwei Ji and Matus Telgarsky.
\newblock Risk and parameter convergence of logistic regression.
\newblock \emph{arXiv preprint arXiv:1803.07300}, 2018.

\bibitem[Ji \& Telgarsky(2020)Ji and Telgarsky]{tel2020directional}
Ziwei Ji and Matus Telgarsky.
\newblock Directional convergence and alignment in deep learning.
\newblock \emph{Advances in Neural Information Processing Systems}, 33:\penalty0 17176--17186, 2020.

\bibitem[Ji \& Telgarsky(2021)Ji and Telgarsky]{telgarskyngd2021}
Ziwei Ji and Matus Telgarsky.
\newblock Characterizing the implicit bias via a primal-dual analysis.
\newblock In \emph{Algorithmic Learning Theory}, pp.\  772--804. PMLR, 2021.

\bibitem[Ji et~al.(2021)Ji, Srebro, and Telgarsky]{tel2021dualacc}
Ziwei Ji, Nathan Srebro, and Matus Telgarsky.
\newblock Fast margin maximization via dual acceleration.
\newblock In \emph{International Conference on Machine Learning}, pp.\  4860--4869. PMLR, 2021.

\bibitem[Jiang* et~al.(2020)Jiang*, Neyshabur*, Mobahi, Krishnan, and Bengio]{Jiang*2020Fantastic}
Yiding Jiang*, Behnam Neyshabur*, Hossein Mobahi, Dilip Krishnan, and Samy Bengio.
\newblock Fantastic generalization measures and where to find them.
\newblock In \emph{International Conference on Learning Representations}, 2020.
\newblock URL \url{https://openreview.net/forum?id=SJgIPJBFvH}.

\bibitem[Jumper et~al.(2021)Jumper, Evans, Pritzel, Green, Figurnov, Ronneberger, Tunyasuvunakool, Bates, Z{\'i}dek, Potapenko, Bridgland, Meyer, Kohl, Ballard, Cowie, Romera-Paredes, Nikolov, Jain, Adler, Back, Petersen, Reiman, Clancy, Zielinski, Steinegger, Pacholska, Berghammer, Bodenstein, Silver, Vinyals, Senior, Kavukcuoglu, Kohli, and Hassabis]{Jumper2021HighlyAP}
John~M. Jumper, Richard Evans, Alexander Pritzel, Tim Green, Michael Figurnov, Olaf Ronneberger, Kathryn Tunyasuvunakool, Russ Bates, Augustin Z{\'i}dek, Anna Potapenko, Alex Bridgland, Clemens Meyer, Simon A~A Kohl, Andy Ballard, Andrew Cowie, Bernardino Romera-Paredes, Stanislav Nikolov, Rishub Jain, Jonas Adler, Trevor Back, Stig Petersen, David~A. Reiman, Ellen Clancy, Michal Zielinski, Martin Steinegger, Michalina Pacholska, Tamas Berghammer, Sebastian Bodenstein, David Silver, Oriol Vinyals, Andrew~W. Senior, Koray Kavukcuoglu, Pushmeet Kohli, and Demis Hassabis.
\newblock Highly accurate protein structure prediction with alphafold.
\newblock \emph{Nature}, 596:\penalty0 583 -- 589, 2021.
\newblock URL \url{https://api.semanticscholar.org/CorpusID:235959867}.

\bibitem[Keskar et~al.(2017)Keskar, Mudigere, Nocedal, Smelyanskiy, and Tang]{keskar2017on}
Nitish~Shirish Keskar, Dheevatsa Mudigere, Jorge Nocedal, Mikhail Smelyanskiy, and Ping Tak~Peter Tang.
\newblock On large-batch training for deep learning: Generalization gap and sharp minima.
\newblock In \emph{International Conference on Learning Representations}, 2017.
\newblock URL \url{https://openreview.net/forum?id=H1oyRlYgg}.

\bibitem[Kirichenko et~al.(2022)Kirichenko, Izmailov, and Wilson]{Kirichenko2022LastLR}
P.~Kirichenko, Pavel Izmailov, and Andrew~Gordon Wilson.
\newblock Last layer re-training is sufficient for robustness to spurious correlations.
\newblock \emph{ArXiv}, abs/2204.02937, 2022.

\bibitem[Kobayashi et~al.(2024)Kobayashi, Kuribayashi, Yokoi, and Inui]{kobayashi2024analyzing}
Goro Kobayashi, Tatsuki Kuribayashi, Sho Yokoi, and Kentaro Inui.
\newblock Analyzing feed-forward blocks in transformers through the lens of attention maps.
\newblock In \emph{The Twelfth International Conference on Learning Representations}, 2024.
\newblock URL \url{https://openreview.net/forum?id=mYWsyTuiRp}.

\bibitem[Kou et~al.(2023)Kou, Chen, and Gu]{qqgimplicit2023}
Yiwen Kou, Zixiang Chen, and Quanquan Gu.
\newblock Implicit bias of gradient descent for two-layer relu and leaky relu networks on nearly-orthogonal data, 2023.

\bibitem[Kovaleva et~al.(2019)Kovaleva, Romanov, Rogers, and Rumshisky]{kovaleva-etal-2019-revealing}
Olga Kovaleva, Alexey Romanov, Anna Rogers, and Anna Rumshisky.
\newblock Revealing the dark secrets of {BERT}.
\newblock In Kentaro Inui, Jing Jiang, Vincent Ng, and Xiaojun Wan (eds.), \emph{Proceedings of the 2019 Conference on Empirical Methods in Natural Language Processing and the 9th International Joint Conference on Natural Language Processing (EMNLP-IJCNLP)}, pp.\  4365--4374, Hong Kong, China, November 2019. Association for Computational Linguistics.
\newblock \doi{10.18653/v1/D19-1445}.
\newblock URL \url{https://aclanthology.org/D19-1445}.

\bibitem[Krizhevsky(2009)]{cifar}
Alex Krizhevsky.
\newblock Learning multiple layers of features from tiny images.
\newblock pp.\  32--33, 2009.
\newblock URL \url{https://www.cs.toronto.edu/~kriz/learning-features-2009-TR.pdf}.

\bibitem[LeCun \& Cortes(2005)LeCun and Cortes]{LeCun2005TheMD}
Yann LeCun and Corinna Cortes.
\newblock The mnist database of handwritten digits.
\newblock 2005.

\bibitem[Lee et~al.(2018)Lee, Sohl-dickstein, Pennington, Novak, Schoenholz, and Bahri]{lee2018deep}
Jaehoon Lee, Jascha Sohl-dickstein, Jeffrey Pennington, Roman Novak, Sam Schoenholz, and Yasaman Bahri.
\newblock Deep neural networks as gaussian processes.
\newblock In \emph{International Conference on Learning Representations}, 2018.
\newblock URL \url{https://openreview.net/forum?id=B1EA-M-0Z}.

\bibitem[Lee et~al.(2021)Lee, Lee, and Song]{lee2021vision}
Seung~Hoon Lee, Seunghyun Lee, and Byung~Cheol Song.
\newblock Vision transformer for small-size datasets, 2021.

\bibitem[Li et~al.(2021)Li, Luo, and Lyu]{li2021resolving-deepmatrix-ib}
Zhiyuan Li, Yuping Luo, and Kaifeng Lyu.
\newblock Towards resolving the implicit bias of gradient descent for matrix factorization: Greedy low-rank learning, 2021.

\bibitem[Li et~al.(2022)Li, Wang, Lee, and Arora]{li2022implicit}
Zhiyuan Li, Tianhao Wang, Jason~D Lee, and Sanjeev Arora.
\newblock Implicit bias of gradient descent on reparametrized models: On equivalence to mirror descent.
\newblock \emph{Advances in Neural Information Processing Systems}, 35:\penalty0 34626--34640, 2022.

\bibitem[Lim et~al.(2019)Lim, Kim, Kim, Kim, and Kim]{Lim2019FastA}
Sungbin Lim, Ildoo Kim, Taesup Kim, Chiheon Kim, and Sungwoong Kim.
\newblock Fast autoaugment.
\newblock In \emph{Neural Information Processing Systems}, 2019.

\bibitem[Liu et~al.(2019)Liu, Ott, Goyal, Du, Joshi, Chen, Levy, Lewis, Zettlemoyer, and Stoyanov]{liu2019roberta}
Yinhan Liu, Myle Ott, Naman Goyal, Jingfei Du, Mandar Joshi, Danqi Chen, Omer Levy, Mike Lewis, Luke Zettlemoyer, and Veselin Stoyanov.
\newblock Roberta: A robustly optimized bert pretraining approach, 2019.

\bibitem[Liu et~al.(2022)Liu, Mao, Wu, Feichtenhofer, Darrell, and Xie]{9879745}
Zhuang Liu, Hanzi Mao, Chao-Yuan Wu, Christoph Feichtenhofer, Trevor Darrell, and Saining Xie.
\newblock A convnet for the 2020s.
\newblock In \emph{2022 IEEE/CVF Conference on Computer Vision and Pattern Recognition (CVPR)}, pp.\  11966--11976, 2022.
\newblock \doi{10.1109/CVPR52688.2022.01167}.

\bibitem[Lyu \& Li(2020)Lyu and Li]{lyu2020Gradient}
Kaifeng Lyu and Jian Li.
\newblock Gradient descent maximizes the margin of homogeneous neural networks.
\newblock In \emph{International Conference on Learning Representations}, 2020.

\bibitem[Lyu et~al.(2021)Lyu, Li, Wang, and Arora]{Lyu2021GradientDO}
Kaifeng Lyu, Zhiyuan Li, Runzhe Wang, and Sanjeev Arora.
\newblock Gradient descent on two-layer nets: Margin maximization and simplicity bias.
\newblock In \emph{Neural Information Processing Systems}, 2021.

\bibitem[Mahmood et~al.(2021)Mahmood, Mahmood, and Van~Dijk]{mahmood2021robustness}
Kaleel Mahmood, Rigel Mahmood, and Marten Van~Dijk.
\newblock On the robustness of vision transformers to adversarial examples.
\newblock In \emph{Proceedings of the IEEE/CVF International Conference on Computer Vision}, pp.\  7838--7847, 2021.

\bibitem[Melas-Kyriazi(2021)]{MelasKyriazi2021DoYE}
Luke Melas-Kyriazi.
\newblock Do you even need attention? a stack of feed-forward layers does surprisingly well on imagenet.
\newblock \emph{ArXiv}, abs/2105.02723, 2021.
\newblock URL \url{https://api.semanticscholar.org/CorpusID:233864618}.

\bibitem[Meng et~al.(2022)Meng, Bau, Andonian, and Belinkov]{meng2022locating}
Kevin Meng, David Bau, Alex~J Andonian, and Yonatan Belinkov.
\newblock Locating and editing factual associations in {GPT}.
\newblock In Alice~H. Oh, Alekh Agarwal, Danielle Belgrave, and Kyunghyun Cho (eds.), \emph{Advances in Neural Information Processing Systems}, 2022.
\newblock URL \url{https://openreview.net/forum?id=-h6WAS6eE4}.

\bibitem[Morwani et~al.(2023)Morwani, Batra, Jain, and Netrapalli]{morwani2023simplicity}
Depen Morwani, Jatin Batra, Prateek Jain, and Praneeth Netrapalli.
\newblock Simplicity bias in 1-hidden layer neural networks, 2023.

\bibitem[Nacson et~al.(2019)Nacson, Lee, Gunasekar, Savarese, Srebro, and Soudry]{nacson2019convergence}
Mor~Shpigel Nacson, Jason Lee, Suriya Gunasekar, Pedro Henrique~Pamplona Savarese, Nathan Srebro, and Daniel Soudry.
\newblock Convergence of gradient descent on separable data.
\newblock In \emph{The 22nd International Conference on Artificial Intelligence and Statistics}, pp.\  3420--3428. PMLR, 2019.

\bibitem[Nagarajan et~al.(2021)Nagarajan, Andreassen, and Neyshabur]{nagarajan2021understanding}
Vaishnavh Nagarajan, Anders Andreassen, and Behnam Neyshabur.
\newblock Understanding the failure modes of out-of-distribution generalization.
\newblock In \emph{International Conference on Learning Representations}, 2021.
\newblock URL \url{https://openreview.net/forum?id=fSTD6NFIW_b}.

\bibitem[Nakkiran et~al.(2019)Nakkiran, Kaplun, Kalimeris, Yang, Edelman, Zhang, and Barak]{inc-comp}
Preetum Nakkiran, Gal Kaplun, Dimitris Kalimeris, Tristan Yang, Benjamin~L. Edelman, Fred Zhang, and Boaz Barak.
\newblock \emph{SGD on Neural Networks Learns Functions of Increasing Complexity}.
\newblock Curran Associates Inc., Red Hook, NY, USA, 2019.

\bibitem[Nanda et~al.(2023)Nanda, Chan, Lieberum, Smith, and Steinhardt]{nanda2023progress}
Neel Nanda, Lawrence Chan, Tom Lieberum, Jess Smith, and Jacob Steinhardt.
\newblock Progress measures for grokking via mechanistic interpretability, 2023.

\bibitem[Naseer et~al.(2021)Naseer, Ranasinghe, Khan, Hayat, Khan, and Yang]{Naseer2021IntriguingPO}
Muzammal Naseer, Kanchana Ranasinghe, Salman~Hameed Khan, Munawar Hayat, Fahad~Shahbaz Khan, and Ming-Hsuan Yang.
\newblock Intriguing properties of vision transformers.
\newblock In \emph{Neural Information Processing Systems}, 2021.
\newblock URL \url{https://api.semanticscholar.org/CorpusID:235125781}.

\bibitem[Netzer et~al.(2011)Netzer, Wang, Coates, Bissacco, Wu, and Ng]{svhn}
Yuval Netzer, Tao Wang, Adam Coates, Alessandro Bissacco, Bo~Wu, and Andrew~Y. Ng.
\newblock Reading digits in natural images with unsupervised feature learning.
\newblock In \emph{NIPS Workshop on Deep Learning and Unsupervised Feature Learning 2011}, 2011.
\newblock URL \url{http://ufldl.stanford.edu/housenumbers/nips2011_housenumbers.pdf}.

\bibitem[Neyshabur et~al.(2014)Neyshabur, Tomioka, and Srebro]{Neyshabur2014InSO}
Behnam Neyshabur, Ryota Tomioka, and Nathan Srebro.
\newblock In search of the real inductive bias: On the role of implicit regularization in deep learning.
\newblock \emph{CoRR}, abs/1412.6614, 2014.
\newblock URL \url{https://api.semanticscholar.org/CorpusID:6021932}.

\bibitem[Neyshabur et~al.(2017)Neyshabur, Bhojanapalli, McAllester, and Srebro]{Neyshabur2017ExploringGI}
Behnam Neyshabur, Srinadh Bhojanapalli, David McAllester, and Nathan Srebro.
\newblock Exploring generalization in deep learning.
\newblock In \emph{Neural Information Processing Systems}, 2017.
\newblock URL \url{https://api.semanticscholar.org/CorpusID:9597660}.

\bibitem[Novak et~al.(2018)Novak, Bahri, Abolafia, Pennington, and Sohl-Dickstein]{novak2018sensitivity}
Roman Novak, Yasaman Bahri, Daniel~A Abolafia, Jeffrey Pennington, and Jascha Sohl-Dickstein.
\newblock Sensitivity and generalization in neural networks: an empirical study.
\newblock \emph{arXiv preprint arXiv:1802.08760}, 2018.

\bibitem[Novak et~al.(2019)Novak, Xiao, Bahri, Lee, Yang, Abolafia, Pennington, and Sohl-dickstein]{novak2019bayesian}
Roman Novak, Lechao Xiao, Yasaman Bahri, Jaehoon Lee, Greg Yang, Daniel~A. Abolafia, Jeffrey Pennington, and Jascha Sohl-dickstein.
\newblock Bayesian deep convolutional networks with many channels are gaussian processes.
\newblock In \emph{International Conference on Learning Representations}, 2019.
\newblock URL \url{https://openreview.net/forum?id=B1g30j0qF7}.

\bibitem[O'Donnell(2014)]{o'donnell_2014}
Ryan O'Donnell.
\newblock \emph{Analysis of Boolean Functions}.
\newblock Cambridge University Press, 2014.
\newblock \doi{10.1017/CBO9781139814782}.

\bibitem[Paszke et~al.(2019)Paszke, Gross, Massa, Lerer, Bradbury, Chanan, Killeen, Lin, Gimelshein, Antiga, Desmaison, Köpf, Yang, DeVito, Raison, Tejani, Chilamkurthy, Steiner, Fang, Bai, and Chintala]{paszke2019pytorch}
Adam Paszke, Sam Gross, Francisco Massa, Adam Lerer, James Bradbury, Gregory Chanan, Trevor Killeen, Zeming Lin, Natalia Gimelshein, Luca Antiga, Alban Desmaison, Andreas Köpf, Edward Yang, Zach DeVito, Martin Raison, Alykhan Tejani, Sasank Chilamkurthy, Benoit Steiner, Lu~Fang, Junjie Bai, and Soumith Chintala.
\newblock Pytorch: An imperative style, high-performance deep learning library, 2019.

\bibitem[Paul \& Chen(2022)Paul and Chen]{paul2022vision}
Sayak Paul and Pin-Yu Chen.
\newblock Vision transformers are robust learners.
\newblock In \emph{Proceedings of the AAAI conference on Artificial Intelligence}, volume~36, pp.\  2071--2081, 2022.

\bibitem[Pezeshki et~al.(2020)Pezeshki, Kaba, Bengio, Courville, Precup, and Lajoie]{Pezeshki2020GradientSA}
Mohammad Pezeshki, Sekouba Kaba, Yoshua Bengio, Aaron~C. Courville, Doina Precup, and Guillaume Lajoie.
\newblock Gradient starvation: A learning proclivity in neural networks.
\newblock In \emph{Neural Information Processing Systems}, 2020.

\bibitem[Phuong \& Lampert(2021)Phuong and Lampert]{phuongortho-separable2021}
Mary Phuong and Christoph~H Lampert.
\newblock The inductive bias of re{\{}lu{\}} networks on orthogonally separable data.
\newblock In \emph{International Conference on Learning Representations}, 2021.
\newblock URL \url{https://openreview.net/forum?id=krz7T0xU9Z_}.

\bibitem[Power et~al.(2022)Power, Burda, Edwards, Babuschkin, and Misra]{power2022grokking}
Alethea Power, Yuri Burda, Harri Edwards, Igor Babuschkin, and Vedant Misra.
\newblock Grokking: Generalization beyond overfitting on small algorithmic datasets, 2022.

\bibitem[Quirke \& Barez(2024)Quirke and Barez]{quirke2024understanding}
Philip Quirke and Fazl Barez.
\newblock Understanding addition in transformers.
\newblock In \emph{International Conference on Learning Representations (ICLR)}, Vienna, Austria, 2024.

\bibitem[Raghu et~al.(2021)Raghu, Unterthiner, Kornblith, Zhang, and Dosovitskiy]{raghu2021do}
Maithra Raghu, Thomas Unterthiner, Simon Kornblith, Chiyuan Zhang, and Alexey Dosovitskiy.
\newblock Do vision transformers see like convolutional neural networks?
\newblock In A.~Beygelzimer, Y.~Dauphin, P.~Liang, and J.~Wortman Vaughan (eds.), \emph{Advances in Neural Information Processing Systems}, 2021.
\newblock URL \url{https://openreview.net/forum?id=Gl8FHfMVTZu}.

\bibitem[Rahaman et~al.(2019{\natexlab{a}})Rahaman, Baratin, Arpit, Draxler, Lin, Hamprecht, Bengio, and Courville]{rahaman19a}
Nasim Rahaman, Aristide Baratin, Devansh Arpit, Felix Draxler, Min Lin, Fred Hamprecht, Yoshua Bengio, and Aaron Courville.
\newblock On the spectral bias of neural networks.
\newblock In Kamalika Chaudhuri and Ruslan Salakhutdinov (eds.), \emph{Proceedings of the 36th International Conference on Machine Learning}, volume~97 of \emph{Proceedings of Machine Learning Research}, pp.\  5301--5310. PMLR, 09--15 Jun 2019{\natexlab{a}}.
\newblock URL \url{https://proceedings.mlr.press/v97/rahaman19a.html}.

\bibitem[Rahaman et~al.(2019{\natexlab{b}})Rahaman, Baratin, Arpit, Draxler, Lin, Hamprecht, Bengio, and Courville]{rahaman2019spectral}
Nasim Rahaman, Aristide Baratin, Devansh Arpit, Felix Draxler, Min Lin, Fred Hamprecht, Yoshua Bengio, and Aaron Courville.
\newblock On the spectral bias of neural networks.
\newblock In \emph{International conference on machine learning}, pp.\  5301--5310. PMLR, 2019{\natexlab{b}}.

\bibitem[Rebuffi et~al.(2021)Rebuffi, Gowal, Calian, Stimberg, Wiles, and Mann]{Rebuffi2021DataAC}
Sylvestre-Alvise Rebuffi, Sven Gowal, Dan~Andrei Calian, Florian Stimberg, Olivia Wiles, and Timothy Mann.
\newblock Data augmentation can improve robustness.
\newblock In \emph{Neural Information Processing Systems}, 2021.

\bibitem[Robbins(1951)]{Robbins1951ASA}
Herbert~E. Robbins.
\newblock A stochastic approximation method.
\newblock \emph{Annals of Mathematical Statistics}, 22:\penalty0 400--407, 1951.

\bibitem[Russakovsky et~al.(2015)Russakovsky, Deng, Su, Krause, Satheesh, Ma, Huang, Karpathy, Khosla, Bernstein, Berg, and Fei-Fei]{imagenet15russakovsky}
Olga Russakovsky, Jia Deng, Hao Su, Jonathan Krause, Sanjeev Satheesh, Sean Ma, Zhiheng Huang, Andrej Karpathy, Aditya Khosla, Michael Bernstein, Alexander~C. Berg, and Li~Fei-Fei.
\newblock {ImageNet Large Scale Visual Recognition Challenge}.
\newblock \emph{International Journal of Computer Vision (IJCV)}, 115\penalty0 (3):\penalty0 211--252, 2015.
\newblock \doi{10.1007/s11263-015-0816-y}.

\bibitem[Sagawa et~al.(2020)Sagawa, Koh, Hashimoto, and Liang]{Sagawa2020Distributionally}
Shiori Sagawa, Pang~Wei Koh, Tatsunori~B. Hashimoto, and Percy Liang.
\newblock Distributionally robust neural networks.
\newblock In \emph{International Conference on Learning Representations}, 2020.
\newblock URL \url{https://openreview.net/forum?id=ryxGuJrFvS}.

\bibitem[Shah et~al.(2020)Shah, Tamuly, Raghunathan, Jain, and Netrapalli]{shah2020pitfalls}
Harshay Shah, Kaustav Tamuly, Aditi Raghunathan, Prateek Jain, and Praneeth Netrapalli.
\newblock The pitfalls of simplicity bias in neural networks.
\newblock \emph{Advances in Neural Information Processing Systems}, 33, 2020.

\bibitem[Shao et~al.(2021)Shao, Shi, Yi, Chen, and Hsieh]{shao2021adversarial}
Rulin Shao, Zhouxing Shi, Jinfeng Yi, Pin-Yu Chen, and Cho-Jui Hsieh.
\newblock On the adversarial robustness of visual transformers.
\newblock \emph{arXiv preprint arXiv:2103.15670}, 1\penalty0 (2), 2021.

\bibitem[Shen et~al.(2023)Shen, Pu, Ji, Li, Zhang, Ge, and Wang]{shen2023improving}
Lujia Shen, Yuwen Pu, Shouling Ji, Changjiang Li, Xuhong Zhang, Chunpeng Ge, and Ting Wang.
\newblock Improving the robustness of transformer-based large language models with dynamic attention.
\newblock \emph{arXiv preprint arXiv:2311.17400}, 2023.

\bibitem[Shi et~al.(2020)Shi, Zhang, Chang, Huang, and Hsieh]{shi2020robustness}
Zhouxing Shi, Huan Zhang, Kai-Wei Chang, Minlie Huang, and Cho-Jui Hsieh.
\newblock Robustness verification for transformers.
\newblock \emph{arXiv preprint arXiv:2002.06622}, 2020.

\bibitem[Shi et~al.(2021)Shi, Wang, Zhang, Yi, and Hsieh]{shi2021fast}
Zhouxing Shi, Yihan Wang, Huan Zhang, Jinfeng Yi, and Cho-Jui Hsieh.
\newblock Fast certified robust training with short warmup.
\newblock \emph{Advances in Neural Information Processing Systems}, 34:\penalty0 18335--18349, 2021.

\bibitem[Soudry et~al.(2018)Soudry, Hoffer, Nacson, Gunasekar, and Srebro]{soudry2018implicit}
Daniel Soudry, Elad Hoffer, Mor~Shpigel Nacson, Suriya Gunasekar, and Nathan Srebro.
\newblock The implicit bias of gradient descent on separable data.
\newblock \emph{The Journal of Machine Learning Research}, 19\penalty0 (1):\penalty0 2822--2878, 2018.

\bibitem[Tarzanagh et~al.(2023{\natexlab{a}})Tarzanagh, Li, Thrampoulidis, and Oymak]{AtaeeTarzanagh2023TransformersAS}
Davoud~Ataee Tarzanagh, Yingcong Li, Christos Thrampoulidis, and Samet Oymak.
\newblock Transformers as support vector machines.
\newblock \emph{ArXiv}, abs/2308.16898, 2023{\natexlab{a}}.

\bibitem[Tarzanagh et~al.(2023{\natexlab{b}})Tarzanagh, Li, Zhang, and Oymak]{tarzanagh2023maxmargin}
Davoud~Ataee Tarzanagh, Yingcong Li, Xuechen Zhang, and Samet Oymak.
\newblock Max-margin token selection in attention mechanism, 2023{\natexlab{b}}.

\bibitem[Tiwari \& Shenoy(2023)Tiwari and Shenoy]{tiwari2023overcoming}
Rishabh Tiwari and Pradeep Shenoy.
\newblock Overcoming simplicity bias in deep networks using a feature sieve, 2023.

\bibitem[Trockman \& Kolter(2022{\natexlab{a}})Trockman and Kolter]{Trockman2022PatchesAA}
Asher Trockman and J.~Zico Kolter.
\newblock Patches are all you need?
\newblock \emph{Transactions on Machine Learning Research}, 2023, 2022{\natexlab{a}}.
\newblock URL \url{https://api.semanticscholar.org/CorpusID:245633423}.

\bibitem[Trockman \& Kolter(2022{\natexlab{b}})Trockman and Kolter]{trockman2022patches}
Asher Trockman and J~Zico Kolter.
\newblock Patches are all you need?, 2022{\natexlab{b}}.
\newblock URL \url{https://openreview.net/forum?id=TVHS5Y4dNvM}.

\bibitem[Valle-Perez et~al.(2019)Valle-Perez, Camargo, and Louis]{valle2018deep}
Guillermo Valle-Perez, Chico~Q Camargo, and Ard~A Louis.
\newblock Deep learning generalizes because the parameter-function map is biased towards simple functions.
\newblock \emph{International Conference on Learning Representations}, 2019.

\bibitem[Vardi(2022)]{vardi2022implicit}
Gal Vardi.
\newblock On the implicit bias in deep-learning algorithms, 2022.

\bibitem[Vasudeva et~al.(2023)Vasudeva, Shahabi, and Sharan]{vasudeva2023mitigating}
Bhavya Vasudeva, Kameron Shahabi, and Vatsal Sharan.
\newblock Mitigating simplicity bias in deep learning for improved ood generalization and robustness, 2023.

\bibitem[Vasudeva et~al.(2024)Vasudeva, Deora, and Thrampoulidis]{vasudeva2024implicit}
Bhavya Vasudeva, Puneesh Deora, and Christos Thrampoulidis.
\newblock Implicit bias and fast convergence rates for self-attention, 2024.

\bibitem[Vaswani et~al.(2017)Vaswani, Shazeer, Parmar, Uszkoreit, Jones, Gomez, Kaiser, and Polosukhin]{Vaswani2017AttentionIA}
Ashish Vaswani, Noam Shazeer, Niki Parmar, Jakob Uszkoreit, Llion Jones, Aidan~N Gomez, \L~ukasz Kaiser, and Illia Polosukhin.
\newblock Attention is all you need.
\newblock In I.~Guyon, U.~Von Luxburg, S.~Bengio, H.~Wallach, R.~Fergus, S.~Vishwanathan, and R.~Garnett (eds.), \emph{Advances in Neural Information Processing Systems}, volume~30. Curran Associates, Inc., 2017.
\newblock URL \url{https://proceedings.neurips.cc/paper_files/paper/2017/file/3f5ee243547dee91fbd053c1c4a845aa-Paper.pdf}.

\bibitem[von Oswald et~al.(2022)von Oswald, Niklasson, Randazzo, Sacramento, Mordvintsev, Zhmoginov, and Vladymyrov]{Oswald2022TransformersLI}
Johannes von Oswald, Eyvind Niklasson, E.~Randazzo, Jo{\~a}o Sacramento, Alexander Mordvintsev, Andrey Zhmoginov, and Max Vladymyrov.
\newblock Transformers learn in-context by gradient descent.
\newblock In \emph{International Conference on Machine Learning}, 2022.

\bibitem[Wang et~al.(2021)Wang, Lin, Liu, Zheng, Wang, and Zha]{wang2021macrobert}
Fali Wang, Zheng Lin, Zhengxiao Liu, Mingyu Zheng, Lei Wang, and Daren Zha.
\newblock Macrobert: Maximizing certified region of bert to adversarial word substitutions.
\newblock In \emph{Database Systems for Advanced Applications: 26th International Conference, DASFAA 2021, Taipei, Taiwan, April 11--14, 2021, Proceedings, Part II 26}, pp.\  253--261. Springer, 2021.

\bibitem[Wang et~al.(2022)Wang, Variengien, Conmy, Shlegeris, and Steinhardt]{wang2022interpretability}
Kevin Wang, Alexandre Variengien, Arthur Conmy, Buck Shlegeris, and Jacob Steinhardt.
\newblock Interpretability in the wild: a circuit for indirect object identification in gpt-2 small, 2022.

\bibitem[Xiao et~al.(2017)Xiao, Rasul, and Vollgraf]{fm}
Han Xiao, Kashif Rasul, and Roland Vollgraf.
\newblock Fashion-mnist: a novel image dataset for benchmarking machine learning algorithms, 2017.

\bibitem[Xu et~al.(2019)Xu, Zhang, Luo, Xiao, and Ma]{xu2019frequency}
Zhi-Qin~John Xu, Yaoyu Zhang, Tao Luo, Yanyang Xiao, and Zheng Ma.
\newblock Frequency principle: Fourier analysis sheds light on deep neural networks.
\newblock \emph{arXiv preprint arXiv:1901.06523}, 2019.

\bibitem[Yang(2021)]{yang2021tensor}
Greg Yang.
\newblock Tensor programs i: Wide feedforward or recurrent neural networks of any architecture are gaussian processes, 2021.

\bibitem[Yang \& Salman(2020)Yang and Salman]{yang2020finegrained}
Greg Yang and Hadi Salman.
\newblock A fine-grained spectral perspective on neural networks, 2020.

\bibitem[Zhang et~al.(2018)Zhang, Cisse, Dauphin, and Lopez-Paz]{zhang2018mixup}
Hongyi Zhang, Moustapha Cisse, Yann~N. Dauphin, and David Lopez-Paz.
\newblock mixup: Beyond empirical risk minimization.
\newblock In \emph{International Conference on Learning Representations}, 2018.
\newblock URL \url{https://openreview.net/forum?id=r1Ddp1-Rb}.

\end{thebibliography}
